\documentclass{article} 
\PassOptionsToPackage{table}{xcolor} 
\usepackage{iclr2025_conference,times}

\usepackage{amsmath,amsfonts,bm}

\def\eqref#1{equation~\ref{#1}}

\def\1{\bm{1}}

\DeclareMathAlphabet{\mathsfit}{\encodingdefault}{\sfdefault}{m}{sl}
\SetMathAlphabet{\mathsfit}{bold}{\encodingdefault}{\sfdefault}{bx}{n}

\usepackage{hyperref}
\usepackage{url}
\usepackage{graphicx}
\usepackage{subcaption}
\title{AutoML Benchmark with shorter time constraints and early stopping}

\author{Israel Campero Jurado \\
Mathematics and Computer Science\\
Eindhoven University of Technology\\
5612 AZ Eindhoven, The Netherlands \\
\texttt{i.campero.jurado@tue.nl} \\
\And
Pieter Gijsbers \\ 
Mathematics and Computer Science\\
Eindhoven University of Technology\\
5612 AZ Eindhoven, The Netherlands \\
\texttt{p.gijsbers@tue.nl} \\
\And
Joaquin Vanschoren \\ 
Mathematics and Computer Science\\
Eindhoven University of Technology\\
5612 AZ Eindhoven, The Netherlands \\
\texttt{j.vanschoren@tue.nl} \\
}

\iclrfinalcopy 
\begin{document}

\maketitle

\begin{abstract}
Automated Machine Learning (AutoML) automatically builds machine learning (ML) models on data. The de facto standard for evaluating new AutoML frameworks for tabular data is the AutoML Benchmark (AMLB). AMLB proposed to evaluate AutoML frameworks using 1- and 4-hour time budgets across 104 tasks. 
We argue that shorter time constraints should be considered for the benchmark because of their practical value, such as when models need to be retrained with high frequency, and to make AMLB more accessible.
This work considers two ways in which to reduce the overall computation used in the benchmark: smaller time constraints and the use of early stopping.
We conduct evaluations of 11 AutoML frameworks on 104 tasks with different time constraints and find the relative ranking of AutoML frameworks is fairly consistent across time constraints, but that using early-stopping leads to a greater variety in model performance.
\end{abstract}







\section{Introduction}

In Machine Learning (ML), manually creating good models is time-consuming and knowledge-intensive. Automated Machine Learning (AutoML) employs efficient automated search methods to create models for new data, often  reducing the computational costs in the process \cite{hutter2019automated, hollmann2022tabpfn}. The AutoML Benchmark (AMLB,~\citealt{gijsbers2024amlb}) has become the standard for the evaluation of AutoML frameworks on tabular data, greatly increasing reproducibility and comparability in AutoML research. Despite the effectiveness of AMLB, evaluating even a single method on the entire benchmark requires 40,000 CPU hours\footnote{40,000 CPU hours $\approx$ 104 datasets * 10 folds * 8 cores * (1 + 4 hour time constraints)}. While new work can compare directly to published results, even the cost for evaluating a single method may be prohibitive.

We identified that the time budgets proposed by \citet{gijsbers2024amlb} were based on what seemed "practically reasonable" at the time, as signified by many frameworks' default time budget of one hour. While the authors motivate evaluating methods on two time budgets as a proxy for anytime performance, they do not motivate the particular choice of 1 hour and 4 hours. 
AutoML frameworks behave under different time constraints.
We conduct similar experiments and analyses for frameworks with early-stopping, offering insights into its potential to reduce energy consumption in AutoML systems. 

\section{Related Literature} \label{cross:relatedwork}
AMLB \cite {amlb2019} introduced benchmarking suites to standardize evaluations in AutoML research, and provided software which can automatically install and evaluate an AutoML framework, increasing reproducibility and a fairer comparison of frameworks. However, we often see that the original benchmarking suite or time constraints (1 hour and 4 hour) are not used as proposed. For example, \citet{wang2021flaml} also use one and ten minutes time constraints, arguing that in some dynamic systems only a few CPU minutes may be allocated to training models. Also, in \cite {zoller2021benchmark} experiments were conducted with a 1-hour cutoff, as tests on 10 randomly selected datasets showed only marginal performance improvements at 4 and 8 hours, making longer timeouts unnecessary. On the other hand, sometimes datasets are omitted from the evaluation due to computational costs~\cite{kozak2023forester, bergman2024don, kenny2024using, bonet2024hyperfast, hajiramezanali2022stab}. 
Finally, black-box optimization, the biggest time sink for most AutoML frameworks, may become less significant to model performance with advancements such as large scale experiment repositories for portfolio selection~\cite{tabrepo, feurer2022auto} and the development of foundation models for tabular data that generate predictions in seconds~\cite{hollmann2025accurate, zhu2023xtab}.

\section{Methodology}\label{cross:methodology}  
This study assesses the effect of reducing AMLB's computational cost, focusing on shorter time constraints (5, 10, 30, and 60 minutes).
Th relative to the existing 1 and 4-hour defaultse purpose is twofold. First, shorter time constraints reflect real-world scenarios where rapid model deployment is crucial, such as in dynamic environments or when computational resources are limited, and thus warrant a benchmark of their own.
Second, we identified that the computational cost of 1 and 4 hour time constraints can be prohibitive, and we hope to learn whether or not evaluations with shorter time constraints may lead to similar conclusions using less resources.
For similar reasons, we also evaluate the effectiveness of early stopping. 

We evaluate AutoML frameworks on the benchmarking suite by \citet{gijsbers2024amlb}, comprising 71 classification and 33 regression tasks from OpenML~\citep{vaschoren2014openml} and also adopt their baselines. The frameworks evaluated are $\mathtt{AutoGluon}$~\cite{erickson2020autogluon}, $\mathtt{autosklearn}$~\cite{feurer2015efficient}, $\mathtt{autosklearn2}$~\cite{feurer2022auto}, $\mathtt{flaml}$~\cite{wang2021flaml}, $\mathtt{GAMA}$~\cite{gijsbers2021gama} (in ``performance'' setup), $\mathtt{H2OAutoML}$~\cite{ledell2020h2o}, $\mathtt{lightAutoML}$~\cite{}, $\mathtt{MLJAR(B)}$~\cite{mljar} (in ``Compete'' setup),  $\mathtt{TPOT}$~\cite{olson2016tpot}. For $\mathtt{AutoGluon}$ multiple configurations are evaluated varying in training and inference time, from slowest  to fastest: $\mathtt{AutoGluon(B)}$,  $\mathtt{AutoGluon(HQ)}$, $\mathtt{AutoGluon(HQIL)}$. In the early stopping experiment, we also evaluate a setting which further optimizes training time,$\mathtt{AutoGluon(FIFTIL)}$, and we include its result in Appendix \ref{cross:appendix_performance_training_time}.
We utilized various configurations of the $\mathtt{AutoGluon}$ framework because it has demonstrated state-of-the-art performance in tabular tasks.

Not all AutoML frameworks allow early stopping, limiting our evaluation to $\mathtt{flaml}$, $\mathtt{H2OAutoML}$, $\mathtt{TPOT}$, $\mathtt{FEDOT}$ and $\mathtt{AutoGluon}$. For each framework, we use early stopping behavior as defined by their defaults or documentation. To see the full setup description, please refer to Appendix \ref{cross:appendix_performance_training_time}. 

\section{Results}\label{cross:results}

\subsection{The Effect of Time Constraints on Performance}
\begin{figure}[h]
\begin{center}

\begin{subfigure}{0.2\textwidth}
\includegraphics[width=\linewidth]{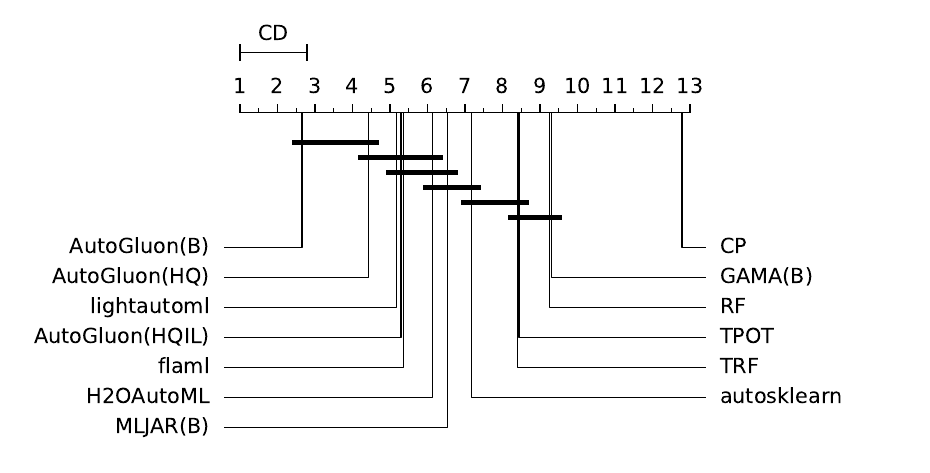}
\caption{5 minutes}
\end{subfigure}
\begin{subfigure}{0.2\textwidth}
\includegraphics[width=\linewidth]{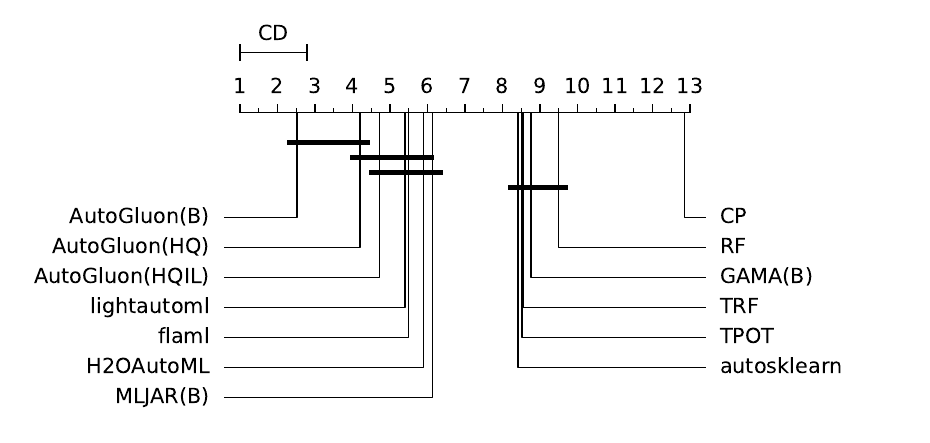}
\caption{10 minutes}
\end{subfigure}
\begin{subfigure}{0.2\textwidth}
\includegraphics[width=\linewidth]{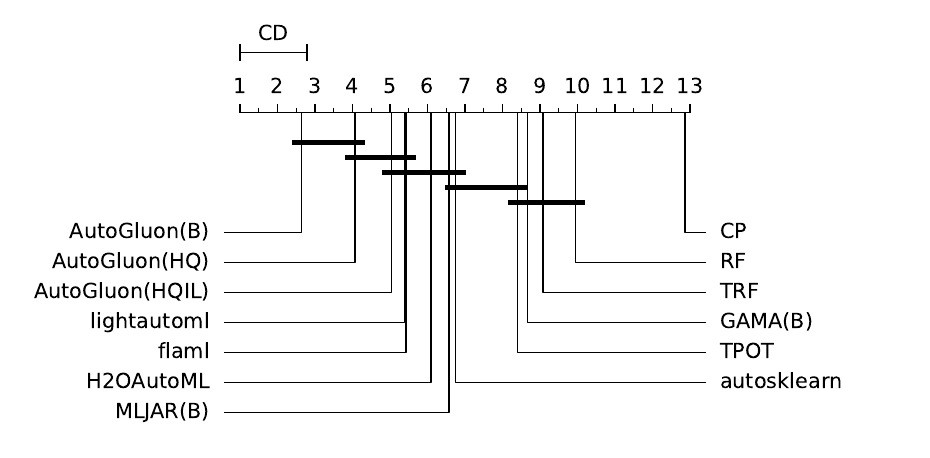}
\caption{30 minutes}
\end{subfigure}
\begin{subfigure}{0.2\textwidth}
\includegraphics[width=\linewidth]{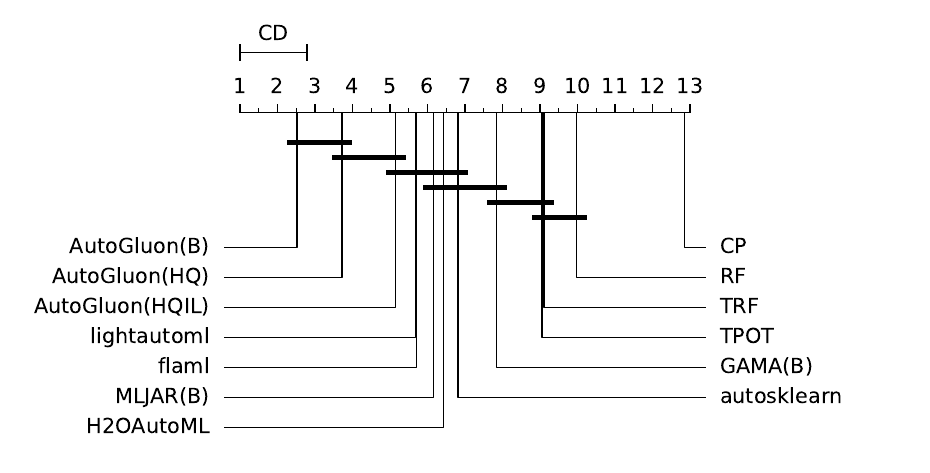}
\caption{60 minutes}
\end{subfigure}

\end{center}
\caption{Critical Diagrams (CD) of the evaluated frameworks by time constraints. These diagrams include the Nemenyi post-hoc test. \label{fig:critical_diagrams_by_time}}
\end{figure}

We first report on the results for shorter time constraints, and then aim to answer the question of whether shorter time constraints may also be used as a substitute or indicator for larger time constraints. Figure \ref{fig:critical_diagrams_by_time} shows the critical difference (CD) diagrams for each time constraint and depicts the mean rank of each framework (lower is better) with a Nemenyi post-hoc test to assess statistical differences in those ranks\footnote{$\mathtt{autosklearn2}$ and $\mathtt{NaiveAutoML}$ are removed in the CDs comparison due to problems described in \citet{gijsbers2024amlb}. FEDOT is similarly omitted due to issues with their integration \url{https://github.com/aimclub/FEDOT/issues/1268}} \cite{demvsar2006statistical}. The ranks are computed as suggested by \citet{gijsbers2024amlb}, by first imputing missing results with constant predictor ($\mathtt{CP}$) performance. AutoML framework ranks are very consistent across time constraints. $\mathtt{AutoGluon(B)}$ achieves the best rank for all time constraints, which may be explained by its use of a meta-learned portfolio, so it does not need much time for model selection. On the other hand, $\mathtt{GAMA}$ and $\mathtt{TPOT}$ do not always outperform the strongest baseline, $\mathtt{TRF}$. These AutoML frameworks use evolutionary optimization and do not employ any portfolio or warm-starting, which means that if they do not have time to evaluate the initial population, they are effectively doing random search. 

Table \ref{table:ranking_overtime} displays the differences in rank by time. The top row shows mean rank with a 60 minute time constraint. Below, each row indicates the difference in rank for different time constraints. Positive values in shorter times reflect worse mean rank compared to 60 minutes (red cells), while negative values indicate an improvement in mean rank (green cells). Since ranks are relative, any improvement or deterioration in the rank of a framework can be attributed either to its own performance changes or to the performance shifts of other frameworks. As we expect, some frameworks experience more errors with smaller time constraints, such as for $\mathtt{GAMA(B)}$ and $\mathtt{AutoGluon(HQIL)}$ on a 5 minute constraint. $\mathtt{AutoGluon(HQIL)}$ improves relative to $\mathtt{AutoGluon(HQ)}$ in 30 and 10 minute constraints, perhaps because without the inference time limitations the ensembling process is slowed down by high inference models.

\begin{table}[h]
\centering
\caption{Ranking performance of AutoML frameworks at varying time constraints (60, 30, 10, 5 minutes). Positive values in shorter times reflect worse performance compared to 60 minutes, while negative values indicate an improvement in ranking. The winner for each time is in bold type $\mathtt{AutoGluon(B)}$ and $\mathtt{AutoGluon(HQ)}$ since it does not have a statistically different rank.}
\label{table:ranking_overtime}
\scalebox{0.6}{
\begin{tabular}{r|rrrrrrrrrr}
\textbf{}      & \rotatebox{90}{\textbf{AutoGluon(B)}}                  & \rotatebox{90}{\textbf{AutoGluon(HQ)}}                 & \rotatebox{90}{\textbf{AutoGluon(HQIL)}}      & \rotatebox{90}{\textbf{lightautoml}}          & \rotatebox{90}{\textbf{flaml}}                & \rotatebox{90}{\textbf{MLJAR(B)}}             & \rotatebox{90}{\textbf{H2OAutoML}}            & \rotatebox{90}{\textbf{autosklearn}}          & \rotatebox{90}{\textbf{GAMA(B)}}              & \rotatebox{90}{\textbf{TPOT}}                                                     \\ \hline
\textbf{60min} & \textbf{2.52}                          & \textbf{3.72}                          & 5.15                          & 5.68                          & 5.71                          & 6.16                          & 6.42                          & 6.82                          & 7.85                          & 9.05                                                              \\
\textbf{30min} & \cellcolor[HTML]{FFCE93}\textbf{$+$0.12} & \cellcolor[HTML]{FFCE93}\textbf{$+$0.35} & \cellcolor[HTML]{A8E7A7}-0.11 & \cellcolor[HTML]{A8E7A7}-0.30 & \cellcolor[HTML]{A8E7A7}-0.28 & \cellcolor[HTML]{FFCE93}$+$0.41 & \cellcolor[HTML]{A8E7A7}-0.33 & \cellcolor[HTML]{A8E7A7}-0.07 & \cellcolor[HTML]{FFCE93}$+$0.82 & \cellcolor[HTML]{A8E7A7}-0.64                                     \\
\textbf{10min} & \cellcolor[HTML]{FFFFFF}\textbf{0.0}   & \cellcolor[HTML]{FFCE93}\textbf{$+$0.48} & \cellcolor[HTML]{A8E7A7}-0.44 & \cellcolor[HTML]{A8E7A7}-0.28 & \cellcolor[HTML]{A8E7A7}-0.21 & \cellcolor[HTML]{A8E7A7}-0.02 & \cellcolor[HTML]{A8E7A7}-0.53 & \cellcolor[HTML]{FFCE93}$+$1.59 & \cellcolor[HTML]{FFCE93}$+$0.91 & \cellcolor[HTML]{A8E7A7}{-0.52} \\
\textbf{5min}  & \cellcolor[HTML]{FFCE93}\textbf{$+$0.13} & \cellcolor[HTML]{FFCE93}\textbf{$+$0.71} & \cellcolor[HTML]{FFCE93}$+$0.14 & \cellcolor[HTML]{A8E7A7}-0.51 & \cellcolor[HTML]{A8E7A7}-0.35 & \cellcolor[HTML]{FFCE93}$+$0.38 & \cellcolor[HTML]{A8E7A7}-0.28 & \cellcolor[HTML]{FFCE93}$+$0.35 & \cellcolor[HTML]{FFCE93}$+$1.46 & \cellcolor[HTML]{A8E7A7}-0.61                                    
\end{tabular}
}
\end{table}

\begin{figure}[h]
\begin{center}

\begin{subfigure}{0.2\textwidth}
\includegraphics[width=\linewidth, height=2cm]{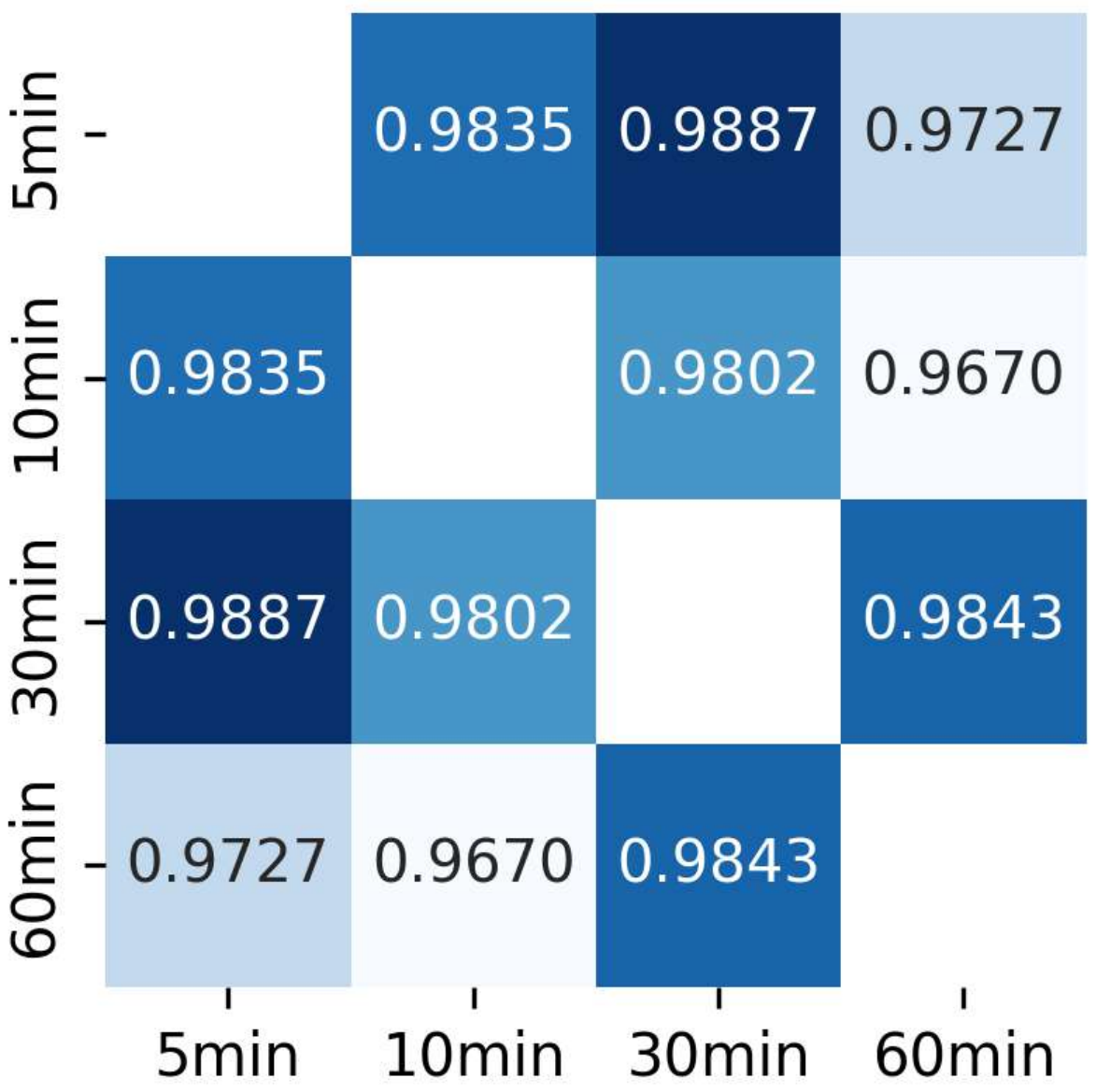}
\caption{Ranking correlation over time}
\end{subfigure}
\hspace{0.05\textwidth}
\begin{subfigure}{0.25\textwidth}
\includegraphics[width=\linewidth, height=2cm]{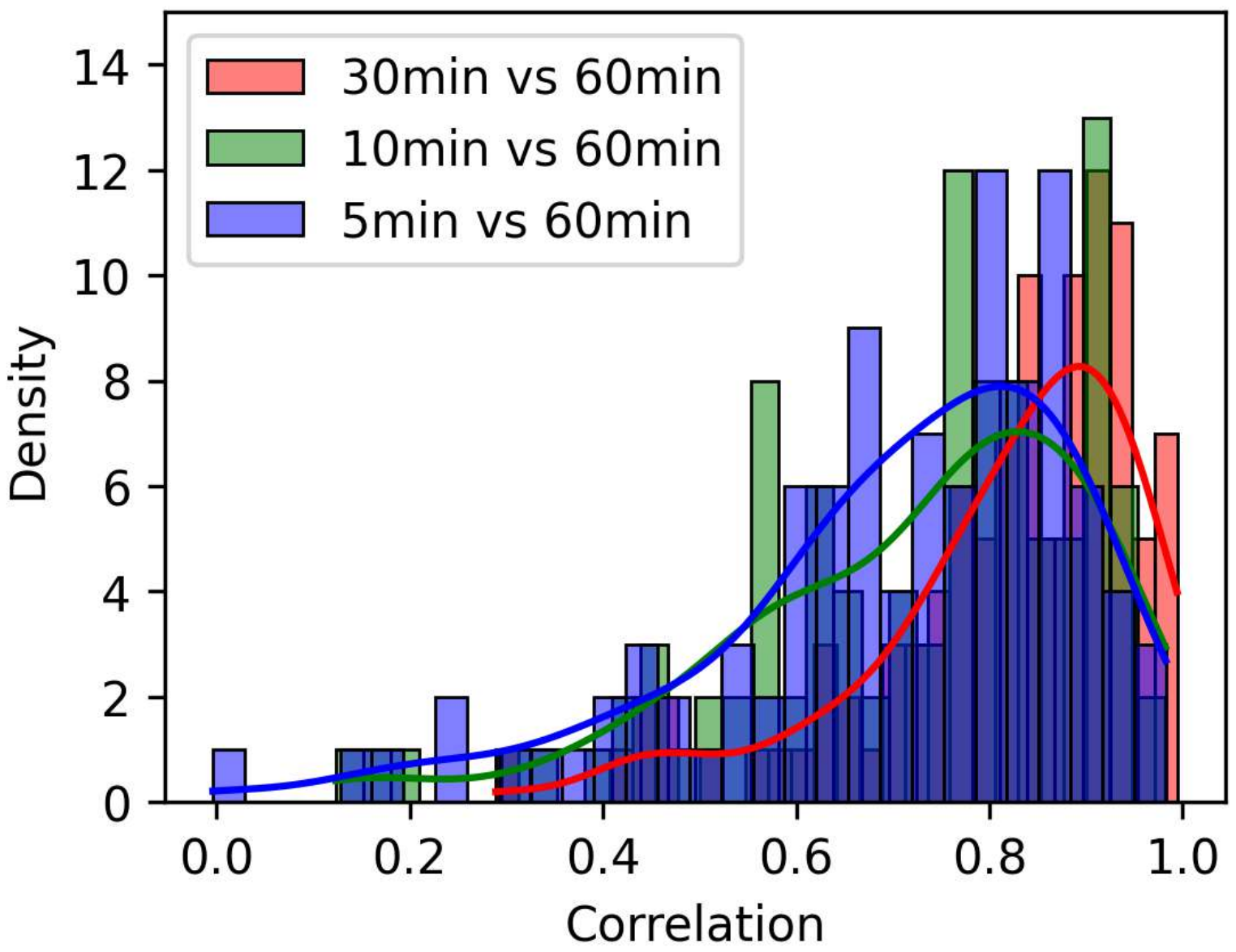}
\caption{Histogram correlation of ranking position over time}
\end{subfigure}
\hspace{0.05\textwidth}
\begin{subfigure}{0.25\textwidth}
\includegraphics[width=\linewidth, height=2cm]{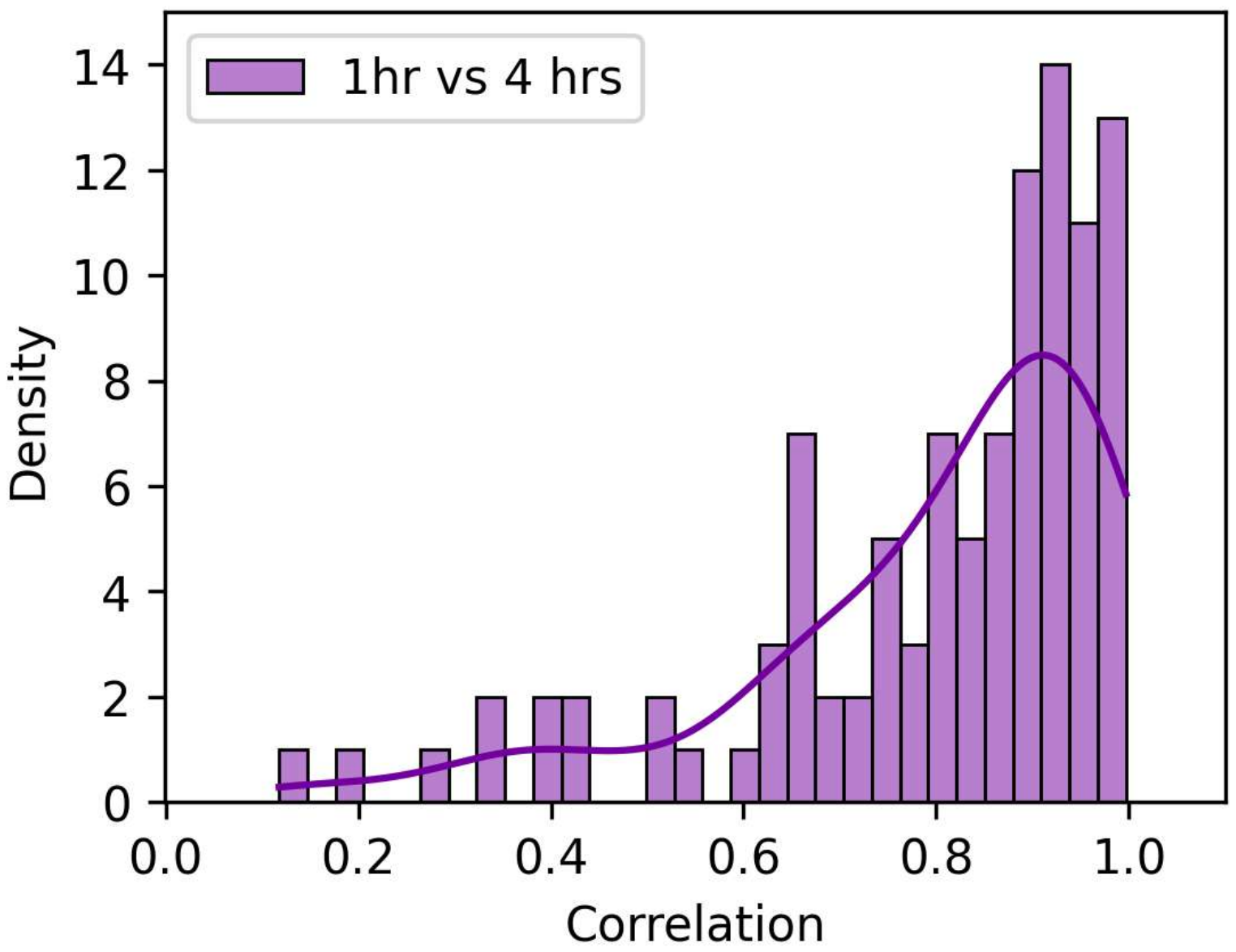}
\caption{Histogram correlation of ranking position 1hr vs 4hr}
\end{subfigure}
\end{center}
\caption{(a) Confusion matrix based on ranking's correlation mean performance per task over the times. (b) Histogram and Density plot of the correlation from the ranking of each framework by task at each time (c) Histogram and Density plot of the correlation from the ranking of each framework by task from the original AMLB results 1 hour vs 4 hours \label{fig:rankingpositions}}
\end{figure}

We examine the correlation between results across time constraints to assess if shorter ones can substitute when resources are scarce. Figure \ref{fig:rankingpositions} (a) shows the Pearson correlation (since it is the most sensitive to outliers) of the rankings from the frameworks over time. We see that across time constraints, the correlation of the ranking is very high ($>0.96$). There may be multiple explanations for this. For example, it is possible that most optimization takes place in the first 5 minutes. It is also possible that rankings are preserved as they are largely defined by the frameworks' optimization methods and search spaces. Another reason might be the nature of the datasets used. If datasets are not excessively complex or if they are of moderate size, the computational burden is lower, allowing frameworks to explore the search space more thoroughly, even within shorter time constraints. Figure \ref{fig:rankingpositions} (b) shows the distribution of the correlation of the rankings per dataset between specific time pairs (over 104 datasets 10 folds). As observed earlier, rankings across all the time constraints have strong positive correlations with the 60-minute runs. In 5 minutes, many frameworks are likely only exploring a small portion of the search space, leading to more variability in performance compared to the 60-minute runs. With 10 minutes, the search space exploration is more complete. The narrow, tall peak around high correlation values in the 30 vs. 60 minutes comparison indicates that 30 minutes might be considered sufficient to achieve results close to the 60-minute runs. By 30 minutes, the AutoML frameworks have likely had enough time to find a strong local optimum in the search space, including tuning important hyperparameters, selecting appropriate models, and possibly building robust ensembles. To contrast this idea, the original results from the AMLB paper are compared in Figure \ref{fig:rankingpositions} (c), where the evaluation was conducted under 1-hour and 4-hour time constraints. The peak around 0.9 and above, along with a tighter distribution, indicates that results between 1-hour and 4-hour runs are highly consistent, with few instances where the correlation drops below 0.7. This suggests after 60 minutes, any additional search might help to refine the solution marginally, leading to a plateau in performance.

\subsection{Behaviour over Time Constraints}

\begin{figure}[h]
\centering
\begin{subfigure}{0.4\textwidth}
    \includegraphics[width=\linewidth, height=3cm]{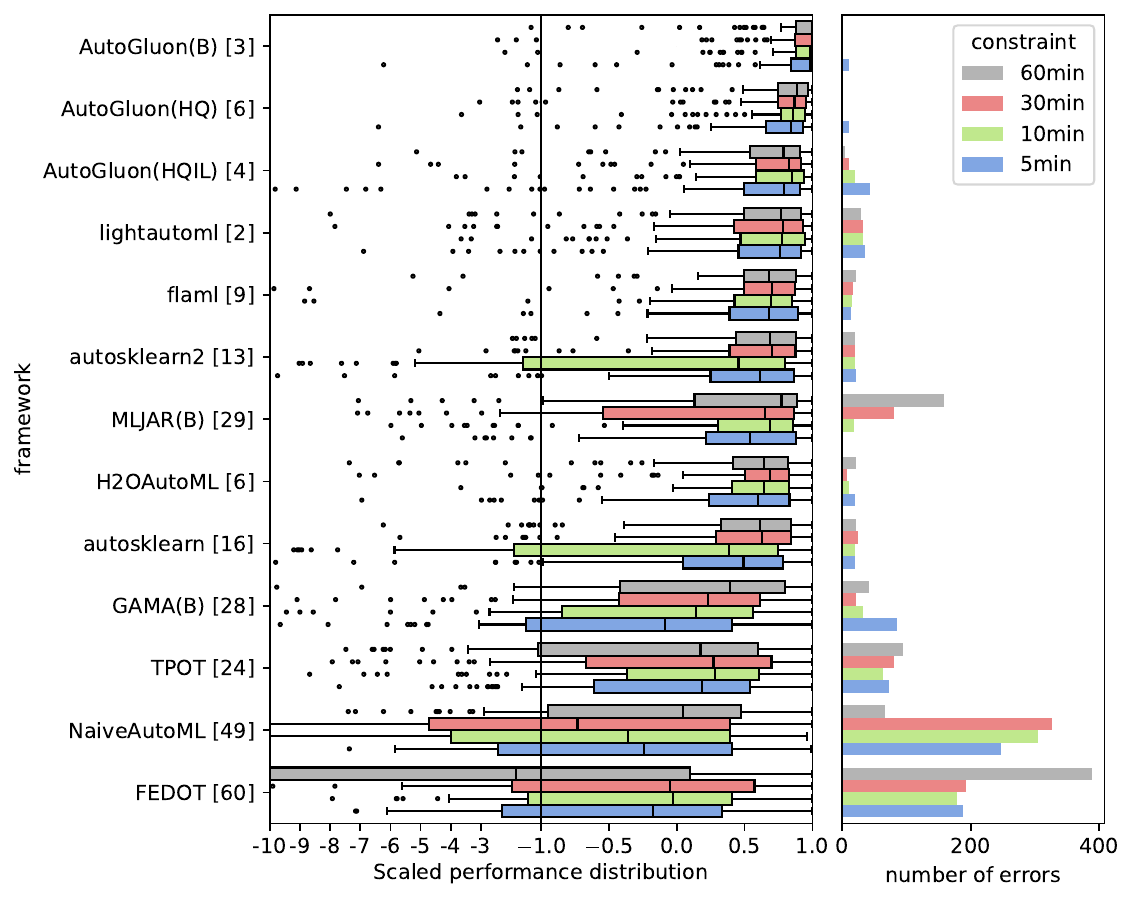}
    \caption{The performance distribution of each framework across different time constraints, along with the number of errors generated. Runs lying outside the graph for each framework are summarized on the y-axis in brackets, where the displayed number represents the total count of such experiments across all four-time constraints.}
\end{subfigure}
\hfill
\begin{subfigure}{0.5\textwidth}
    \includegraphics[width=\linewidth, height=5cm]{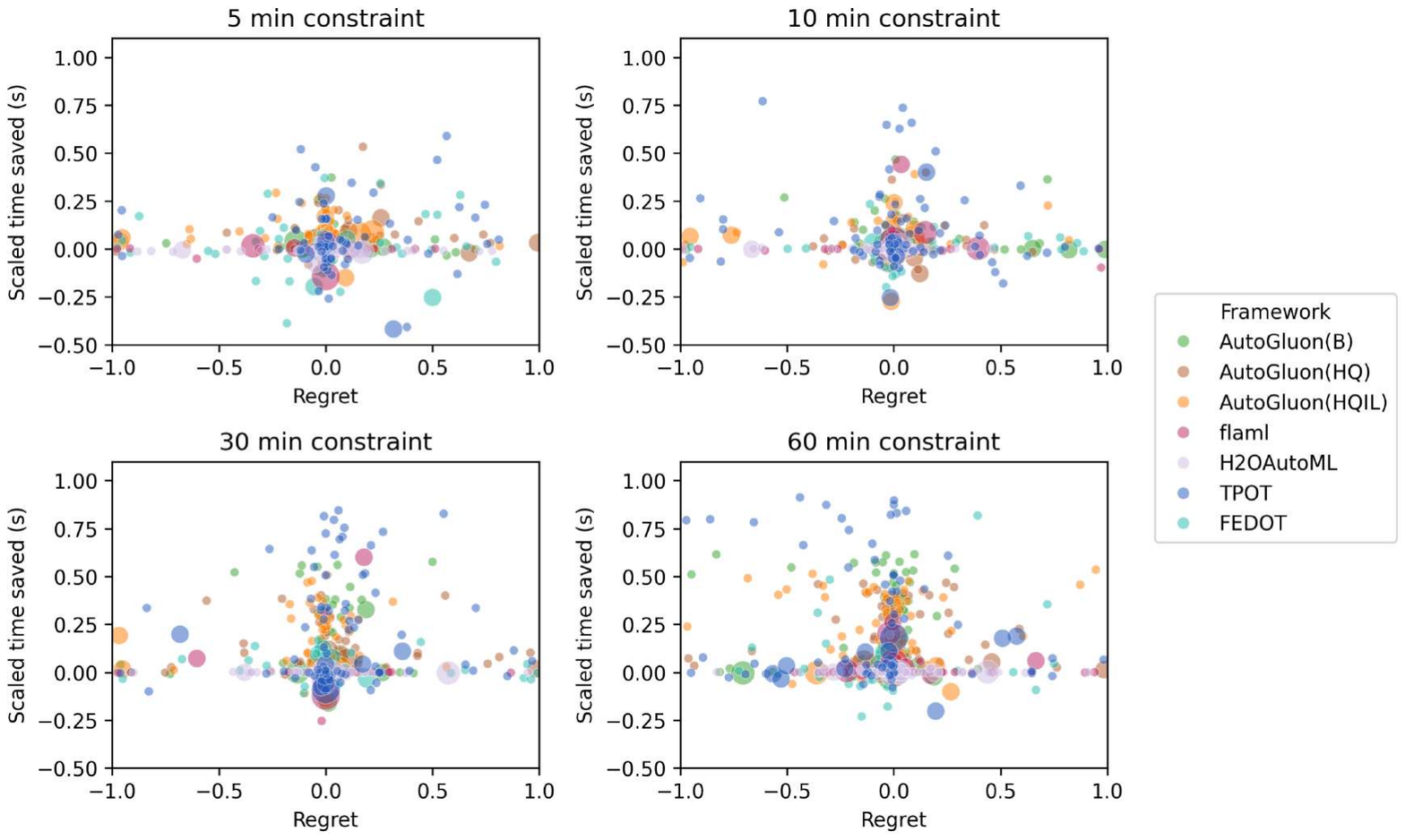}
    \caption{\textit{regret} and \textit{scaled time saved} across time constraints. The size of the dot corresponds to the dataset size, from instances $< 100k$ to instances $> 5M$. The values in the Scaled time saved (s) represent the proportion of time saved relative by the constraint time. The maximum value could approach 1 (or 100\%) and the minimum value could be 0, meaning no time was saved.}
\end{subfigure}
\caption{Performance distribution (a) and \textit{regret} and \textit{time saved} (b) across the AutoML frameworks. \label{fig:performance_and_early}}
\end{figure}

To illustrate to what extent the frameworks have improved over time, Figure \ref{fig:performance_and_early} (a) shows the scaled performance distribution of each framework under each time constraint. The results are scaled by task using the performance of the random forest ($\mathtt{RF}$) performance (0) and the best-observed performance (1). It also includes the number of errors encountered per framework at each time constraint. Some frameworks show notable improvements with increased time, indicating that they leverage extended computation more effectively. In particular, $\mathtt{AutoGluon}$ consistently exhibited the lowest error rates, while $\mathtt{H2OAutoML}$ and $\mathtt{flaml}$ also maintained relatively stable performance with fewer failures across different time constraints. The relationship between error rates and the performance distribution is particularly evident in frameworks such as $\mathtt{NaiveAutoML}$ and $\mathtt{FEDOT}$, where a higher number of failures leads to a downward shift in performance. This is a direct consequence of the way imputation is handled with $\mathtt{CP}$, as more frequent failures result in greater performance degradation. Some frameworks, such as $\mathtt{MLJAR(B)}$ and $\mathtt{FEDOT}$, appear to perform better under shorter time constraints, where fewer errors contribute to more stable outcomes.

\subsection{Early stopping in AutoML systems}

To evaluate early stopping, we measure two metrics: (1) \textit{regret} (-1 to 1), the performance difference between setups with and without early stopping (higher values indicate worse outcomes); and (2) \textit{scaled time saved}, the training time reduction (as a proportion of the constraint time) when using early stopping, where higher values are preferable. 

Figure \ref{fig:performance_and_early} (b) visualizes the relationship between \textit{regret} and \textit{scaled time saved} across time constraints. Each dot represents an AutoML framework on a dataset, with its size indicating the size of the dataset. 5 and 10 minutes, early stopping has limited impact under these time constraints (the runs rarely surpass 50\% of \textit{time saved}), likely because the overall training process is already very short. In 30 minutes, we see more variance in \textit{scaled time saved}, particularly for larger datasets. Frameworks like $\mathtt{TPOT}$ show some runs with higher \textit{regret} and \textit{scaled time saved}, which could indicate that early stopping cuts off promising learning trajectories prematurely, leading to suboptimal models, $\mathtt{AutoGluon}$ shows signs of save more time after 30 minutes while maintaining a \textit{regret} near zero. At 60 minutes, the spread of both \textit{regret} and \textit{time saved} becomes more pronounced. $\mathtt{AutoGluon}$ and $\mathtt{TPOT}$ can save time while reducing \textit{regret}, but sometimes also sacrifice time for performance. $\mathtt{FEDOT}$ and $\mathtt{H2OAutoML}$ maintain stable performance with smaller \textit{time saved} values, suggesting that these frameworks may not have been strongly influenced by early stopping.

The prevalence of negative regret suggests that early stopping not only reduces training time but often improves model performance. This phenomenon may arise because frameworks that converge quickly to near-optimal solutions may avoid wasteful exploration of redundant configurations when stopped early, effectively ``locking in'' better-performing models. For larger datasets, early stopping could prevent computationally expensive over-refinement of models with diminishing returns. Finally, some frameworks may synergize with early stopping by dynamically reallocating saved time to more impactful components of the pipeline (e.g., feature engineering or ensemble refinement). Please refer to Appendix \ref{cross:appendix_time_early} to see further analysis.

\section{Conclusions}\label{cross:conclusions}


Shorter time constraints are inherently interesting as they reflect real-world scenarios where rapid model deployment is critical, such as in dynamic environments or when computational resources are constrained. We showed that shorter time constraints show discriminatory power very similar to that of larger time budgets. However, these results hold for the current benchmark design and with specific versions and configurations of the evaluated AutoML frameworks, and may not generally be true. The lack of generalization is true not only for the new time constraints, but also for the old ones. Similarly to~\cite{gijsbers2024amlb}, we hope to demonstrate the degree of generalization through the evaluation of multiple time constraints and configurations, and similar to~\cite{gijsbers2024amlb} we encourage you to run your own experiments that mimic conditions that are important to you. 

The results presented in this paper suggest that using shorter time constraints than 60 minutes could be considered for AutoML benchmarking, since significant optimization is achieved in the early stages of the search process. If the goal is to achieve a fair comparison between all frameworks regarding statistical differences, ranking positions remain relatively stable even with reduced time constraints. As observed, a time budget as short as 5 minutes can still provide a reasonably consistent comparison across frameworks. On the other hand, the incorporation of early stopping mechanisms across AutoML frameworks reveals a mixed landscape where effectiveness is highly context and framework-dependent. Although early stopping can save substantial time on datasets, the risk of truncating model optimization too early remains high.

This work aims to motivate the AutoML community to reconsider excessive resource consumption in benchmarking and to foster the development of more efficient optimization techniques. Shorter time constraints and the integration of early stopping mechanisms may help reduce the duration of long-running experiments while still achieving comparable benchmarking results. This could lead to not only better optimization methods but also more sustainable practices in both research and practical applications of AutoML systems.

\subsubsection*{Acknowledgments}

We sincerely thank Nick Erickson from Amazon Web Services (AWS) for his support and contributions to this work. His insights and assistance were instrumental in shaping our research. While we were unable to include his name as a co-author due to internal AWS processing timelines, we deeply appreciate his efforts.


\bibliography{iclr2025_conference}
\bibliographystyle{iclr2025_conference}

\clearpage
\appendix
\section{Appendix \label{cross:appendix}}

\subsection{Reproducibility}


This work can be reproduced by following the next link:
\url{https://github.com/israel-cj/automlbenchmark.git}. The AutoML framework versions are also listed in the data file.

\subsection{Extra experimental setup}

AutoGluon was set up in the next way for constrained time analysis:

\begin{itemize}
    \item $\mathtt{AutoGluon(B)}$ $=$ best quality: \\presets: $\mathtt{``best\_quality"}$ 
    \item $\mathtt{AutoGluon(HQ)}$ $=$ high quality: \\ presets: $\mathtt{[``high\_quality", ``optimize\_for\_deployment"]}$
    \item $\mathtt{AutoGluon(HQIL)}$$=$ high quality with an inference time limit: \\ presets: $\mathtt{[``high\_quality", ``optimize\_for\_deployment"]}$, \\ $\mathtt{infer\_limit}$: 0.0001
    \item $\mathtt{AutoGluon(FIFTIL)}$ $=$ decent accuracy, fast inference, and extremely fast training: \\ presets: $\mathtt{[``medium\_quality", ``optimize\_for\_deployment"]}$, $\mathtt{infer\_limit}$: 0.0001 \footnote{This setup was only evaluated for early stopping.}
\end{itemize}

Not all frameworks allow the use of early stopping features. In this work, we consider five which have such capability and were activated in the following way:

\begin{itemize}
    \item $\mathtt{AutoGluon}$: it has integrated a new callback \\
    $\mathtt{EarlyStoppingEnsembleCallback(patience=5)}$
    The \verb|patience| was fixed by the developer team directly. 
    \item $\mathtt{FEDOT}$: It can be activated via $\mathtt{early\_stopping\_iterations}$, which will stop after $n$ generation without improving. We set up the value to 3 to keep it consistent with $\mathtt{H2OAutoML}$.
    \item $\mathtt{flaml}$: it uses $\mathtt{EarlyStopping}$ callback, which can be used to monitor a metric and stop the training when no improvement is observed. This can be activated with the parameter $\mathtt{early\_stop}$ argument.
    \item $\mathtt{H2OAutoML}$: to turn on early stopping a number $>=1$ in $\mathtt{stopping\_rounds}$ should be defined. $\mathtt{stopping\_rounds}$ is used to stop model training when the metric does not improve for this specified number of training rounds based on a simple moving average. It was set up to the default value of 3.
    \item $\mathtt{TPOT}$: it uses the parameter $\mathtt{early\_stop}$, which is an integer, optional ($\mathtt{default: None}$). This represents how many generations TPOT checks whether there is no improvement in the optimization process. We set up the value to 3 to keep it consistent with $\mathtt{H2OAutoML}$.
    
\end{itemize}

\subsection{Behaviour over time constraints extended}

\label{cross:relativeImprovement} 
\begin{figure}[h]
\begin{center}
\begin{subfigure}{0.45\textwidth}
\includegraphics[width=\linewidth]{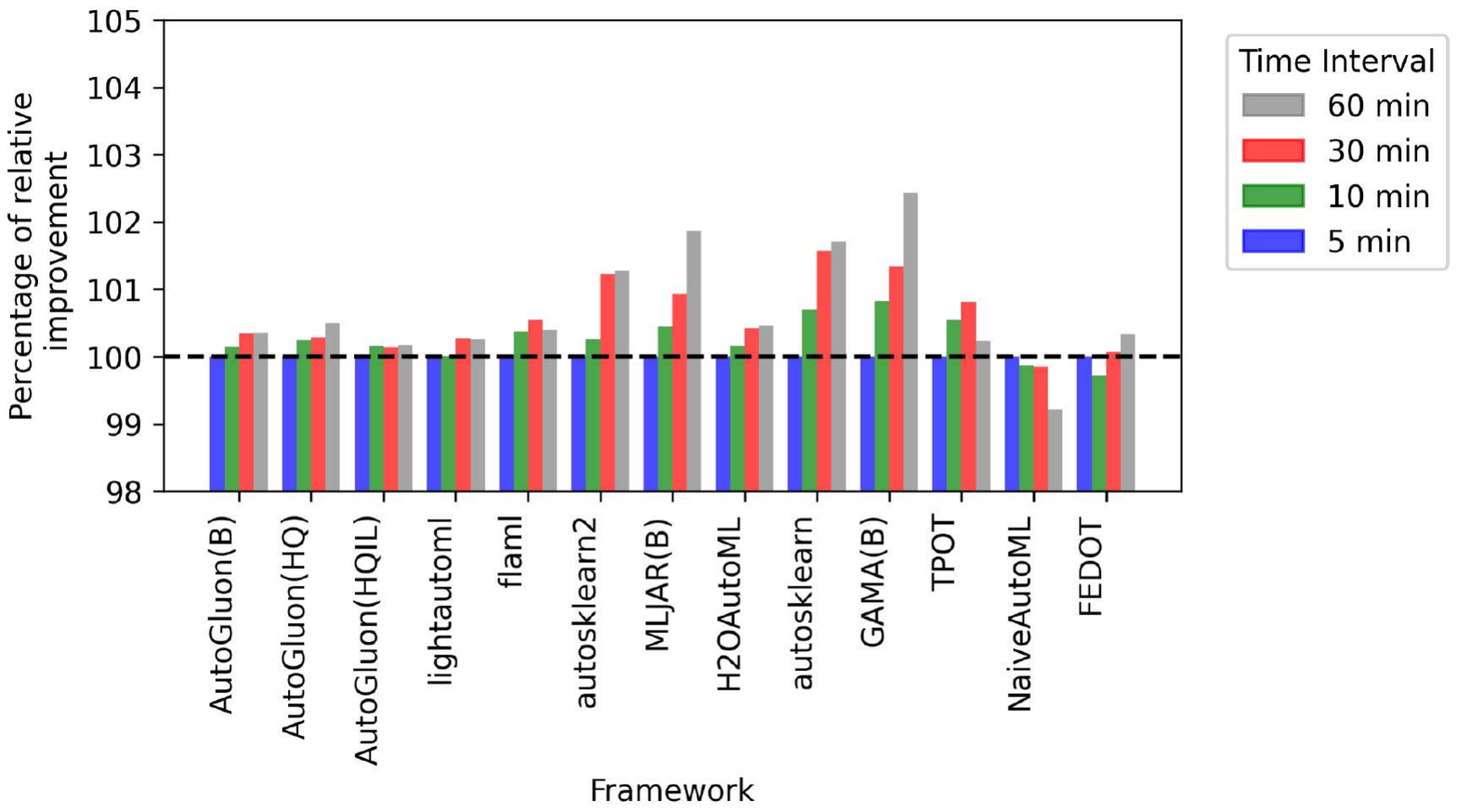}
\caption{Mean performance defined in percentage per each time constraint, the missed values were imputed across all time constraints}
\end{subfigure}
\hspace{0.02\textwidth} 
\begin{subfigure}{0.45\textwidth}
\includegraphics[width=\linewidth]{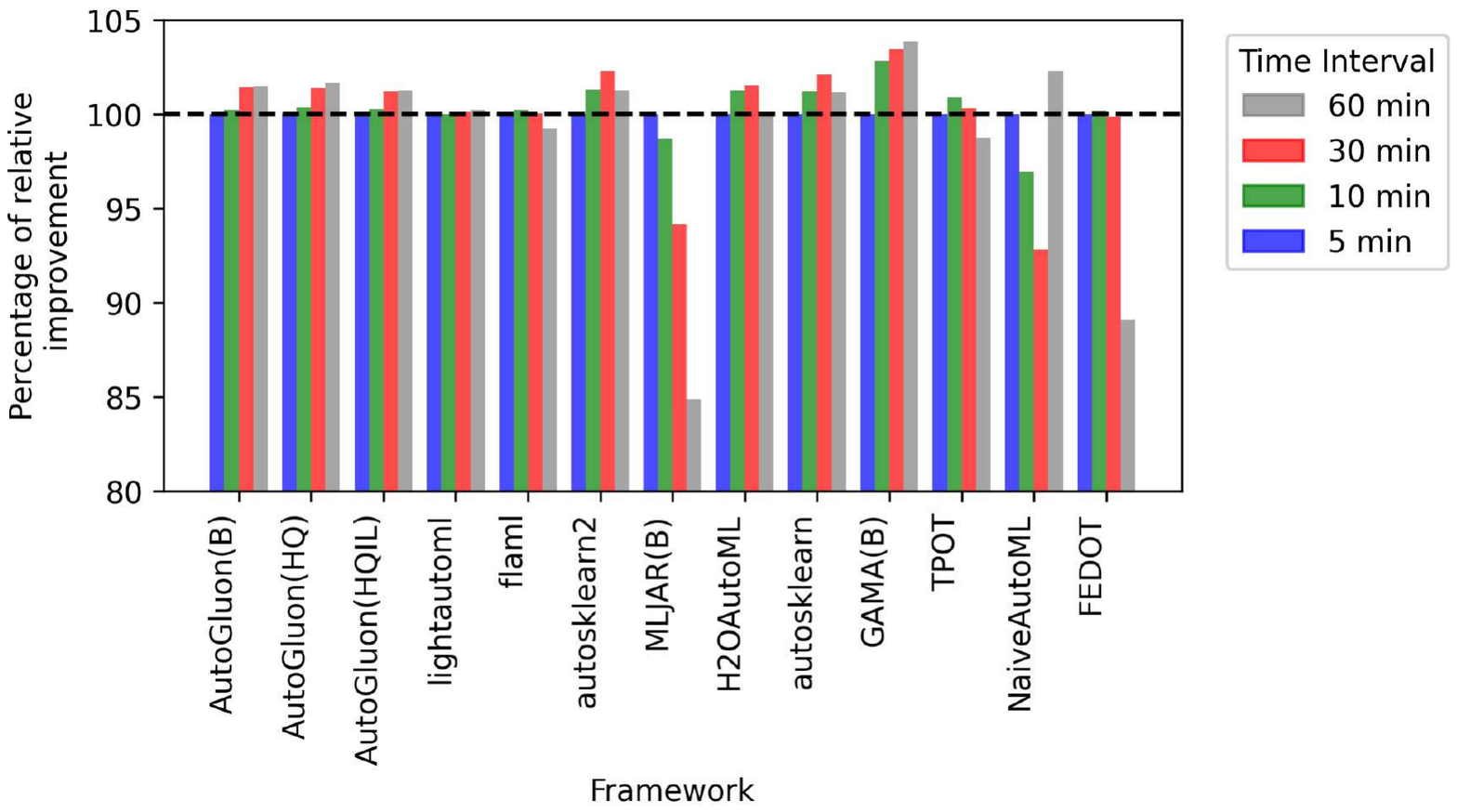}
\caption{Mean performance defined in percentage per each time constraint, missing values imputed with $\mathtt{CP}$}
\end{subfigure}

\caption{Relative improved performance reached in each time constraint when considering 5 minutes as 100\% over binary tasks. \label{fig:improvedpercentage}}
\end{center}
\end{figure}

To exemplify to which extent the frameworks have improved over time, Figure \ref{fig:improvedpercentage} illustrates the relative improvement after 5 minutes per framework on binary tasks (area under the curve is the metric used, thus is straightforward to interpret). The Figure is presented in two scenarios: (a) missed values across times were just imputed to keep a homogeneous number of tasks per framework, or (b) missing tasks were replaced with $\mathtt{CP}$ (producing an effect where the more errors in the framework, the worse the results will turn in most cases). It is appreciated that most optimization actually happens in the first 5 minutes for binary datasets. In fact, the highest a framework improved relatively after 5 minutes is close to 3\% ($\mathtt{GAMA}$). One possibility is that AutoML frameworks are designed to quickly optimize core hyperparameters early on, leveraging efficient search strategies that focus on the most promising regions of the search space. The diminishing returns over longer time intervals might suggest that the initial optimizations capture much of the predictive power, with further fine-tuning yielding marginal gains. As we mentioned, this can also be related to the nature of the datasets used since only 14 tasks have a high dimensionality ($>1k$ features), and 9 are large ($>500k$ instances); this allows frameworks to explore the search space more thoroughly in shorter budgets.

\subsection{Ranking correlation with meta-features}

\begin{figure}[h]
\begin{center}

\begin{subfigure}{0.90\textwidth}
\includegraphics[width=\linewidth]{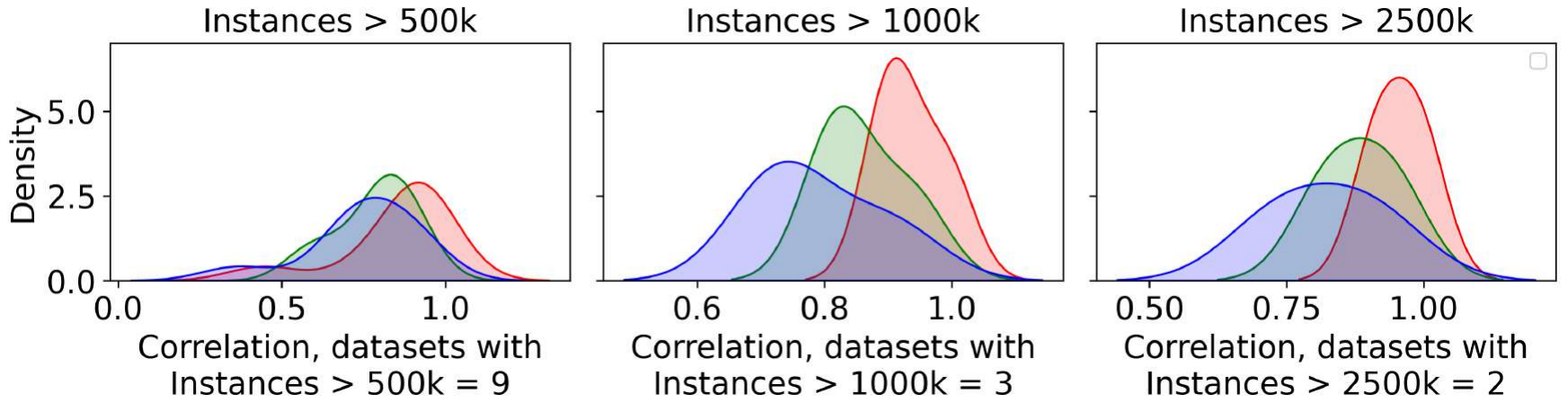}
\caption{Ranking correlation vs. Instances}
\end{subfigure}
\begin{subfigure}{0.90\textwidth}
\includegraphics[width=\linewidth]{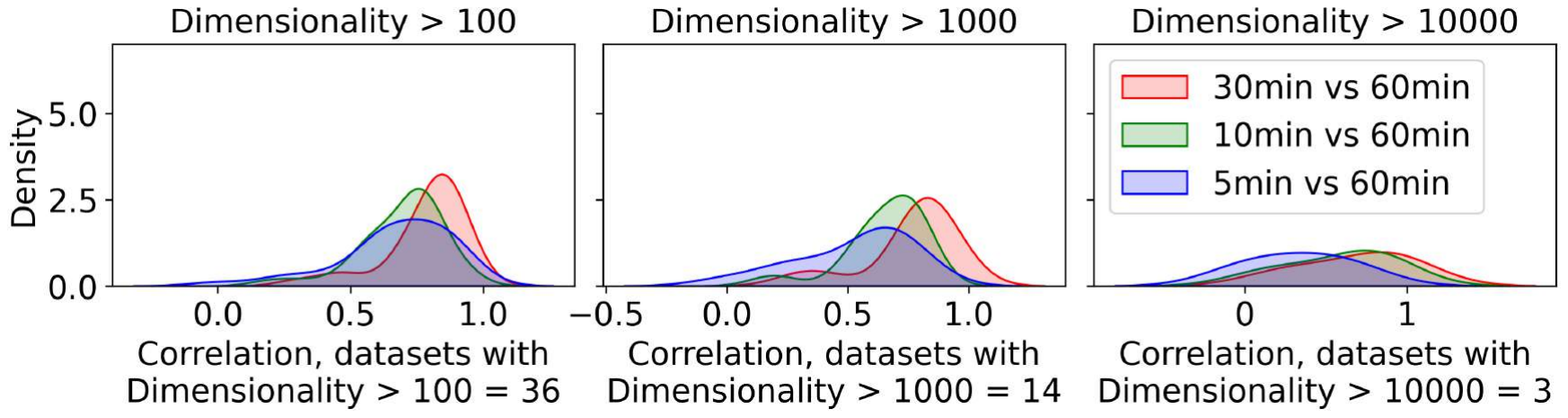}
\caption{Ranking correlation vs. Features}
\end{subfigure}

\end{center}
\caption{Density plot of the correlation from the ranking of the means of each framework by each task at each time, divided into instances and features number \label{fig:densityplotsinstances}}
\end{figure}

We perform an analysis at the meta-features level. Figure \ref{fig:densityplotsinstances} (a) reveals how the number of instances in a dataset affects the ranking correlation between AutoML performance across different time constraints. For datasets with over 500k instances, the 5 vs 60-minute comparison shows broader variation, with correlations spread between 0.6 and 1.0. For datasets with over 1 million instances, correlations generally improve, with the 30 vs 60-minute comparison peaking close to 1.0, indicating a highly comparable ranking position per dataset. The same applies to datasets over 2.5 million instances; 30 minutes might be a viable alternative for large-scale datasets. Nevertheless, the limited number of datasets in this category (only 2 datasets) suggests that while these observations are interesting, this effect may not generalize across all very large datasets.

Related to dimensionality,
Figure \ref{fig:densityplotsinstances} (b) illustrates the ranking correlation vs dataset features. For datasets with more than 100 features, the distributions are concentrated near higher correlation values. However, as dimensionality exceeds 1k, the distributions widen, particularly for the 5 vs 60-minute comparison, where there is a shift below 0.5, indicating a growing divergence in ranking. The early time limits may not be sufficient for all the tasks to effectively optimize models for high-dimensional data, hence the sharp decline in correlation for larger datasets. 
For datasets with more than 10,000 features, the ranking gap becomes even more pronounced, as shown by the wide spread of the correlation distribution. Likely due to the increased error, the larger datasets get, the harder it is to finish anything within the time constraints. Evaluating all possible interactions and combinations becomes computationally intensive, requiring more time for both model training and hyperparameter optimization.

\begin{figure}[h]
\centering

\begin{subfigure}{0.90\textwidth}
\includegraphics[width=\linewidth]{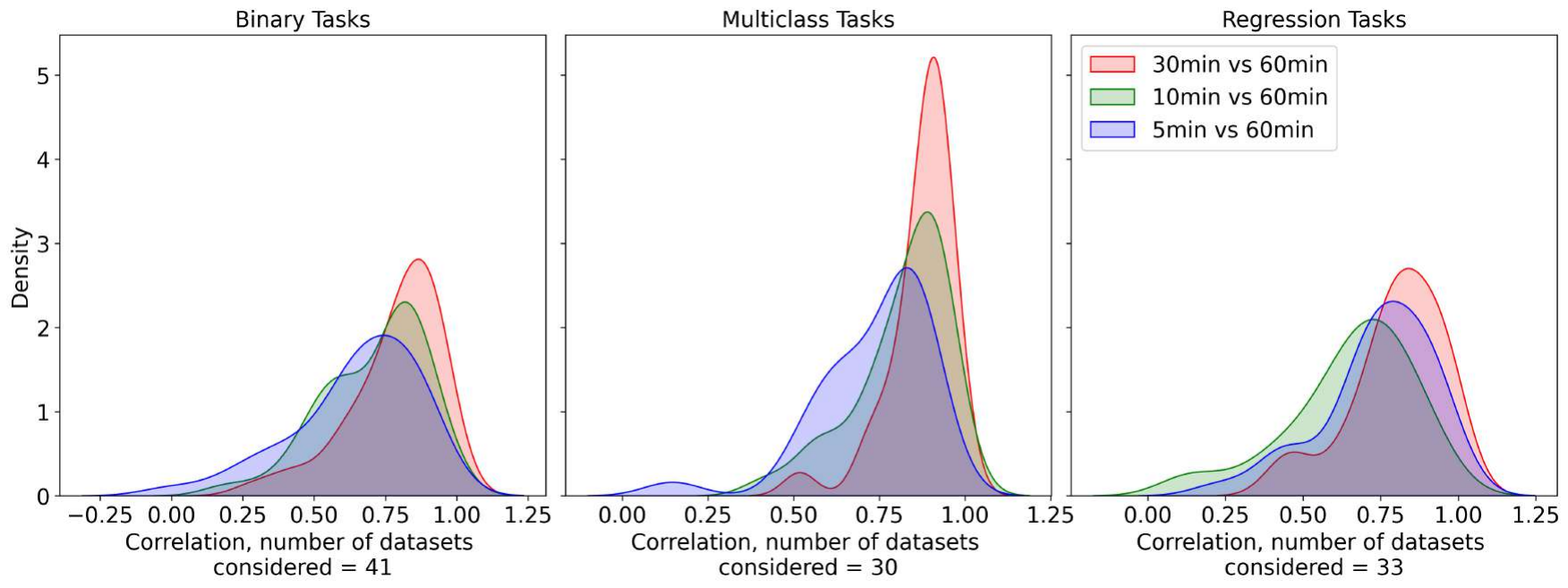}
\caption{}
\end{subfigure}
\newline
\begin{subfigure}{0.80\textwidth}
\includegraphics[width=\linewidth]{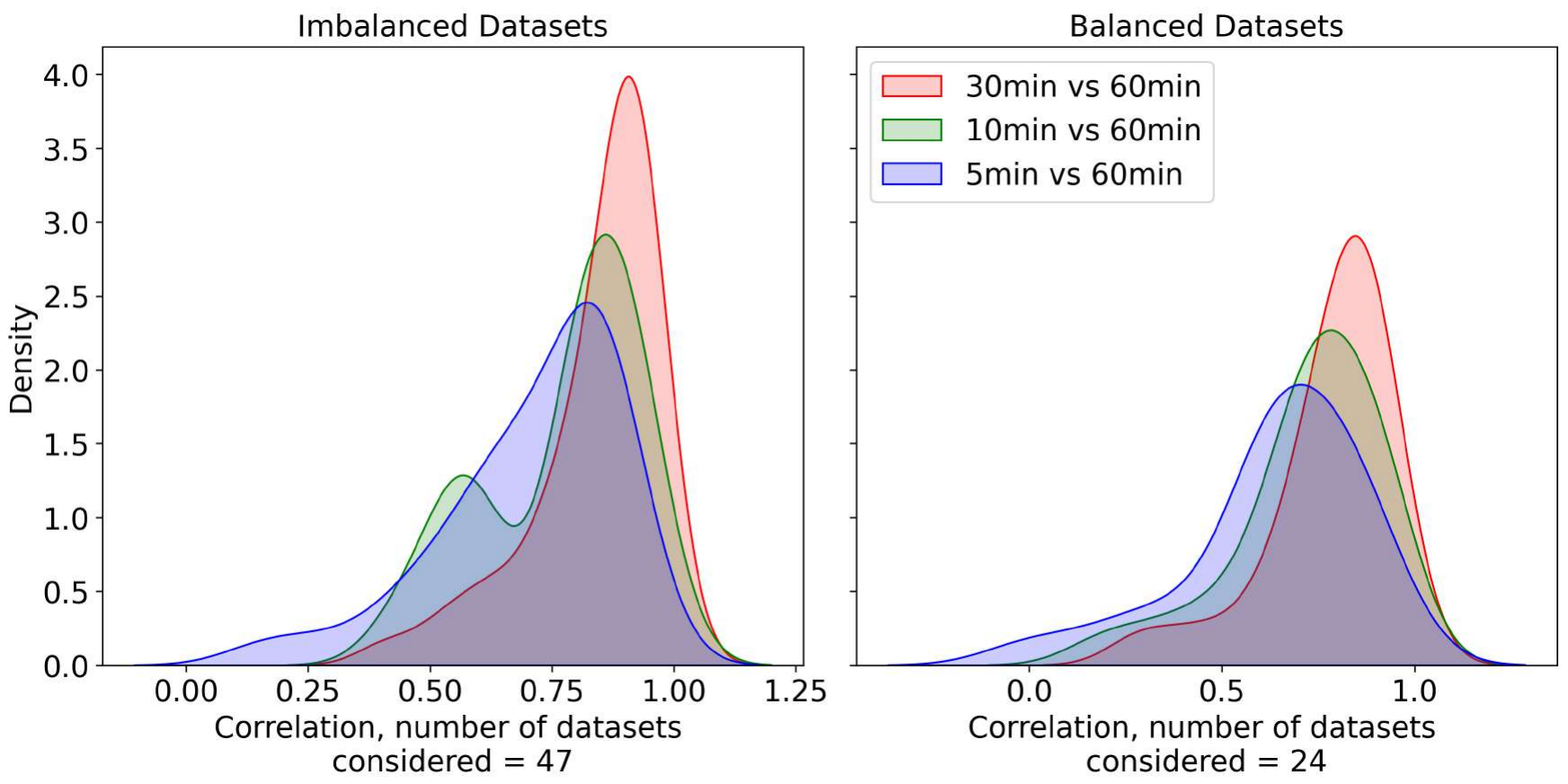}
\caption{}
\end{subfigure}

\caption{(a) Density plot of the correlation from the ranking of the means of each framework per type (b) Density plot of the correlation from the ranking of the means of each framework per imbalance ratio (classification tasks) \label{fig:rankingpositions_extra}}
\end{figure}

Similarly, Figure \ref{fig:rankingpositions_extra} (a), we evaluated the ranking of each framework per type (binary, multiclass and regression). Binary tasks in shorter time budgets (5 and 10 minutes) exhibit a broader spread of correlation values compared to 30 minutes vs 60 minutes. The 30-minute curve is narrower, peaking closer to 1, indicating that performance is more consistent with the 60-minute benchmark. A clearer trend is observed for multiclass tasks, with the correlation improving as the time budget increases. The 30-minute curve has the highest peak and is closer to 1, showing a stronger alignment with the 60-minute performance.
Regression Tasks: There is a noticeable broader distribution for 10 minutes, indicating greater variability. As the time budget increases, particularly with 30 minutes, the correlation with the 60-minute performance improves again, peaking around 0.8 to 1.

In Figure \ref{fig:rankingpositions_extra} (b) it was analyzed the ranking performance vs. imbalance ratio (defined by a threshold of 1.5  $MajorityClassPercentage / MinorityClassPercentage$). For imbalanced datasets, short time constraints (5 and 10 minutes) lead to more variation in performance. The 30-minute time budget shows a tighter distribution, with a higher peak closer to 1, indicating better alignment with the 60-minute performance.
In balanced datasets, there is less variation across all time constraints. The 30-minute correlation curve has a sharp peak near 1, showing that for balanced datasets, the performance between 30 and 60 minutes is highly consistent. The 5-minute and 10-minute comparisons are more spread out but still perform relatively well, as their peaks are closer to higher correlation values.

To analyze the impact of task types, Table \ref{table:ranking_correlation_task} presents the correlation between meta-features and ranking correlation across different time constraints. For classification tasks, the weak correlations with meta-features suggest that these features have little influence on the consistency of framework rankings over time. A similar pattern is observed for regression tasks, indicating that meta-features alone cannot fully account for the variations in ranking. This suggests that more complex interactions between data characteristics and time limitations likely play a role. Contributing factors could include algorithm sensitivity, the curse of dimensionality, and the sensitivity of evaluation metrics, among others.

\begin{table}[h]
\centering
\caption{Ranking correlation per task }
\label{table:ranking_correlation_task}
\scalebox{0.9}{
\begin{tabular}{l|llllllll}
\hline
                        & \textbf{Features} & $p$-value & \textbf{Instances} & $p$-value & \textbf{Ratio} & $p$-value & \textbf{Classes} & $p$-value \\ \hline
\textbf{Classification} & -0.239            & $<$0.001  & 0.075              & 0.275     & -0.115         & 0.09      & -0.014           & 0.82      \\ \hline
\textbf{Regression}     & -0.062            & 0.538     & 0.21               & 0.036     &                &           &                  &           \\ \hline
\end{tabular}
}
\end{table}

Since the meta-features do not allow us to observe a clear trend in the performance of the AutoML systems over time to determine the main discrepancies, we performed a task-level analysis over datasets with correlation in the rankings ($\leq 0.6$). There is an intersection of 7 tasks across the 3-time constraint \textit{Moneyball, sylvine, dna, tecator, Internet-Advertisements, arcene} and \textit{Australian}. What do these tasks have in common?  \textit{Internet-Advertisements} has a 0.16 imbalance ratio (minority class/majority class), and \textit{arcene} contains 10k features. Tasks like \textit{dna} and \textit{sylvine} (a challenge in machine learning) often involve complex relationships between features and target variables. \textit{Moneyball} contains 1118 out of 1232 instances with missing values. \textit{tecator} has a column with 216 distinct values out of 240, which makes it too little to predict with high accuracy, same for \textit{Australian} containing up to 350 different values out of 690. AutoML systems usually can manage complex datasets. However, they still might struggle, this can often depend on how well the inherent strengths of AutoML align with the specific requirements of a given dataset, including complex non-linear interactions between features/instances/targets. 

\subsection{Time constraint evaluations}

\subsubsection{Performance differences}

\begin{figure}[h]
\begin{center}

\begin{subfigure}{0.31\textwidth}
\includegraphics[width=\linewidth]{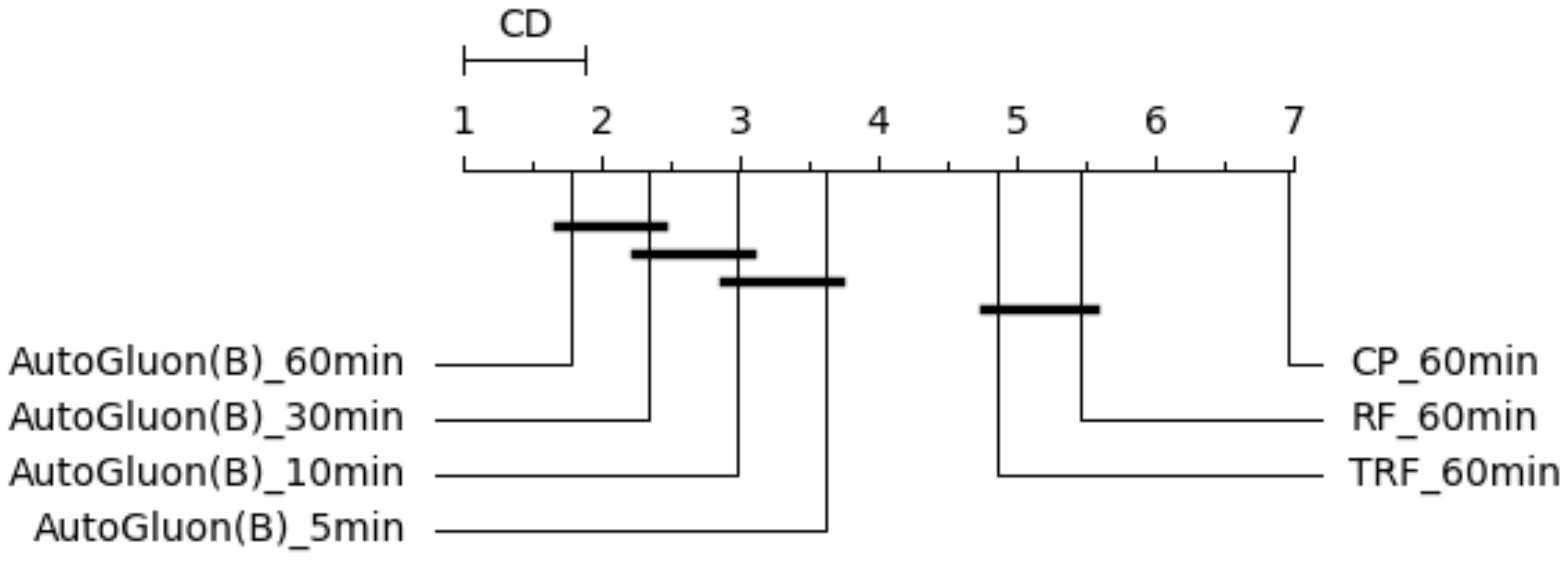}
\caption{AutoGluon(B)}
\end{subfigure}
\begin{subfigure}{0.31\textwidth}
\includegraphics[width=\linewidth]{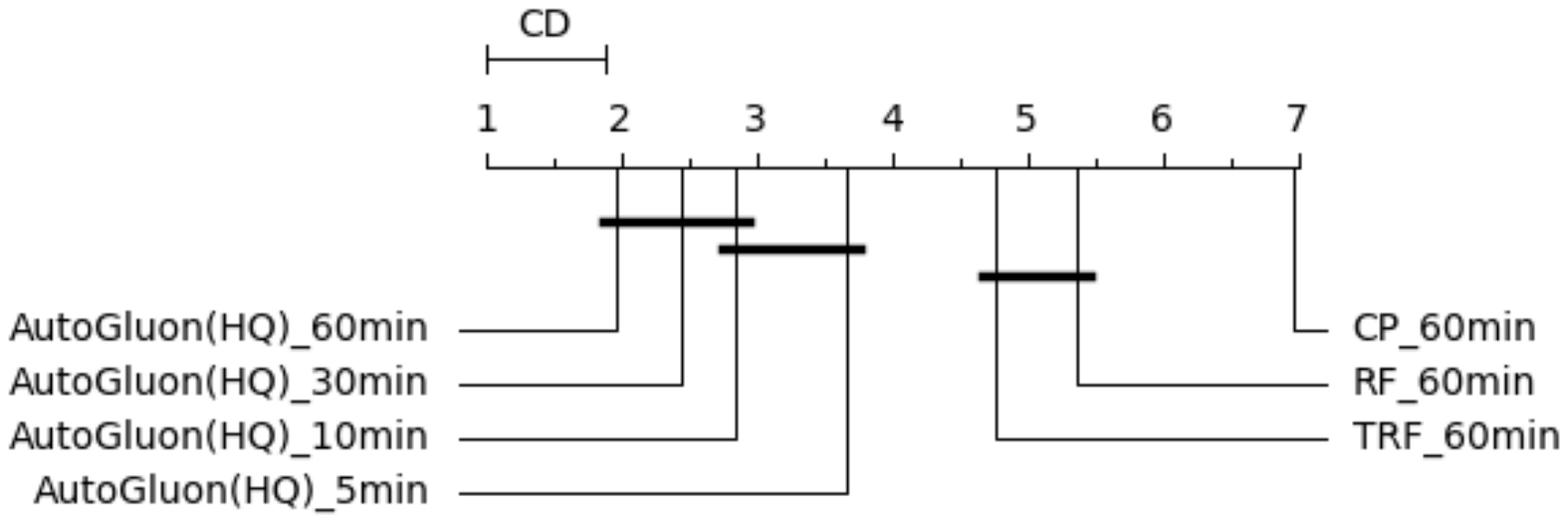}
\caption{AutoGluon(HQ)}
\end{subfigure}
\begin{subfigure}{0.31\textwidth}
\includegraphics[width=\linewidth]{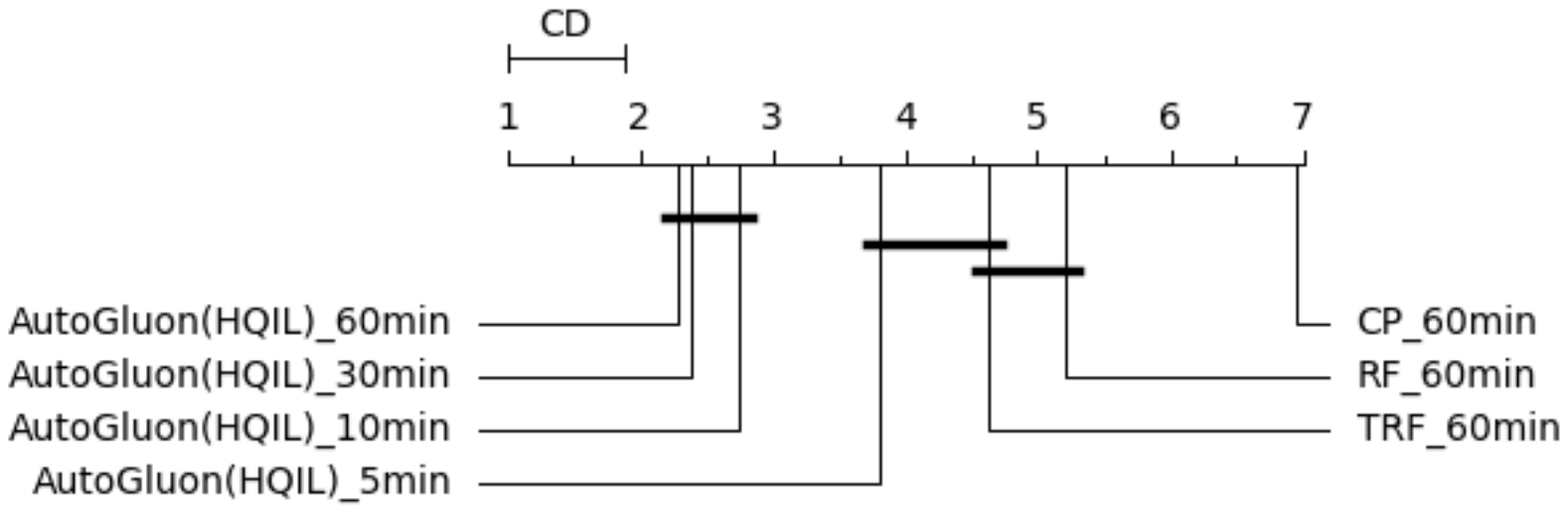}
\caption{AutoGluon(HQIL)}
\end{subfigure}
\newline
\begin{subfigure}{0.31\textwidth}
\includegraphics[width=\linewidth]{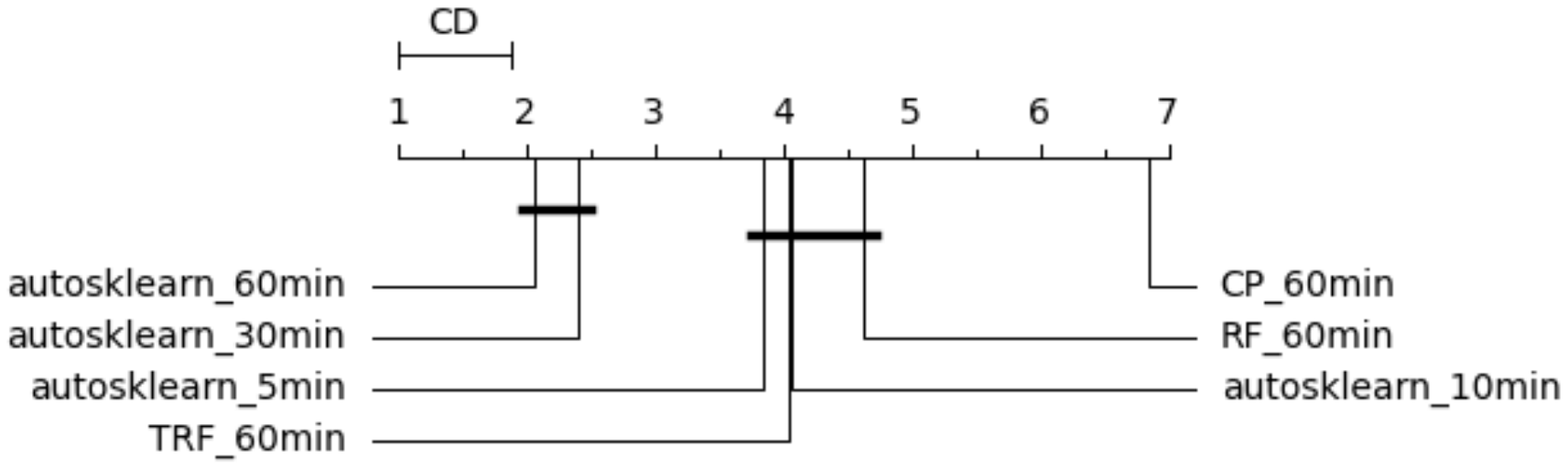}
\caption{autosklearn}
\end{subfigure}
\begin{subfigure}{0.31\textwidth}
\includegraphics[width=\linewidth]{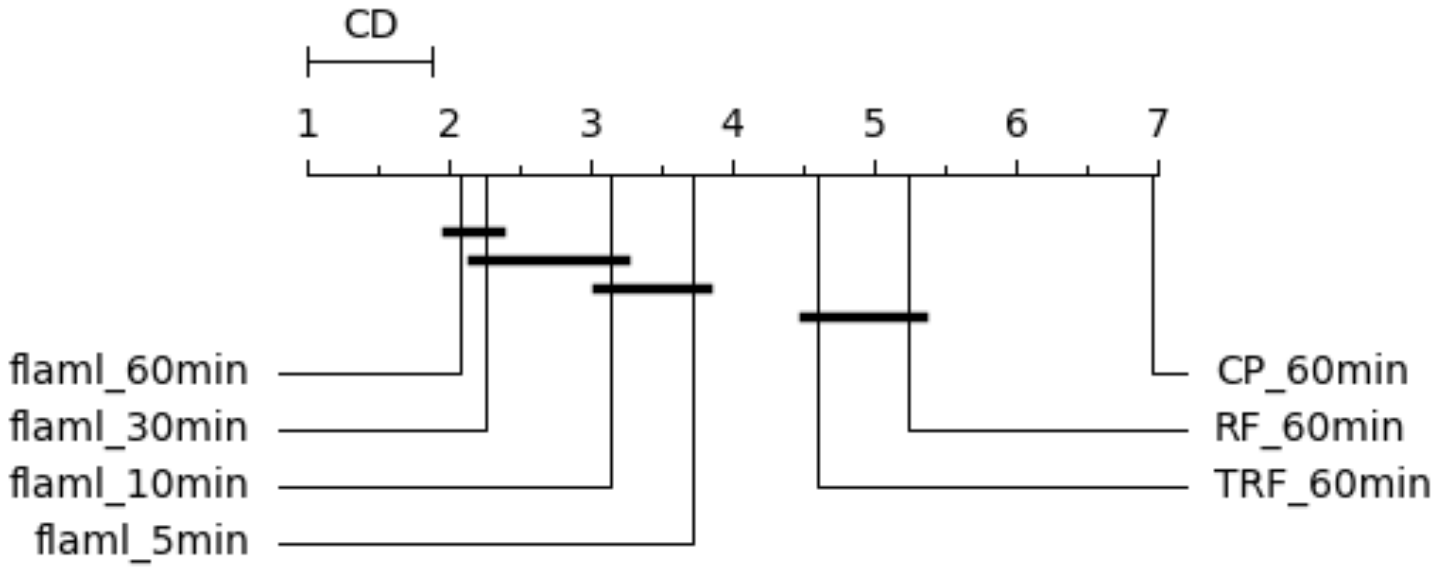}
\caption{flaml}
\end{subfigure}
\begin{subfigure}{0.31\textwidth}
\includegraphics[width=\linewidth]{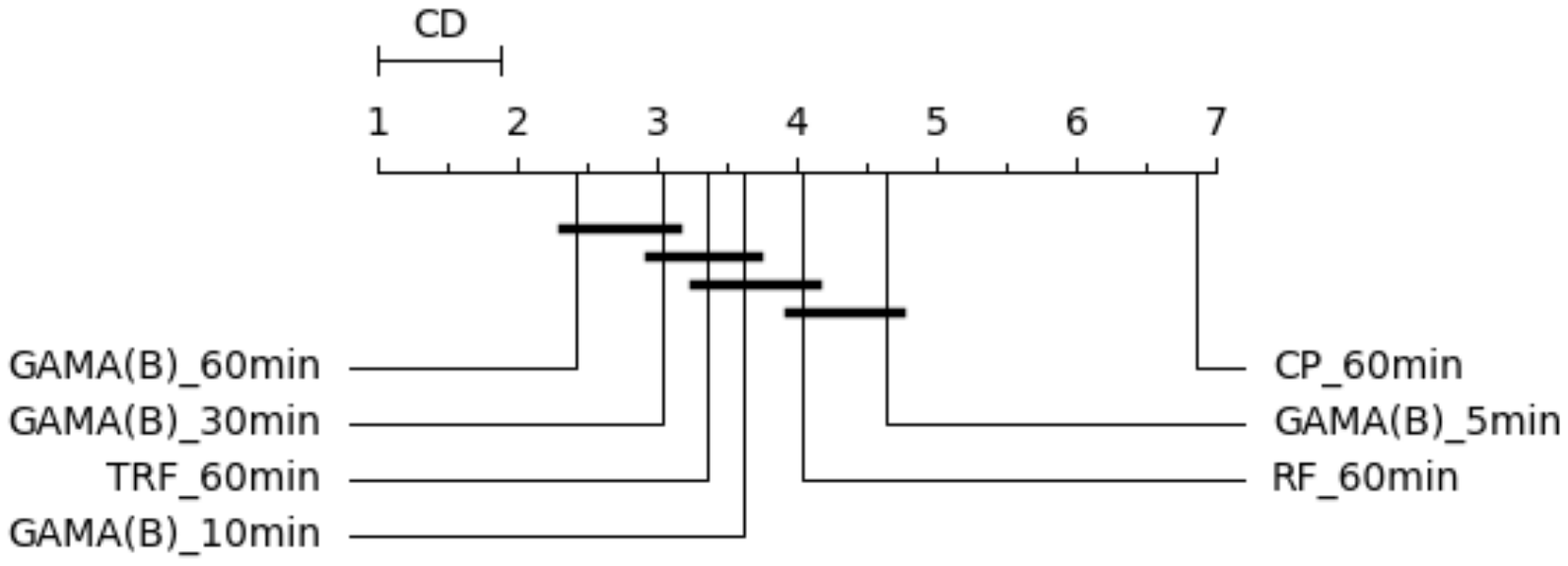}
\caption{GAMA}
\end{subfigure}
\newline
\begin{subfigure}{0.31\textwidth}
\includegraphics[width=\linewidth]{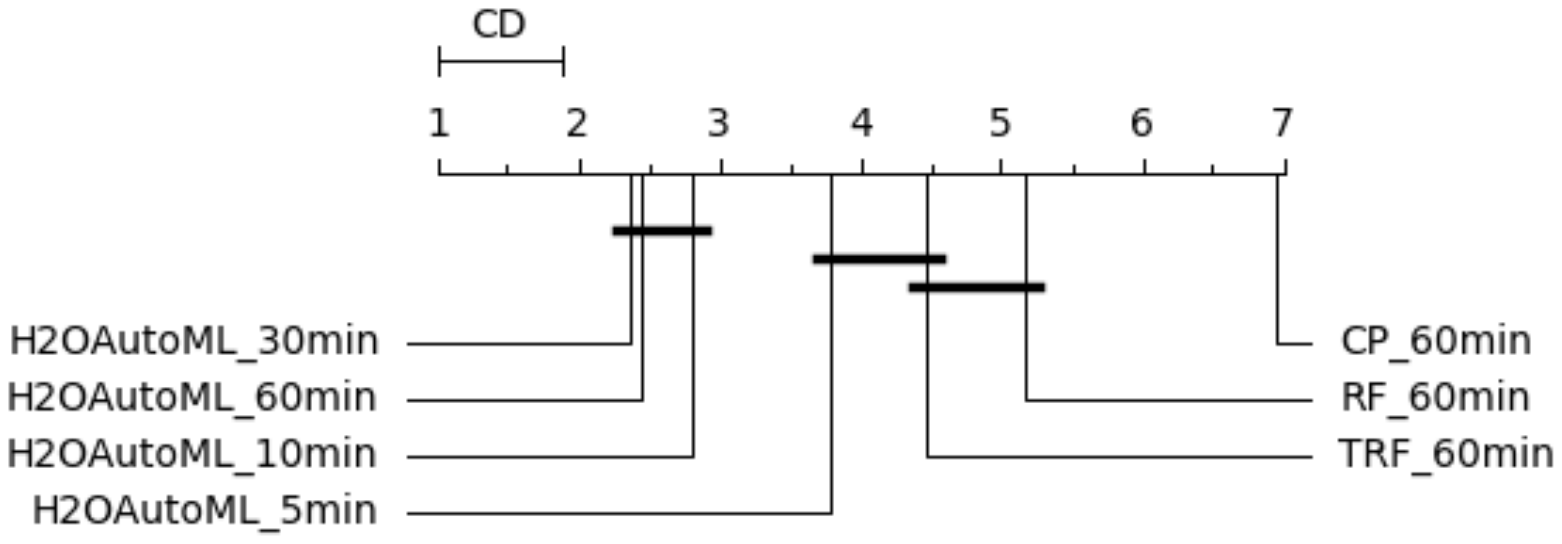}
\caption{H2OAutoML}
\end{subfigure}
\begin{subfigure}{0.31\textwidth}
\includegraphics[width=\linewidth]{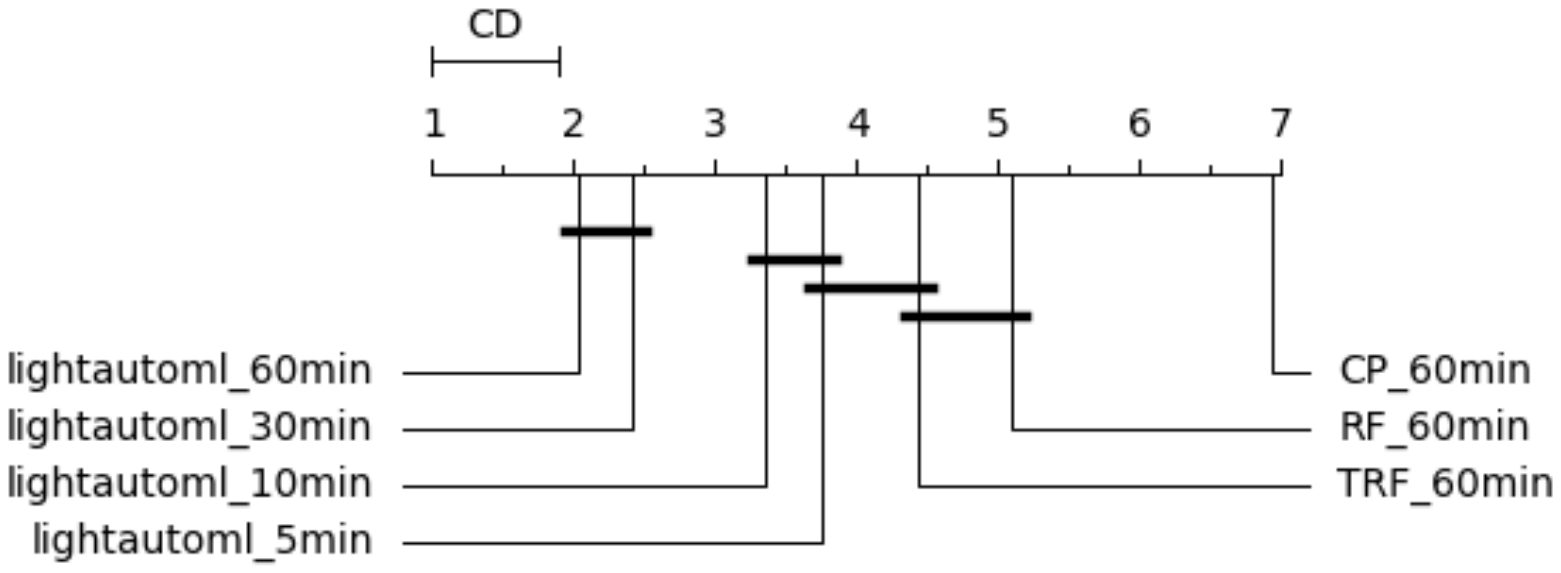}
\caption{lightautoml}
\end{subfigure}
\begin{subfigure}{0.31\textwidth}
\includegraphics[width=\linewidth]{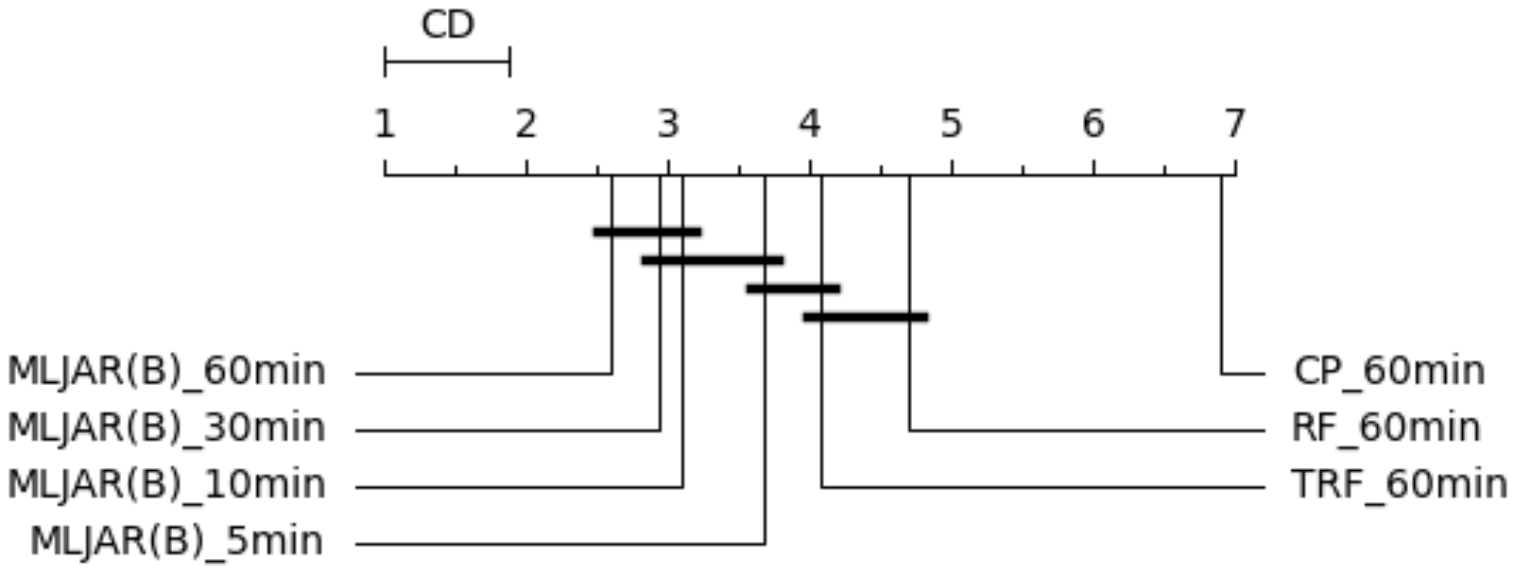}
\caption{MLJAR(B)}
\end{subfigure}
\newline
\begin{subfigure}{0.31\textwidth}
\includegraphics[width=\linewidth]{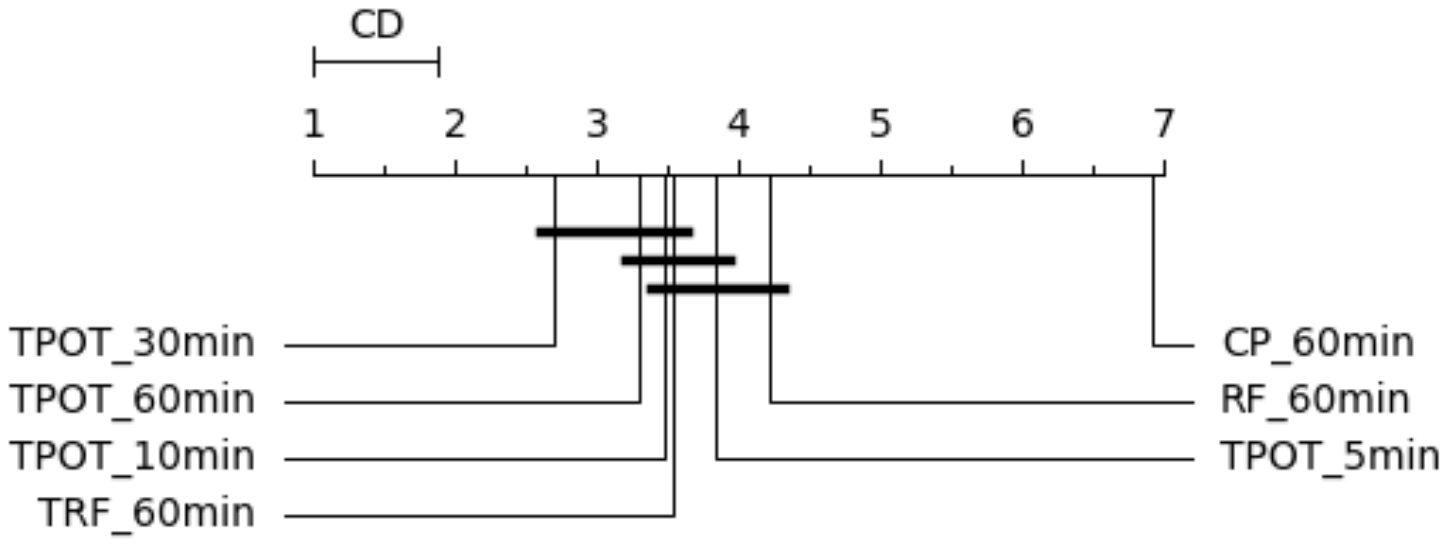}
\caption{TPOT}
\end{subfigure}
\newline

\end{center}
\caption{Critical Diagrams (CD) of the evaluated frameworks by time constraints. These diagrams include the Nemenyi post-hoc test. \label{fig:appendix_critical_diagrams_by_time}}
\end{figure}

Figure \ref{fig:appendix_critical_diagrams_by_time} presents the CD diagrams for each evaluated AutoML framework under varying time constraints. In general, most frameworks exhibit a consistent performance across the time budgets, meaning no decrement in performance behaviour as the time limit is extended. However, exceptions such as $\mathtt{TPOT}$ stand out. This is primarily due to missing values being imputed using $\mathtt{CP}$. Frameworks that encountered multiple errors tended to gravitate toward $\mathtt{CP}$ in the ranking. Several frameworks remain competitive even with reduced time constraints, showing minimal performance differences between the 30-minute and 60-minute evaluations. For instance, frameworks like $\mathtt{H2OAutoML}$, $\mathtt{AutoGluon}$, $\mathtt{flaml}$ and $\mathtt{lightautoml}$ demonstrate relatively stable rankings across different time budgets, indicating they can achieve near-optimal performance within 30 minutes. 

\subsubsection{Performance distributions}

\begin{figure}[h]
\begin{center}

\begin{subfigure}{0.45\textwidth}
\includegraphics[width=\linewidth]{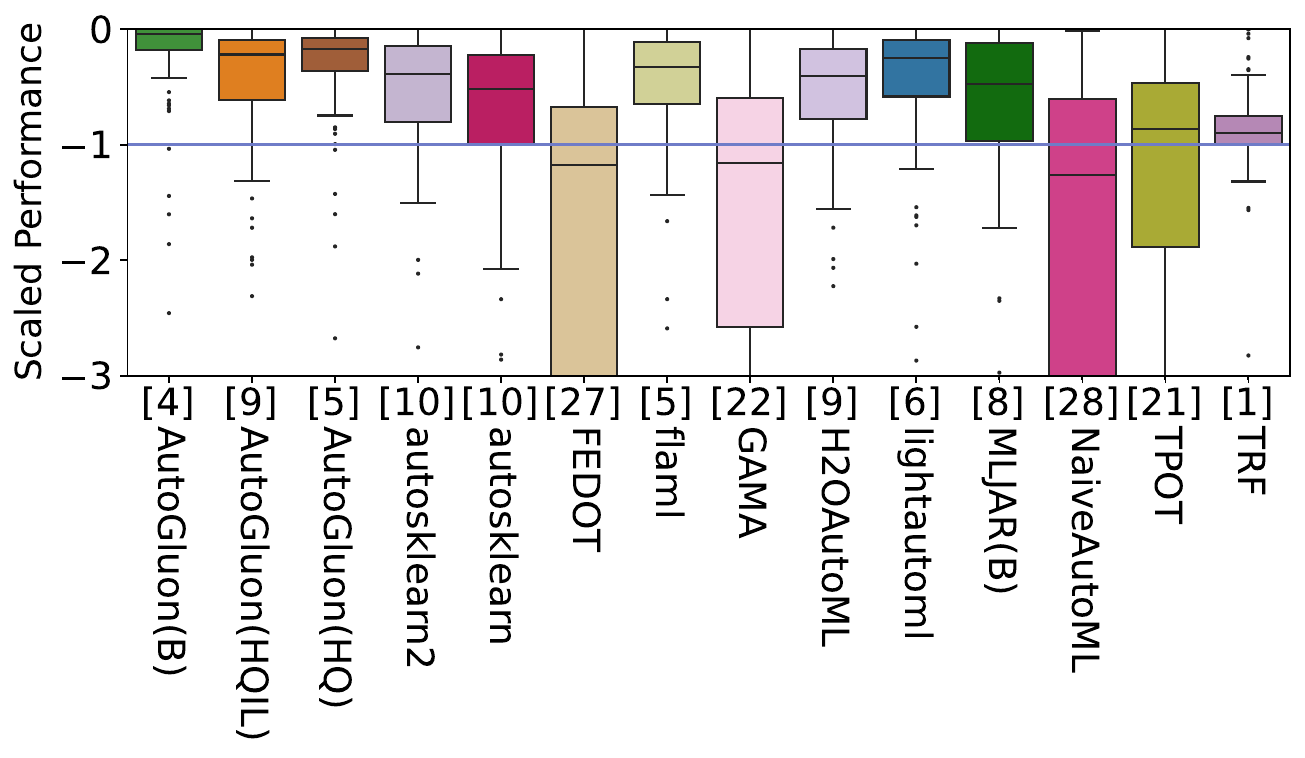}
\caption{\tiny 5 minutes evaluation}
\end{subfigure}
\begin{subfigure}{0.45\textwidth}
\includegraphics[width=\linewidth]{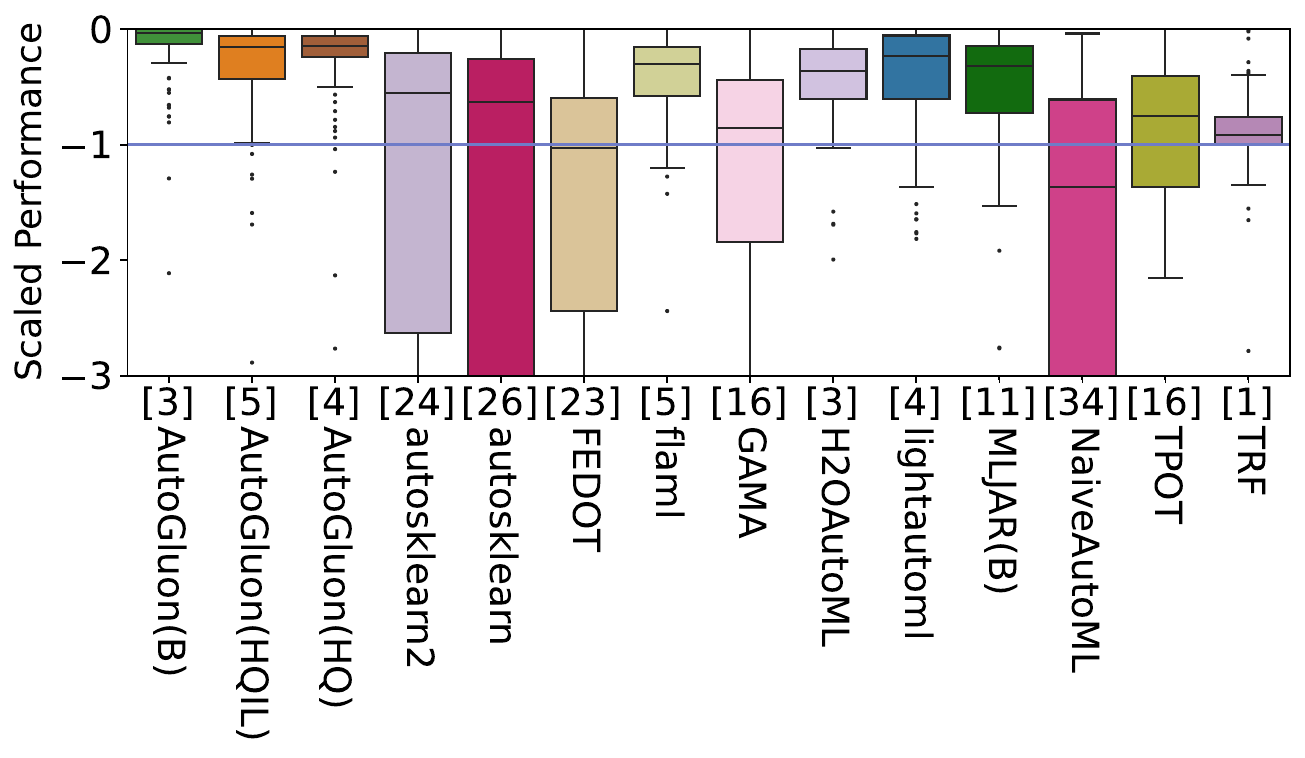}
\caption{\tiny 10 minutes evaluation}
\end{subfigure}
\begin{subfigure}{0.45\textwidth}
\includegraphics[width=\linewidth]{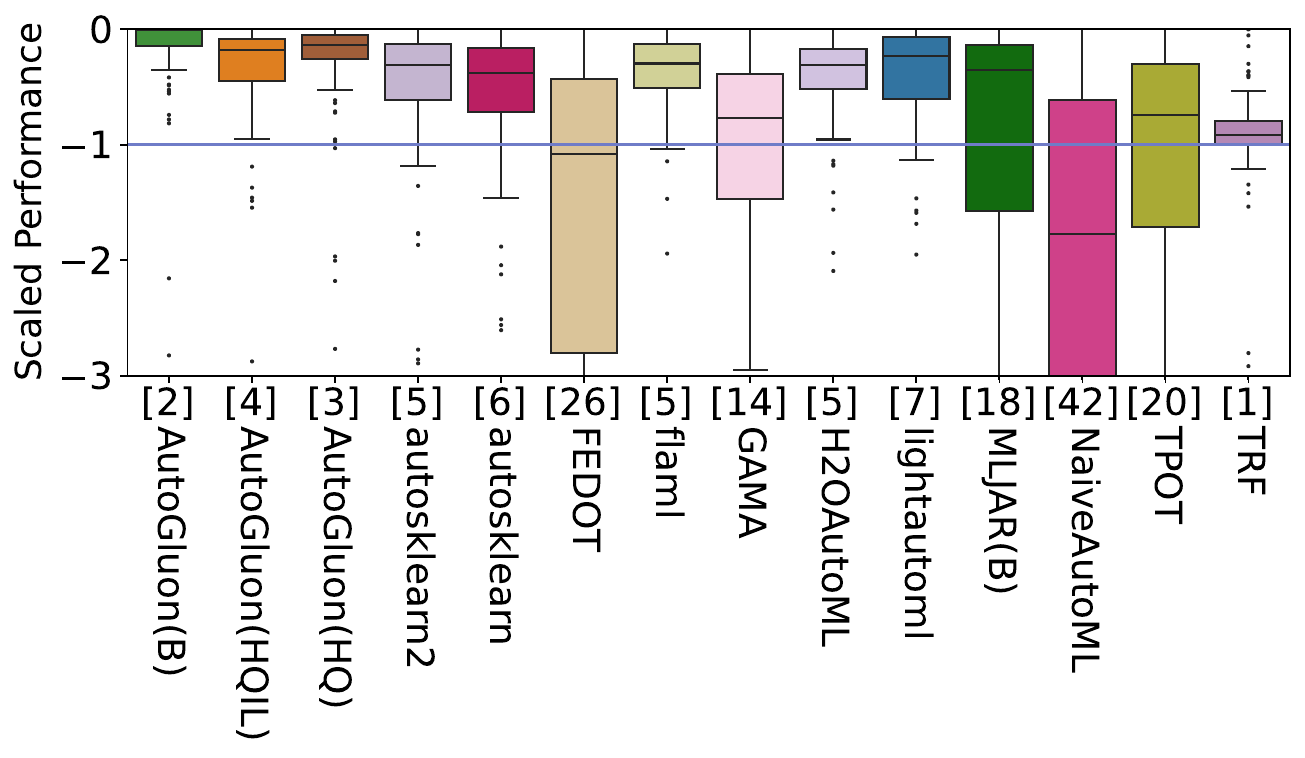}
\caption{\tiny 30 minutes evaluation}
\end{subfigure}
\begin{subfigure}{0.45\textwidth}
\includegraphics[width=\linewidth]{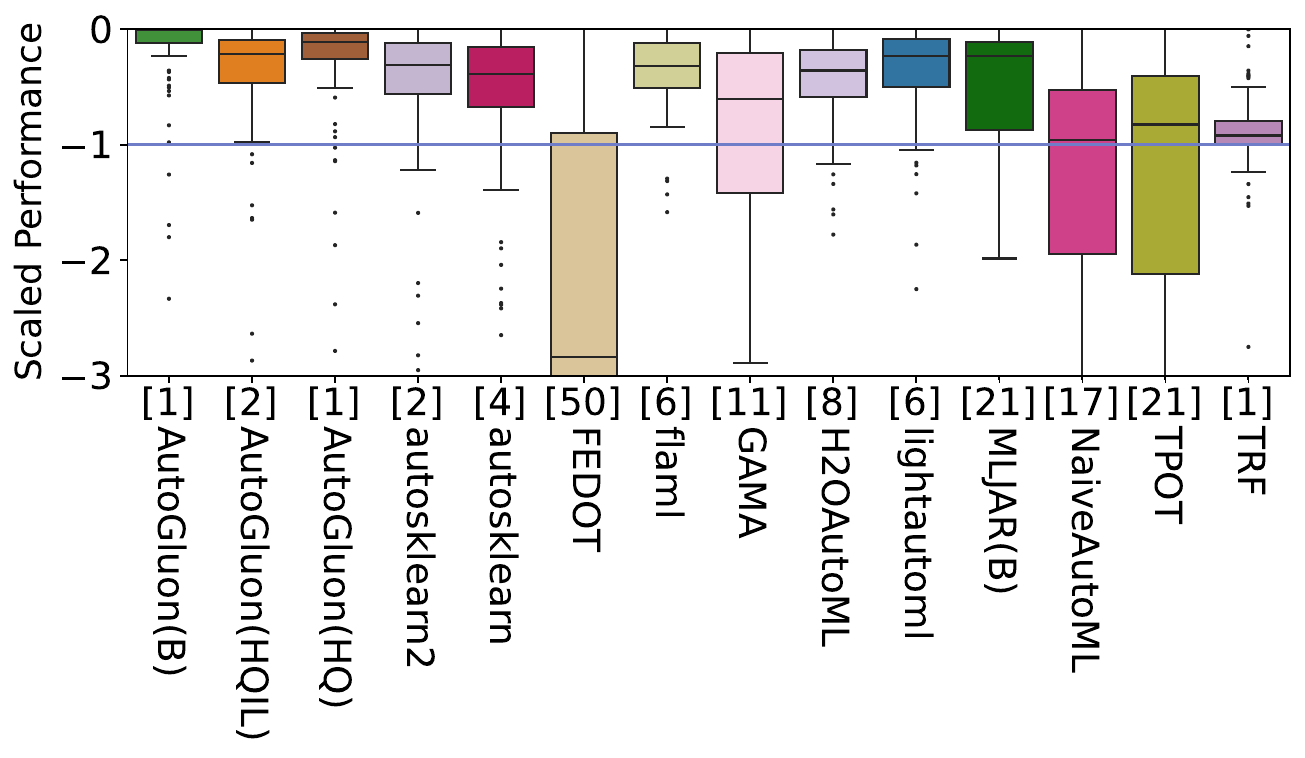}
\caption{\tiny 60 minutes evaluation}
\end{subfigure}

\end{center}

\caption{Boxplots of framework results evaluated by time constraints, distribution after scaling the performance values from the random forest $\mathtt{RF}$ 60 minutes (-1 horizontal line) to best observed (0). In addition, all the frameworks are compared against $\mathtt{TRF}$. The number of outliers for each framework that are not shown in the plot is denoted on the x-axis. \label{fig:boxplot}}
\end{figure}

Figure \ref{fig:boxplot} illustrates the box plots of the performance of the frameworks, such are scaled from the baseline set by a random forest model with 60 minutes $\mathtt{RF}$ (-1) to the best-observed performance (0) as done in the original AMLB paper, in addition, all the frameworks are compared against $\mathtt{TRF}$. Starting with the 5-minute evaluation (a), we see a significant spread in the performance distributions for several frameworks. Notably, frameworks like $\mathtt{AutoGluon}$ and $\mathtt{flaml}$ show the best performance distributions across the tasks. In the 10-minute evaluation (b), the distribution remains steady for most of the frameworks, and the mean improved for frameworks like $\mathtt{GAMA}$ and $\mathtt{TPOT}$, indicating that even a small increase in time budget allows them to explore more effective configurations but the spread is wider for $\mathtt{autosklearn}$ and $\mathtt{autosklearn2}$, intuition is due to the fact that its performance does not importantly change from 5 to 10 minutes so the lower quartile is dragged down when normalized. By the 30-minute (c), we observe a further improvement across the frameworks. $\mathtt{AutoGluon}$, $\mathtt{autosklearn}$, $\mathtt{flaml}$, $\mathtt{lightAutoML}$ and $\mathtt{H2OAutoML}$ continue to maintain strong performance. Finally, in the 60-minute evaluation (d), nearly all frameworks show tightly grouped distributions. The exceptions over all the times with the weakest performances are $\mathtt{NaiveAutoML}$, $\mathtt{TPOT}$ and $\mathtt{FEDOT}$.

\begin{figure}[h]
\begin{center}

\begin{subfigure}{0.31\textwidth}
\includegraphics[width=\linewidth]{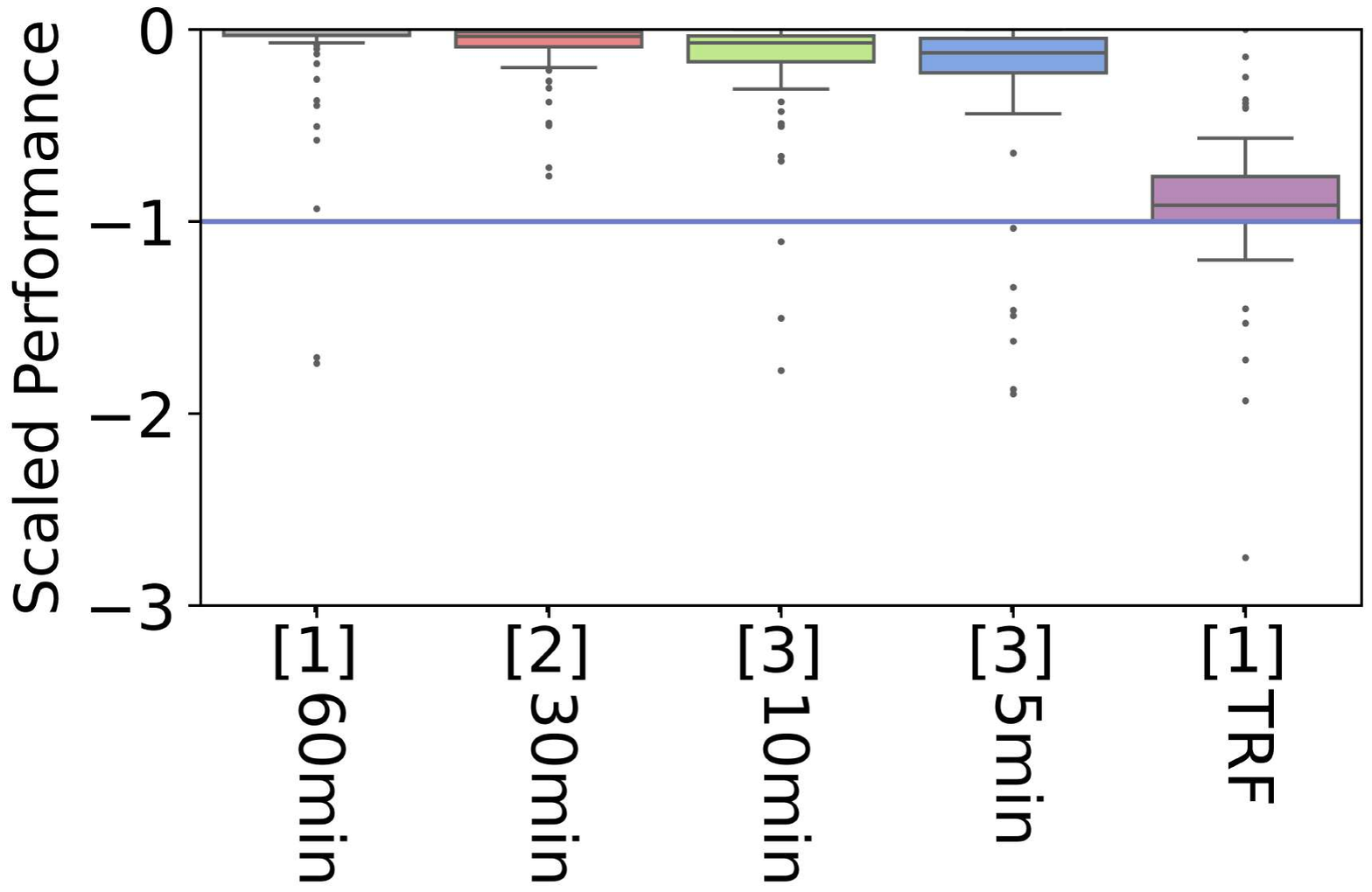}
\caption{\tiny AutoGluon(B)}
\end{subfigure}
\begin{subfigure}{0.31\textwidth}
\includegraphics[width=\linewidth]{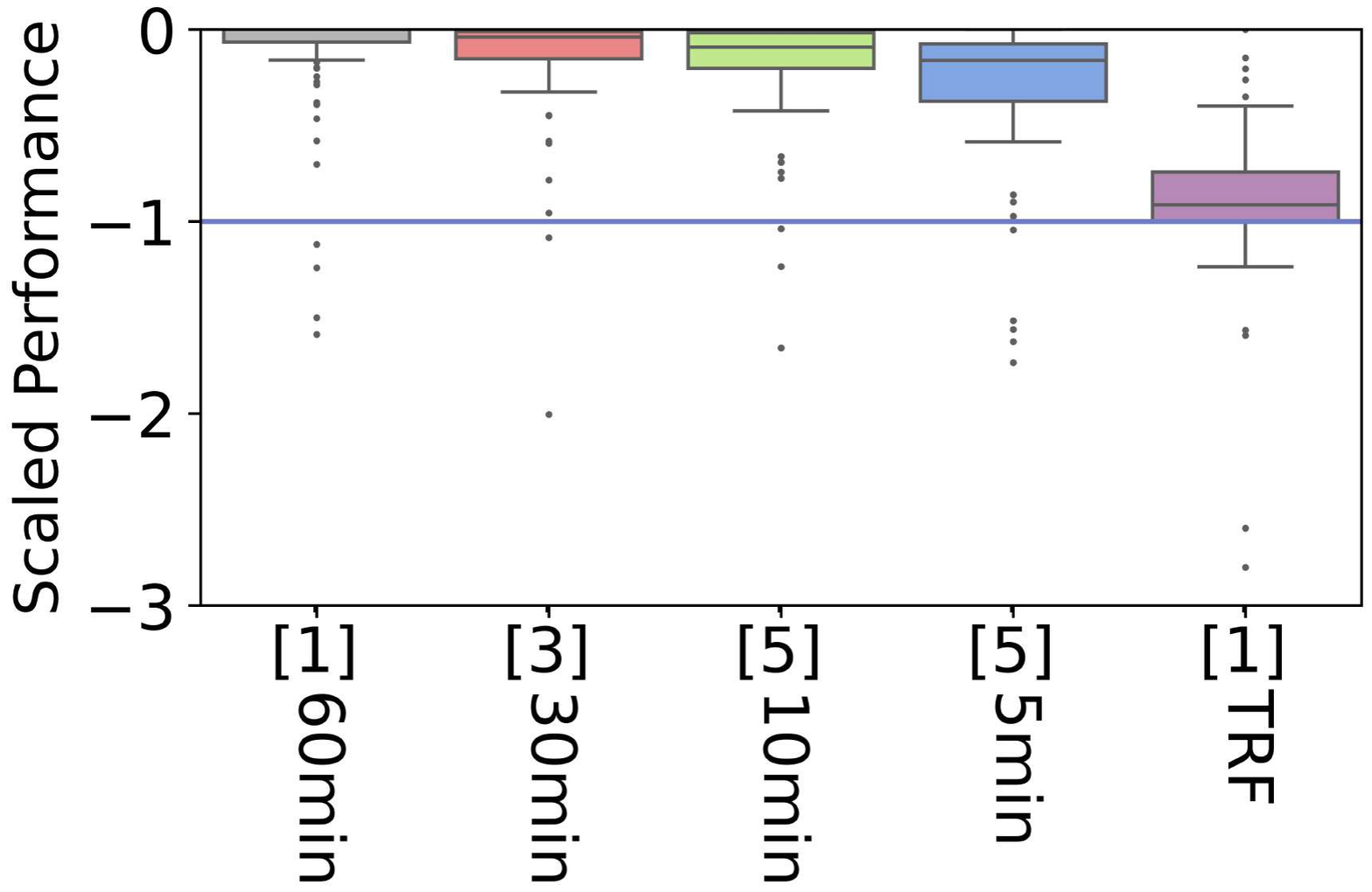}
\caption{\tiny AutoGluon(HQ)}
\end{subfigure}
\begin{subfigure}{0.31\textwidth}
\includegraphics[width=\linewidth]{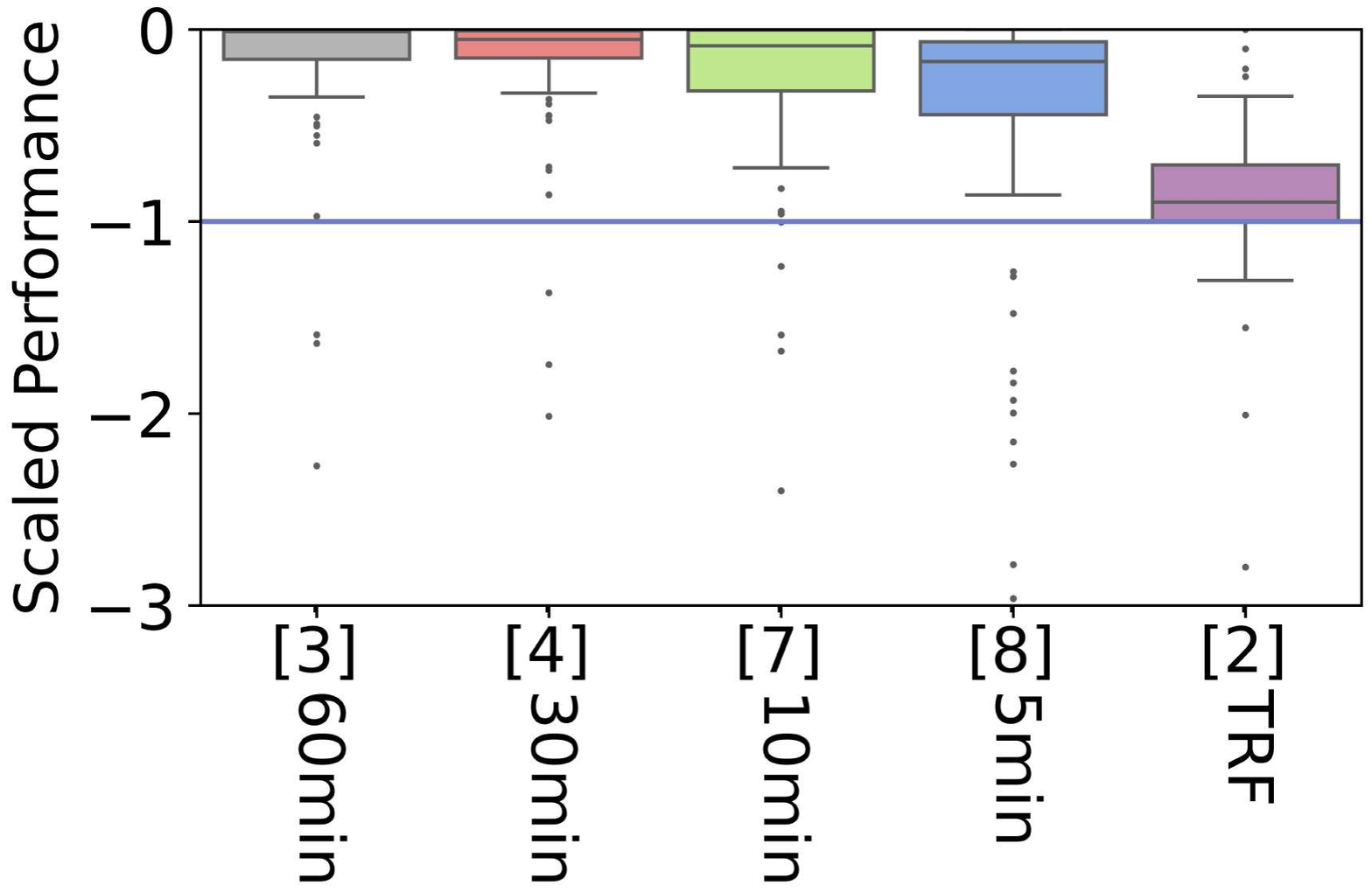}
\caption{\tiny AutoGluon(HQIL)}
\end{subfigure}
\newline
\begin{subfigure}{0.31\textwidth}
\includegraphics[width=\linewidth]{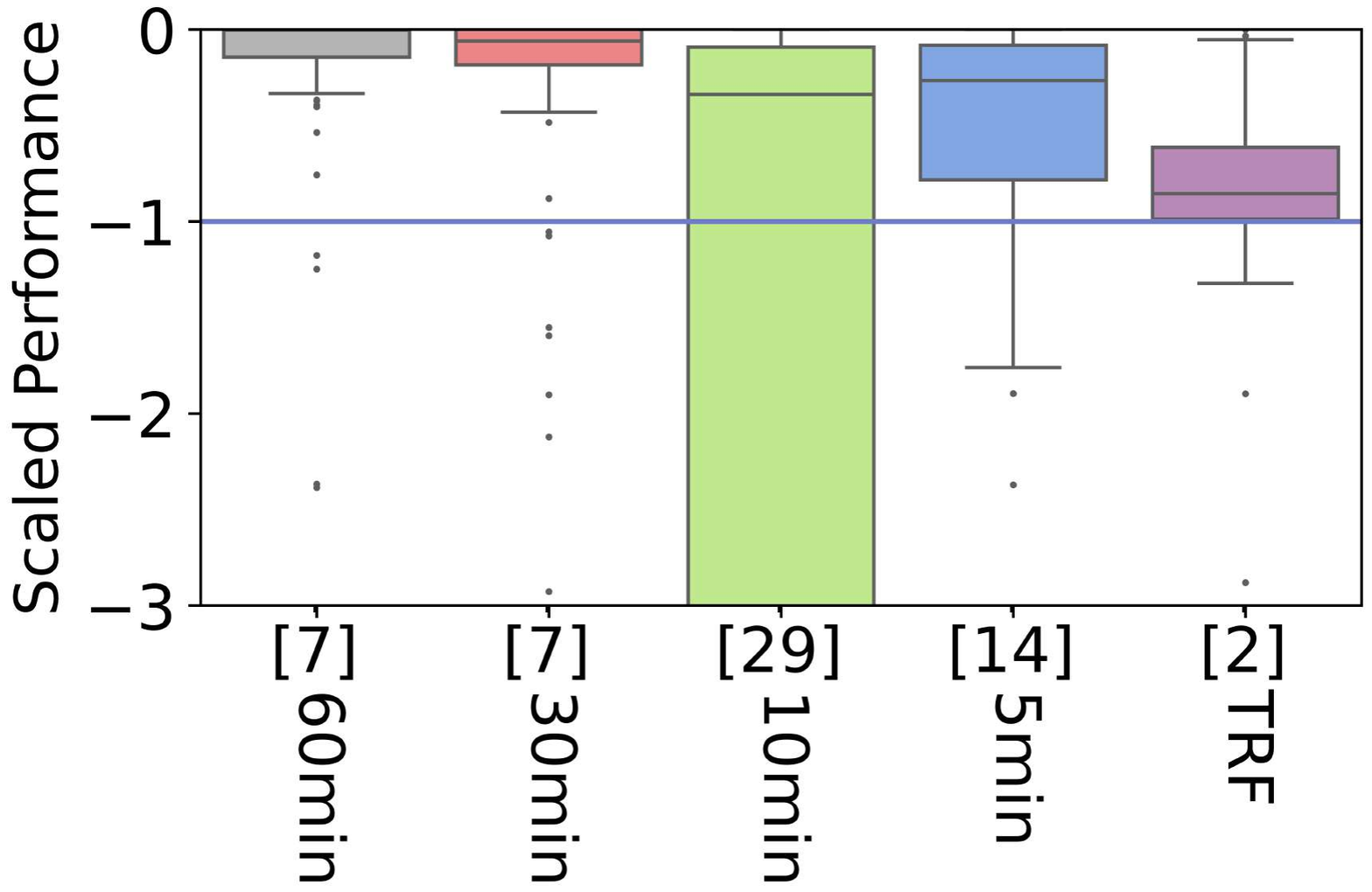}
\caption{\tiny autosklearn}
\end{subfigure}
\begin{subfigure}{0.31\textwidth}
\includegraphics[width=\linewidth]{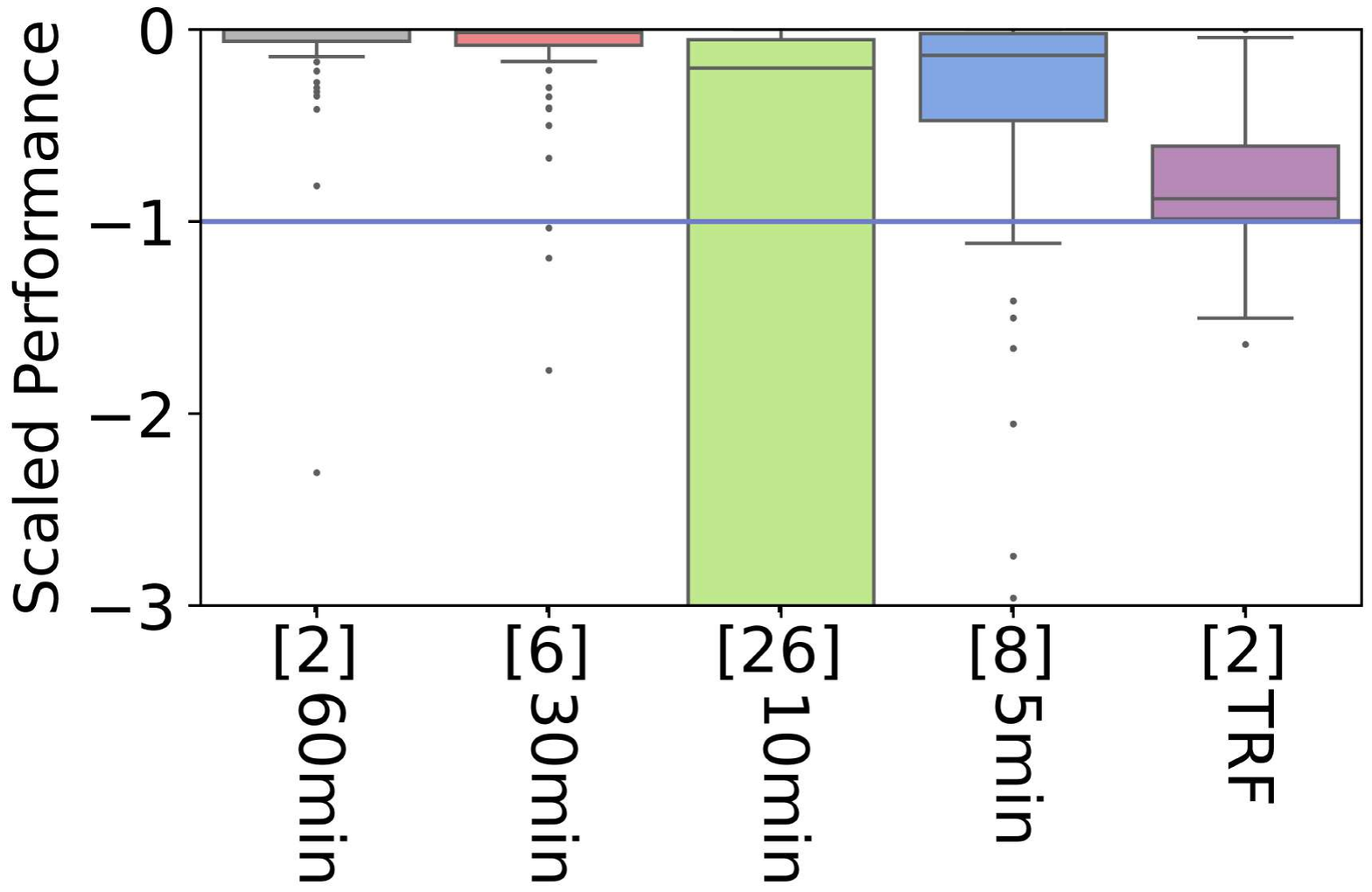}
\caption{\tiny autosklearn2}
\end{subfigure}
\begin{subfigure}{0.31\textwidth}
\includegraphics[width=\linewidth]{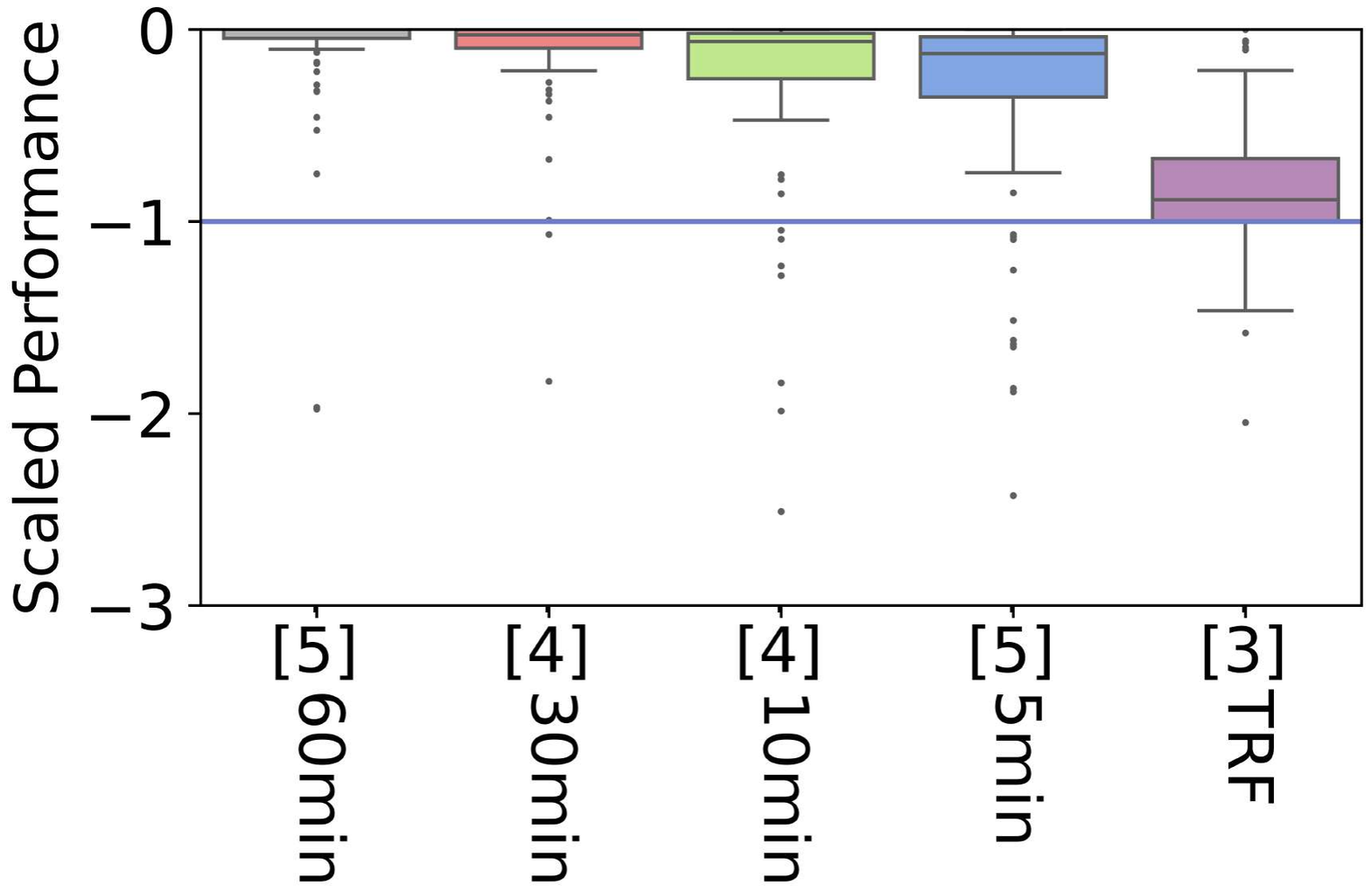}
\caption{\tiny flaml}
\end{subfigure}
\newline
\begin{subfigure}{0.31\textwidth}
\includegraphics[width=\linewidth]{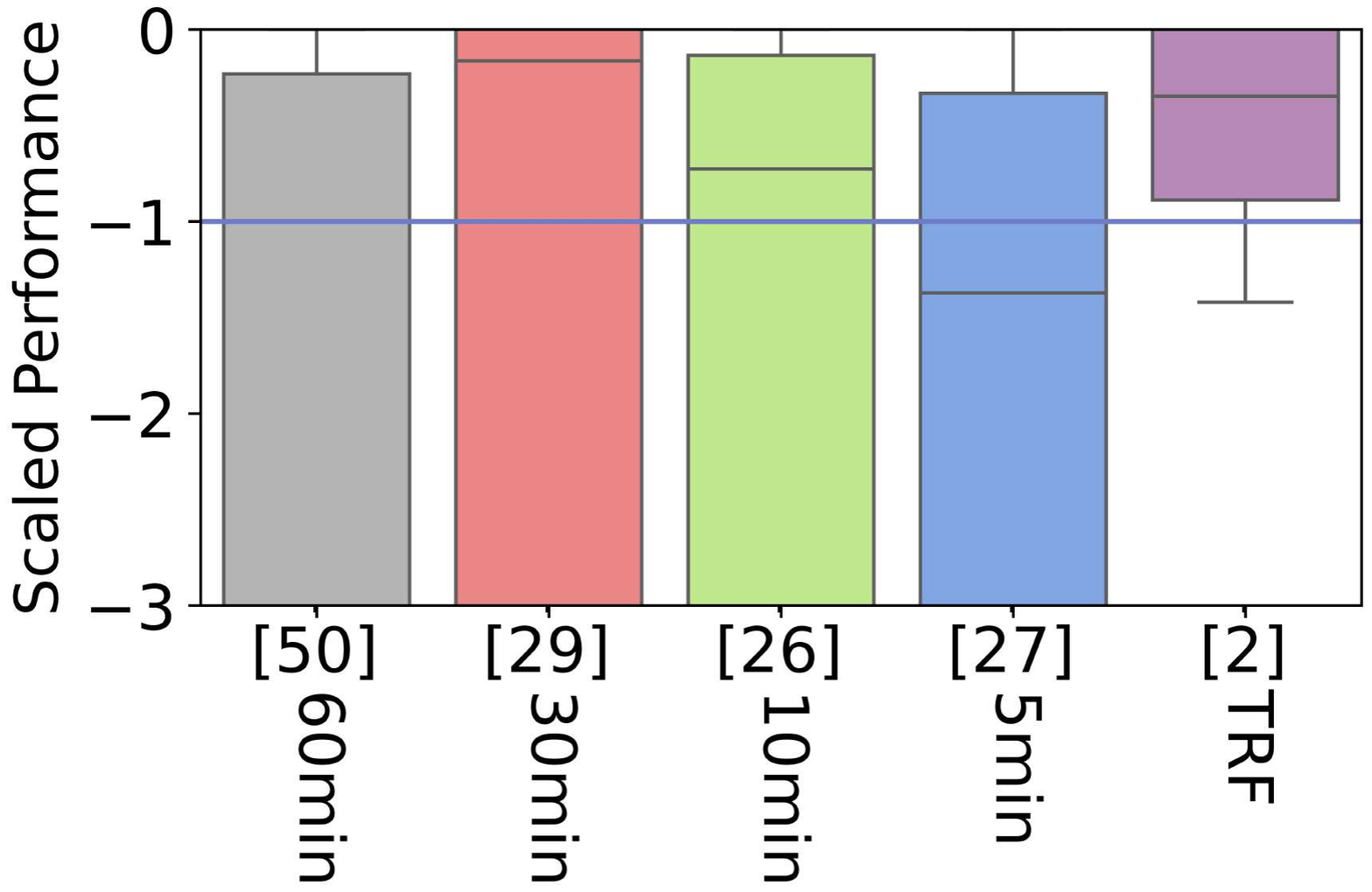}
\caption{\tiny FEDOT}
\end{subfigure}
\begin{subfigure}{0.31\textwidth}
\includegraphics[width=\linewidth]{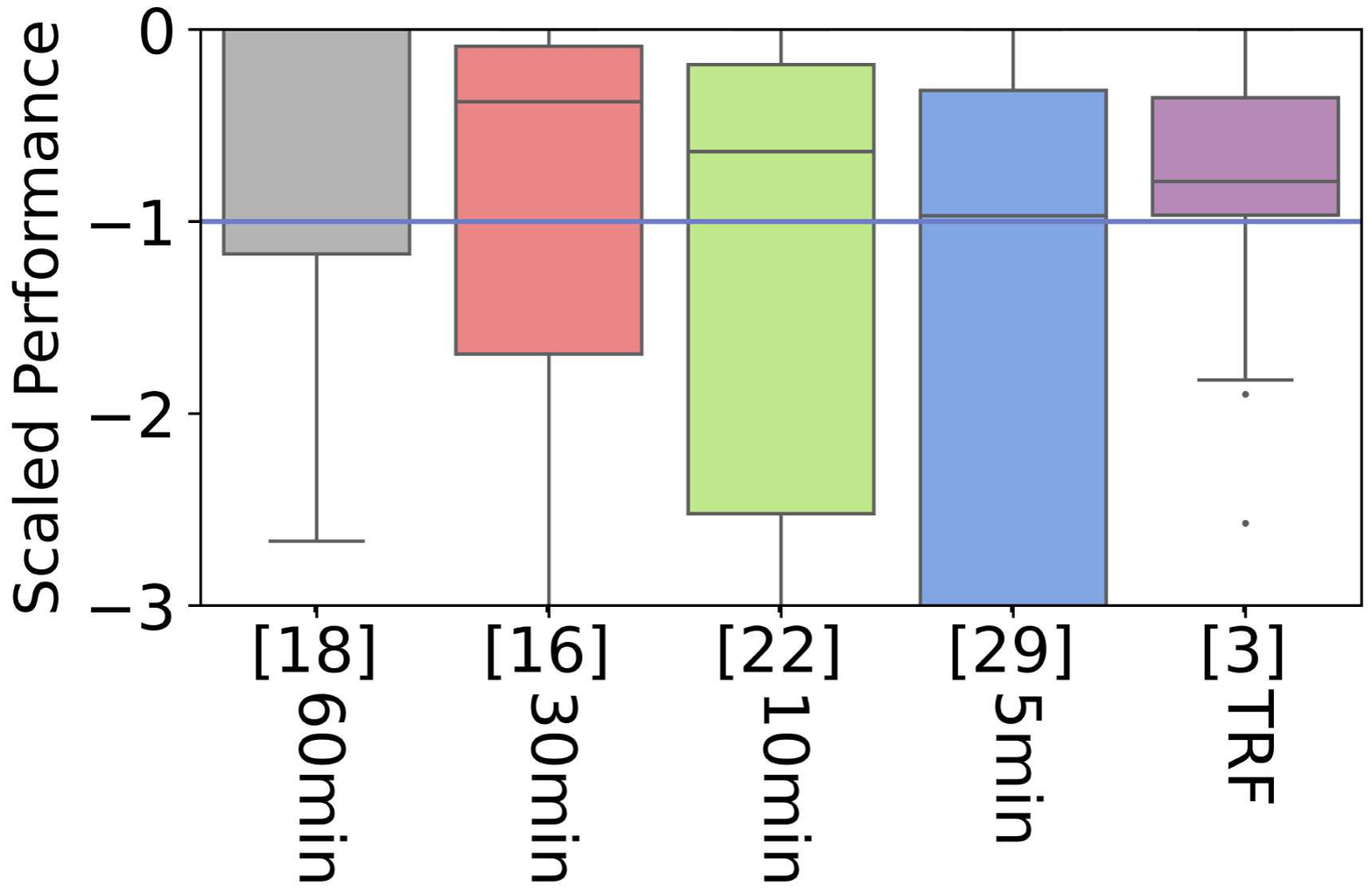}
\caption{\tiny GAMA}
\end{subfigure}
\begin{subfigure}{0.31\textwidth}
\includegraphics[width=\linewidth]{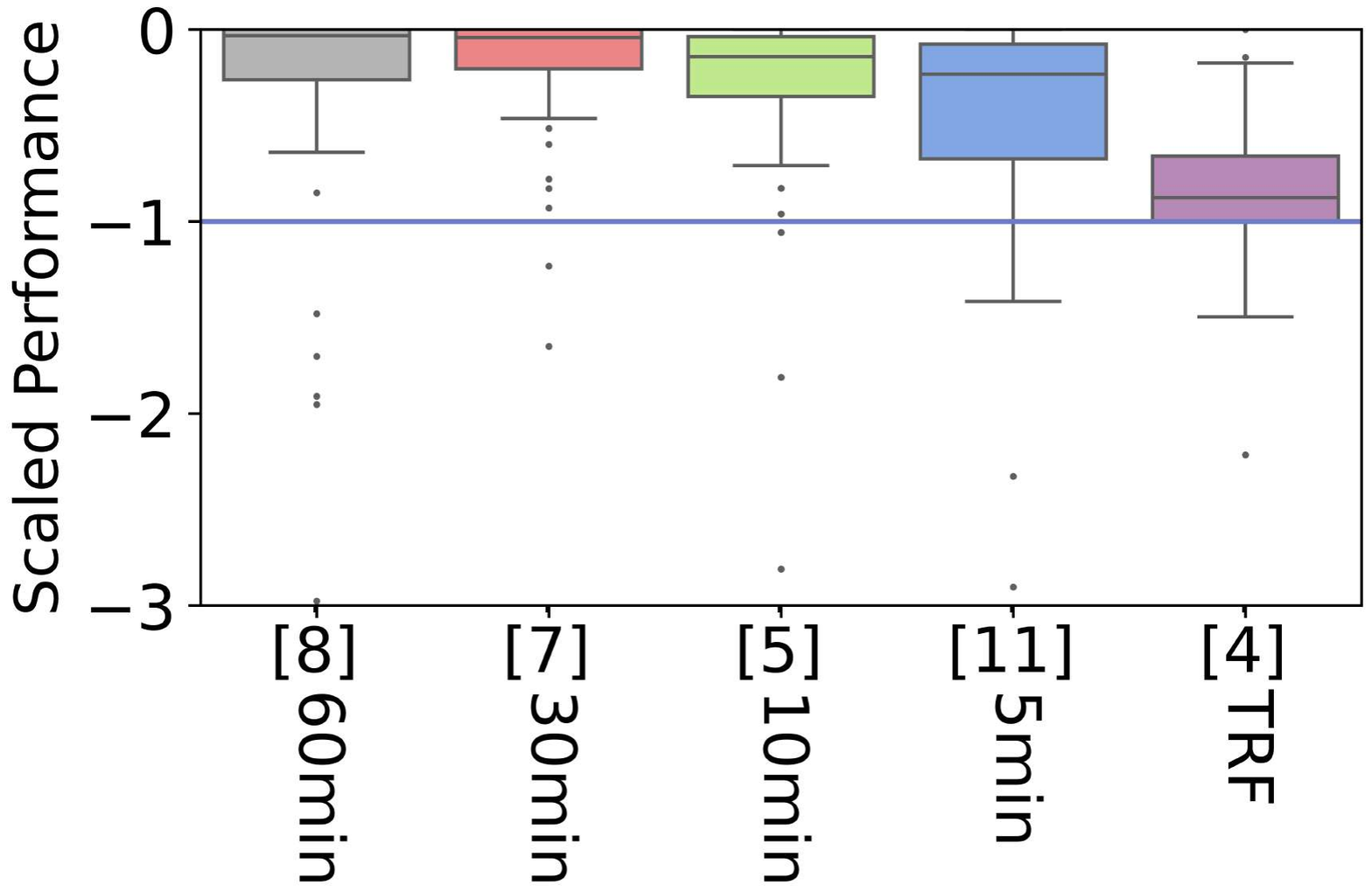}
\caption{\tiny H2OAutoML}
\end{subfigure}
\newline
\begin{subfigure}{0.31\textwidth}
\includegraphics[width=\linewidth]{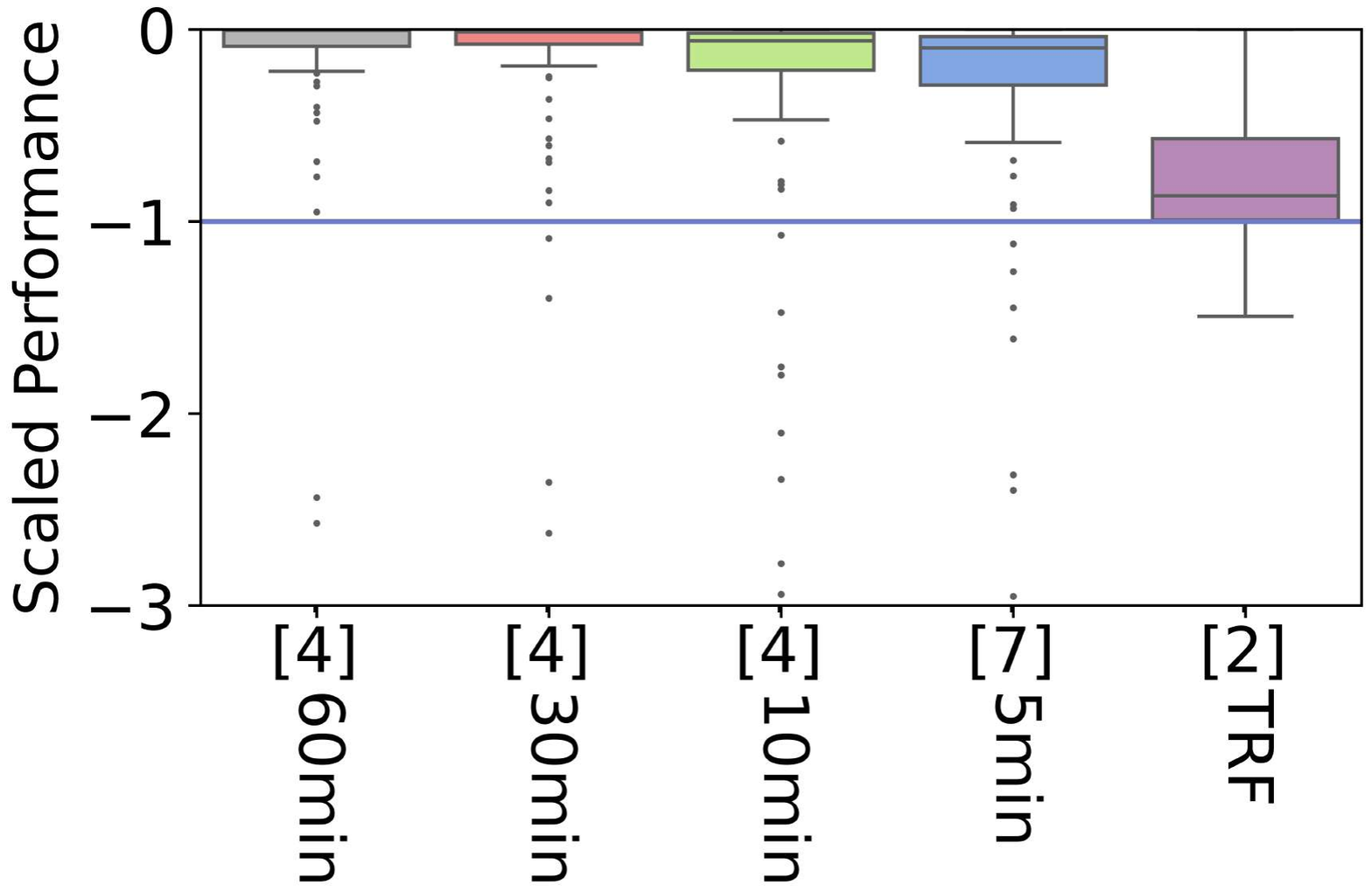}
\caption{\tiny lightautoml}
\end{subfigure}
\begin{subfigure}{0.31\textwidth}
\includegraphics[width=\linewidth]{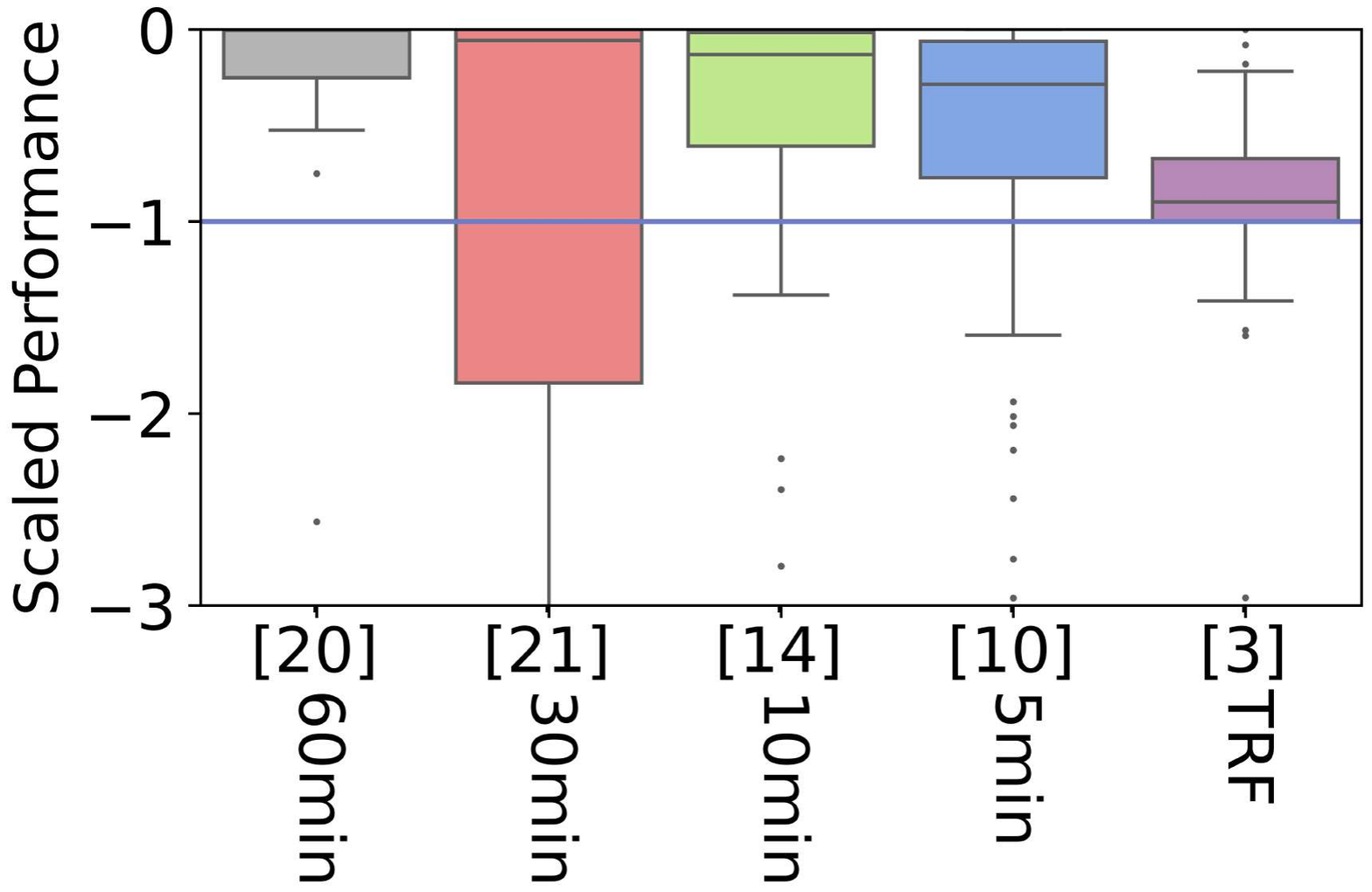}
\caption{\tiny MLJAR(B)}
\end{subfigure}
\begin{subfigure}{0.31\textwidth}
\includegraphics[width=\linewidth]{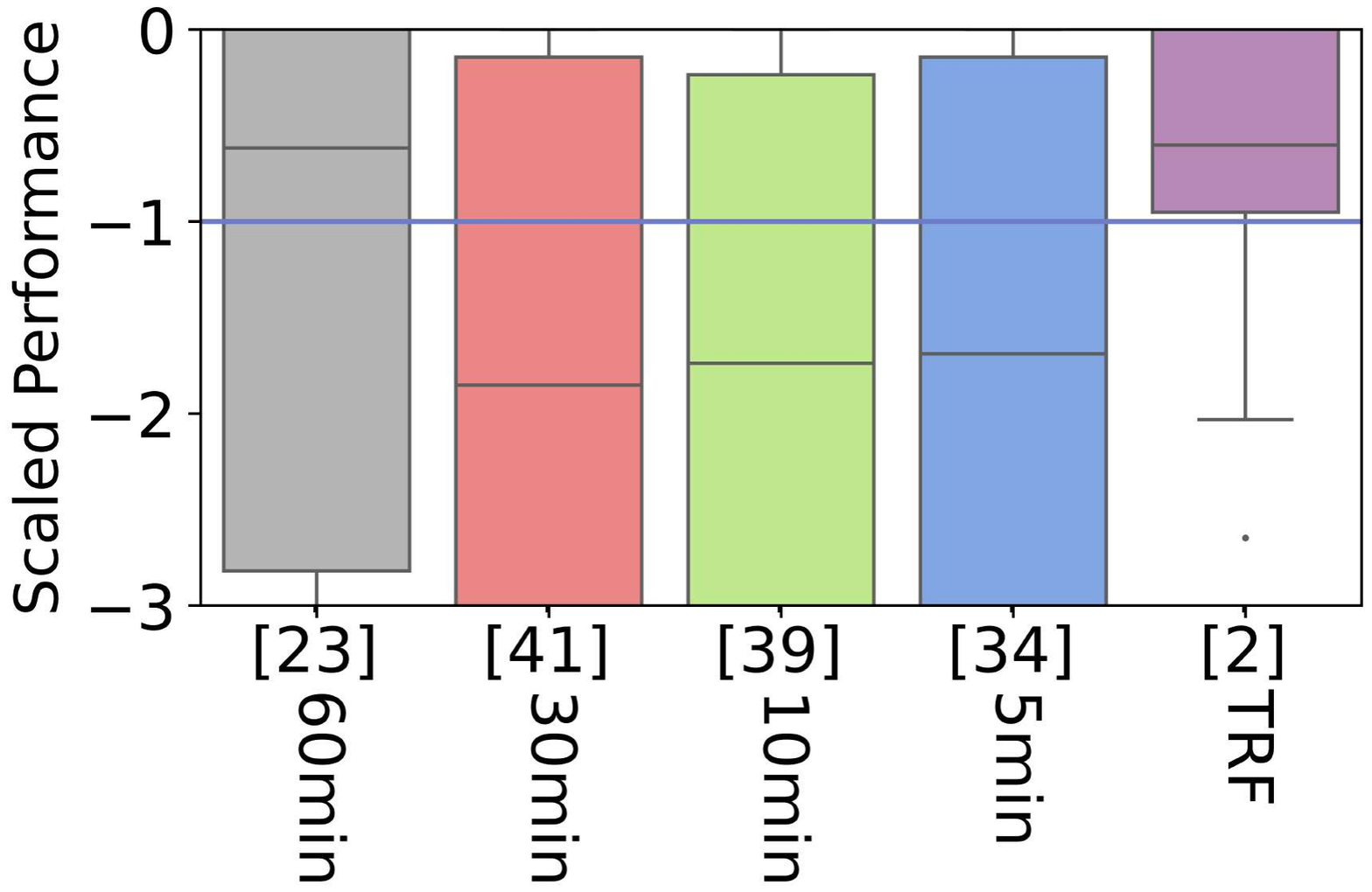}
\caption{\tiny NaiveAutoML}
\end{subfigure}
\newline
\begin{subfigure}{0.31\textwidth}
\includegraphics[width=\linewidth]{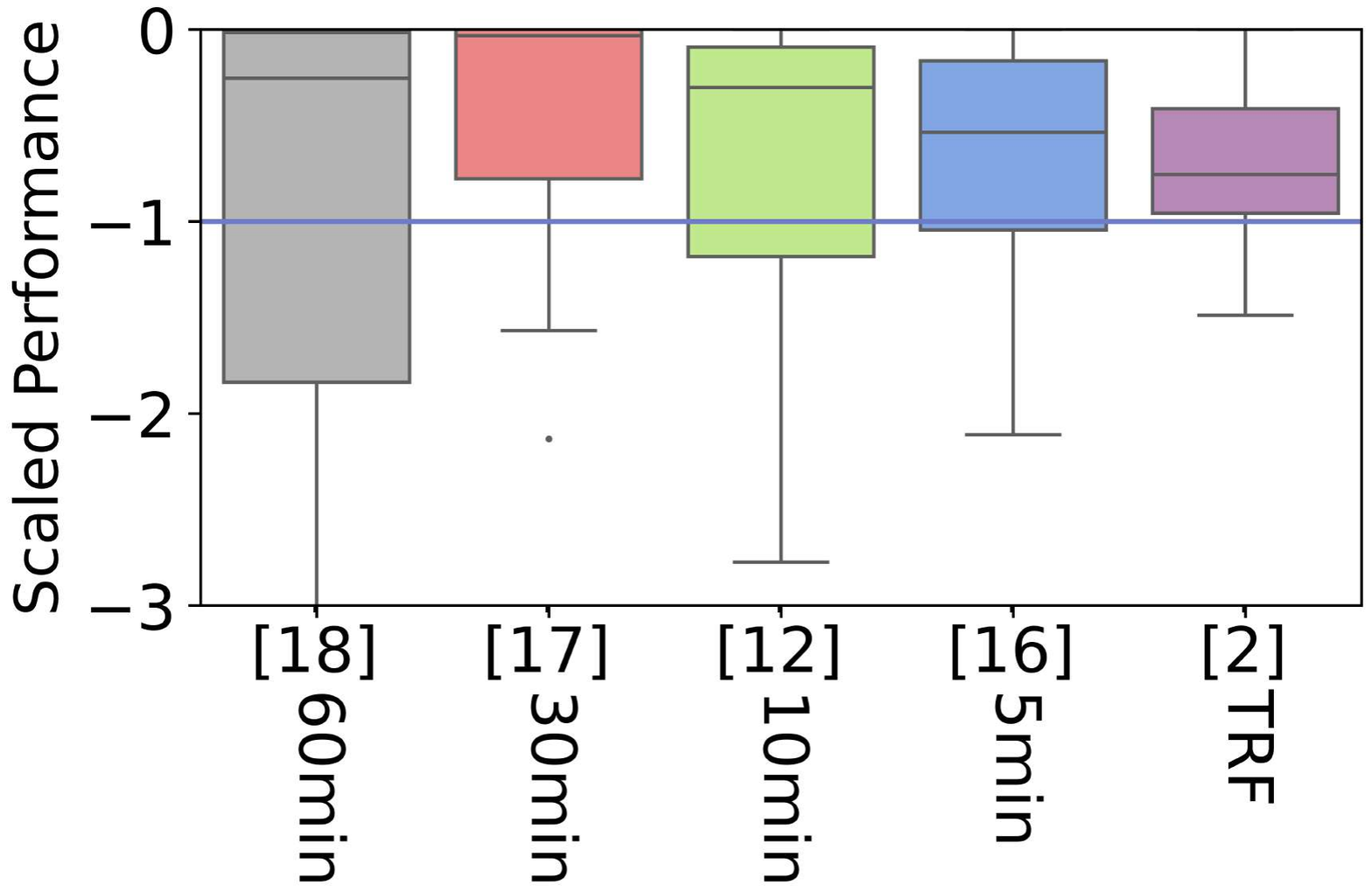}
\caption{\tiny TPOT}
\end{subfigure}

\end{center}
\caption{Boxplots of framework results evaluated by time constraints, distribution after scaling the performance values from the random forest $\mathtt{RF}$ 60 minutes (-1 horizontal line) to best observed (0). The number of outliers for each framework that are not shown in the plot is denoted on the x-axis. \label{fig:appendix_boxplot}}
\end{figure}

Now, the distribution of scaled performance across different time constraints per framework is presented. Figure \ref{fig:appendix_boxplot}, shows that $\mathtt{flaml}$, $\mathtt{H2OAutoML}$, $\mathtt{AutoGluon}$ maintain stable performance distributions across different time budgets, with their performance often clustering closer to 0, suggesting that they can be effective even with 5 or 10 minutes of evaluation. However, frameworks like $\mathtt{NaiveAutoML}$, $\mathtt{FEDOT}$ and $\mathtt{GAMA}$ show more variability in shorter time frames, with larger numbers of outliers, implying that they may require longer evaluations to consistently perform well. While many frameworks demonstrate potential at lower time constraints, extended evaluation times, such as 30 or 60 minutes, still offer more reliable performance, with fewer outliers and better overall stability.

\subsubsection{Performance and inference time}

\begin{figure}[h]
\begin{center}

\begin{subfigure}{0.45\textwidth}
\includegraphics[width=\linewidth]{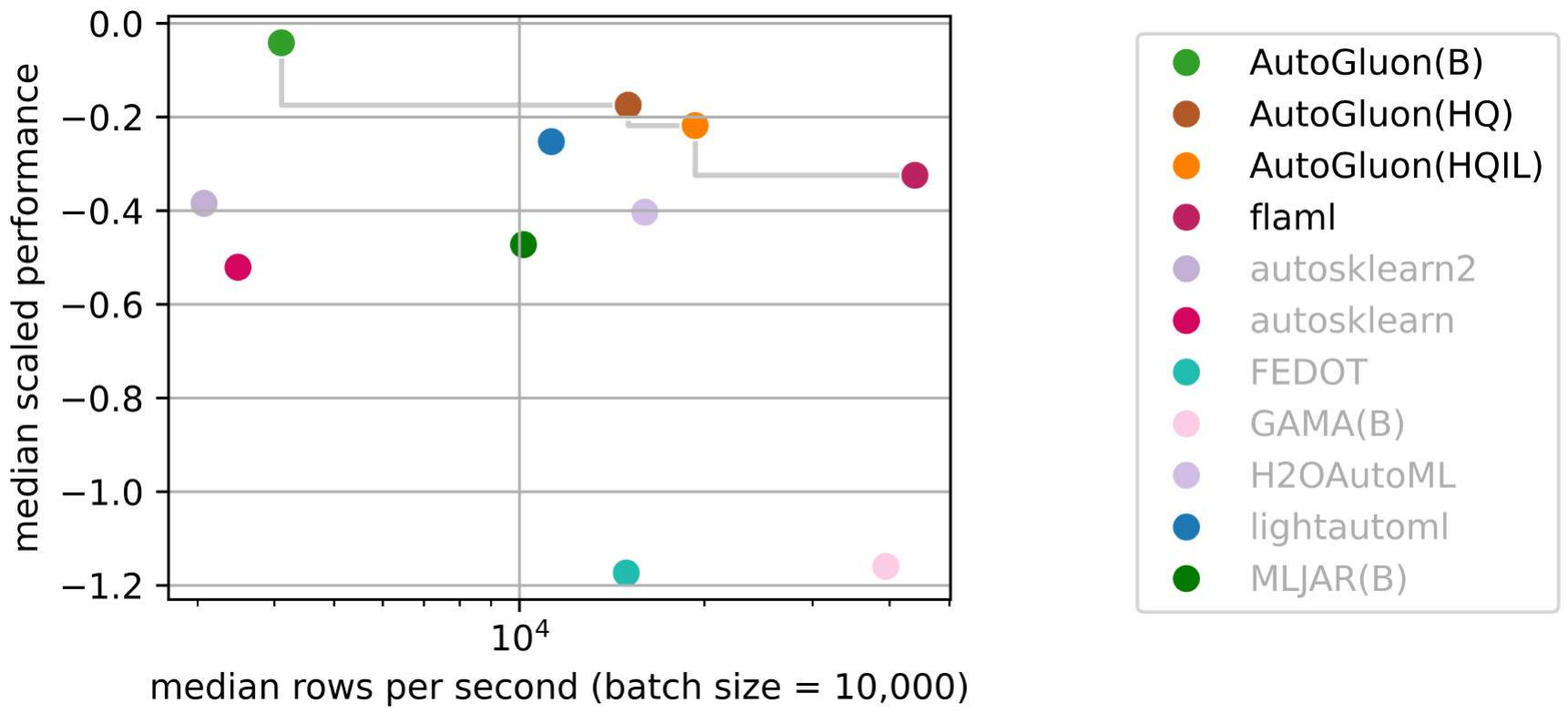}
\caption{5 minutes evaluation}
\end{subfigure}
\begin{subfigure}{0.45\textwidth}
\includegraphics[width=\linewidth]{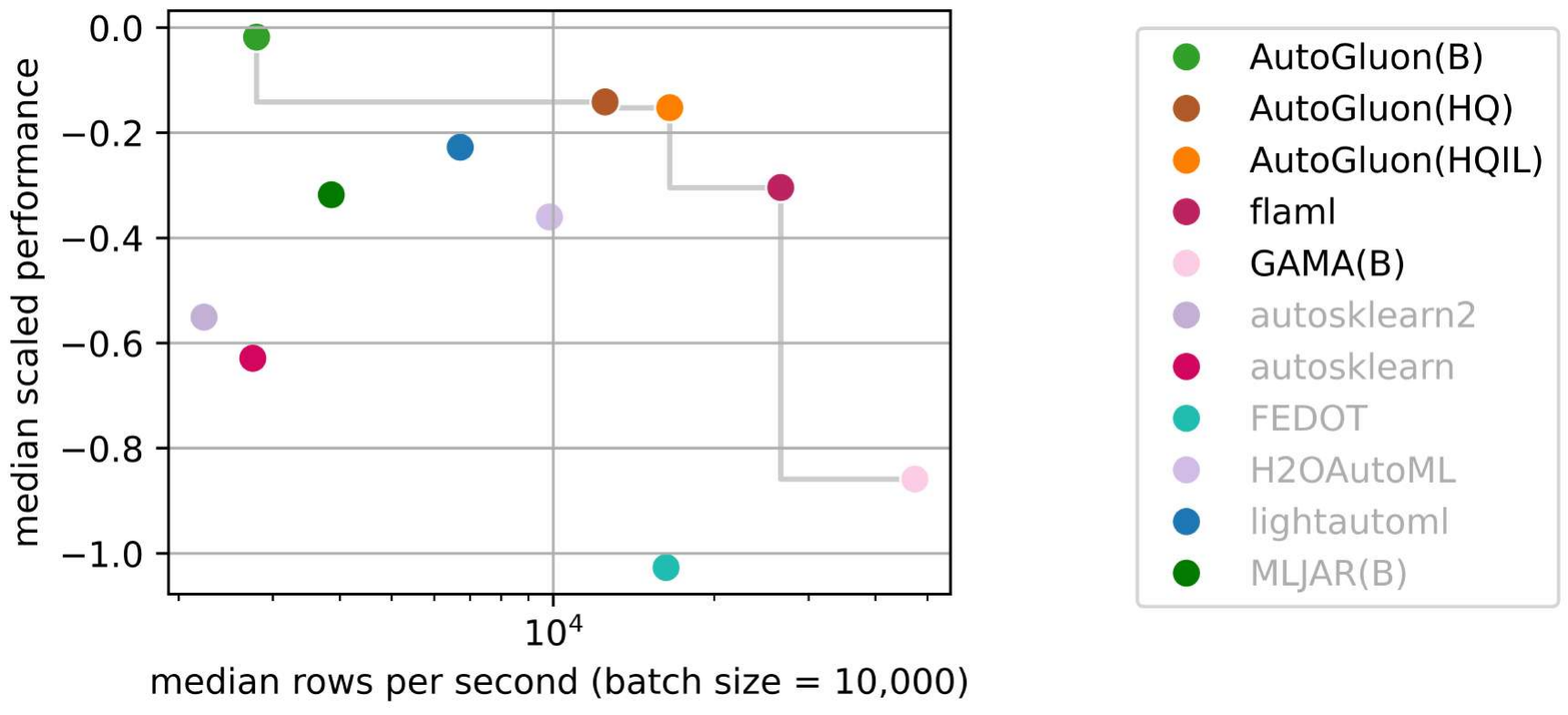}
\caption{10 minutes evaluation}
\end{subfigure}
\newline
\begin{subfigure}{0.45\textwidth}
\includegraphics[width=\linewidth]{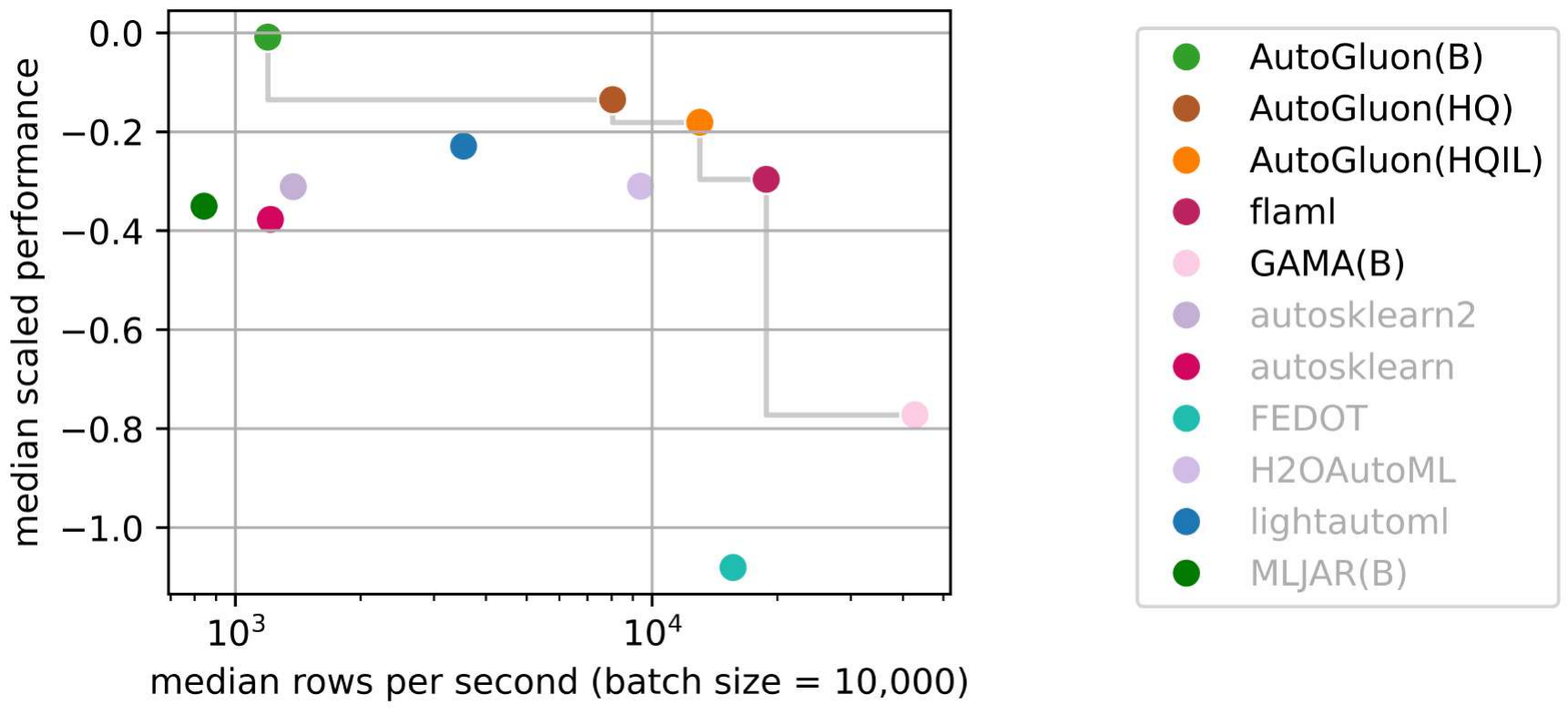}
\caption{30 minutes evaluation}
\end{subfigure}
\begin{subfigure}{0.45\textwidth}
\includegraphics[width=\linewidth]{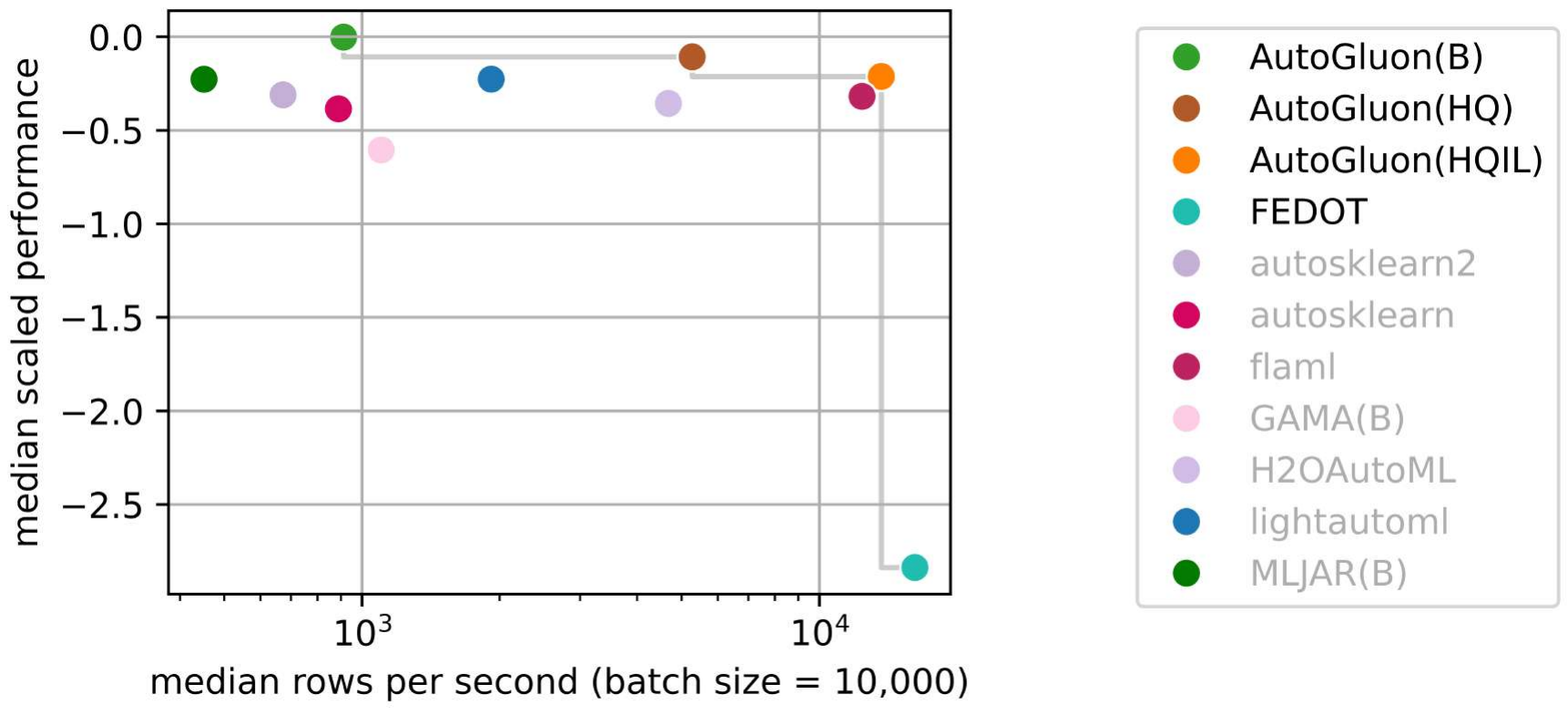}
\caption{60 minutes evaluation}
\end{subfigure}

\end{center}
\caption{Pareto Frontier of framework evaluated by time constraints. The plots show the performance values are scaled from the random forest (-1) to best observed (0) versus the inference speed in median per second with a batch size of 10,000. \label{fig:pf_perf_inference_times}}
\end{figure}

Figure \ref{fig:pf_perf_inference_times} illustrates the Pareto frontiers of the normalized model accuracy against corresponding median per-row prediction speeds. It is observed that $\mathtt{flaml}$ and $\mathtt{AutoGluon}$ setups dominated the Pareto frontier, and $\mathtt{GAMA}$ produce lightweight models accelerating inference but with several times less power in performance. Starting with the 5-minute evaluation (a), we observe that $\mathtt{AutoGluon(HQ)}$ and $\mathtt{AutoGluon(HQIL)}$ achieve the best balance between performance and speed, as both frameworks are situated closer to the top-right corner. These frameworks demonstrate their ability to maintain relatively high accuracy while delivering fast prediction speeds, suggesting that they are well-suited for scenarios where computational resources and time are limited. $\mathtt{flaml}$ displays being competitive in this shortest scenario and trade-off. Similarly, at the 30-minute evaluation (c), $\mathtt{AutoGluon(HQIL)}$ emerges as the front-runner, with $\mathtt{GAMA}$ offering competitive prediction speeds but with at least 3 times lower performance compared to $\mathtt{AutoGluon}$ frameworks. Finally, in the 60-minute evaluation (d), the shift in $\mathtt{FEDOT}$ 's performance is due to a large number of missing values since $\mathtt{FEDOT}$ cannot handle large datasets well and missing values are imputed with the $\mathtt{CP}$, makes to look the models faster than they really are.

\begin{figure}[h]
\begin{center}
\includegraphics[width=0.7\linewidth]{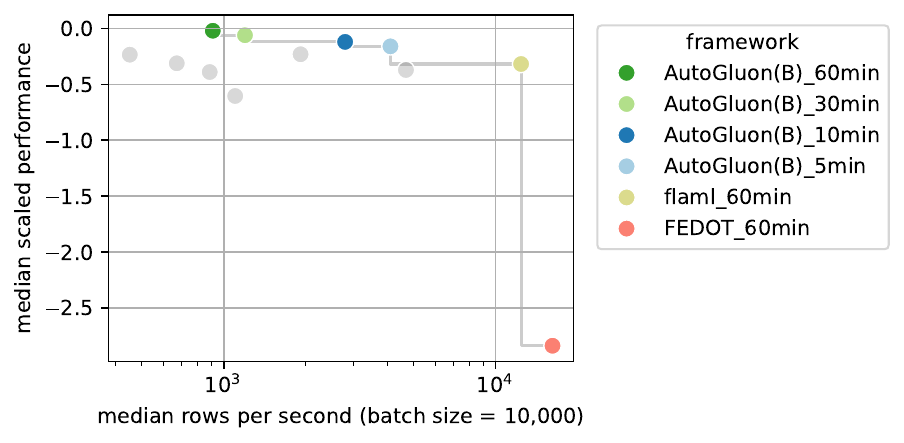}
\end{center}
\caption{Pareto Frontier of all the framework evaluated over 60 minutes versus the best framework performance in all time setups. Only the Pareto efficient methods are displayed in the legend. \label{fig:pf_perf_inference_times_general}}
\end{figure}

In the Figure \ref{fig:pf_perf_inference_times_general}. it is appreciated the Pareto frontier of all the frameworks over 60 minutes versus the best framework found over the four-time constraints in CD. $\mathtt{FEDOT}$ is not really performing well, as mentioned; it has a large scale of missing values, which biased the inference time for the imputation with $\mathtt{CP}$. The objective of this Figure is to find if the best-performed AutoML system in the CD is capable of achieving meaningful results with less than one hour of runtime on many datasets. It can be seen that $\mathtt{AutoGluon}$ is able to succeed on all 104 datasets with 10 folds on \verb|best_quality| using a 10-minute constraint and even a 5-minute constraint. 

\begin{figure}[h]
\begin{center}

\begin{subfigure}{0.31\textwidth}
\includegraphics[width=\linewidth]{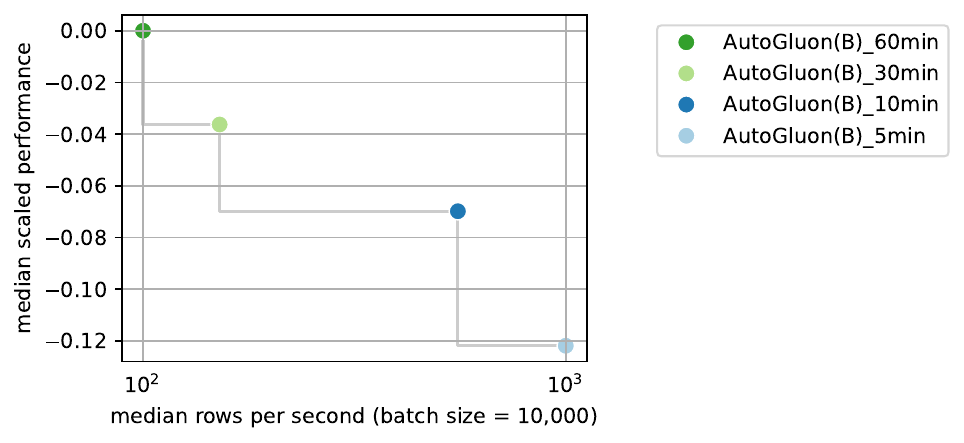}
\caption{AutoGluon(B)}
\end{subfigure}
\begin{subfigure}{0.31\textwidth}
\includegraphics[width=\linewidth]{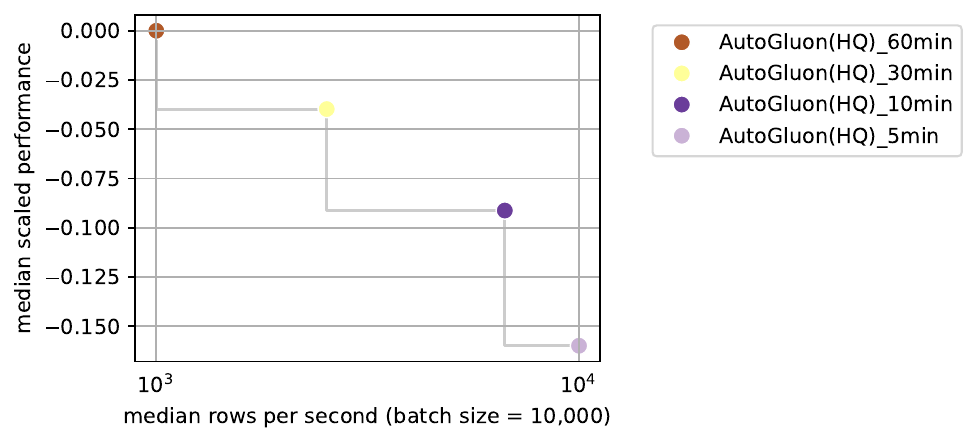}
\caption{AutoGluon(HQ)}
\end{subfigure}
\begin{subfigure}{0.31\textwidth}
\includegraphics[width=\linewidth]{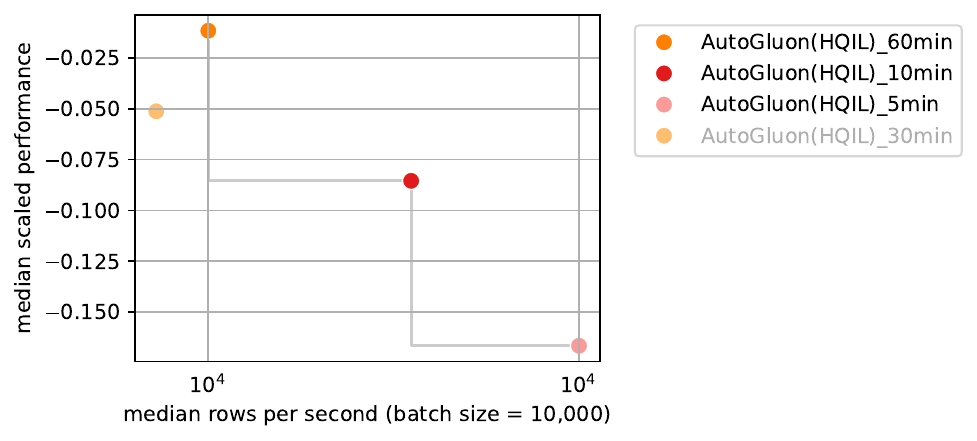}
\caption{\tiny AutoGluon(HQIL)}
\end{subfigure}
\newline
\begin{subfigure}{0.31\textwidth}
\includegraphics[width=\linewidth]{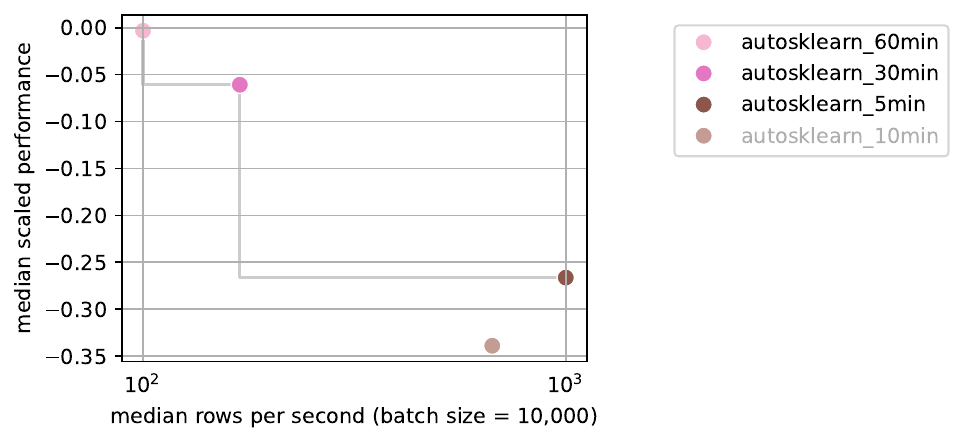}
\caption{autosklearn}
\end{subfigure}
\begin{subfigure}{0.31\textwidth}
\includegraphics[width=\linewidth]{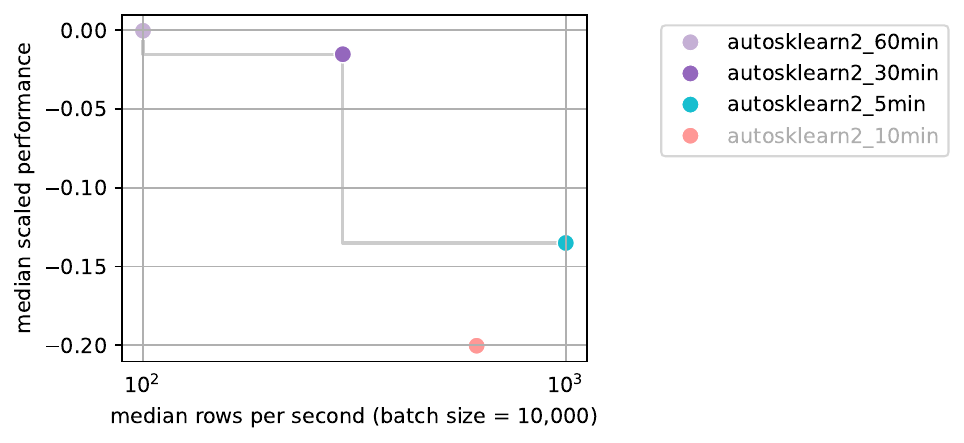}
\caption{autosklearn2}
\end{subfigure}
\begin{subfigure}{0.31\textwidth}
\includegraphics[width=\linewidth]{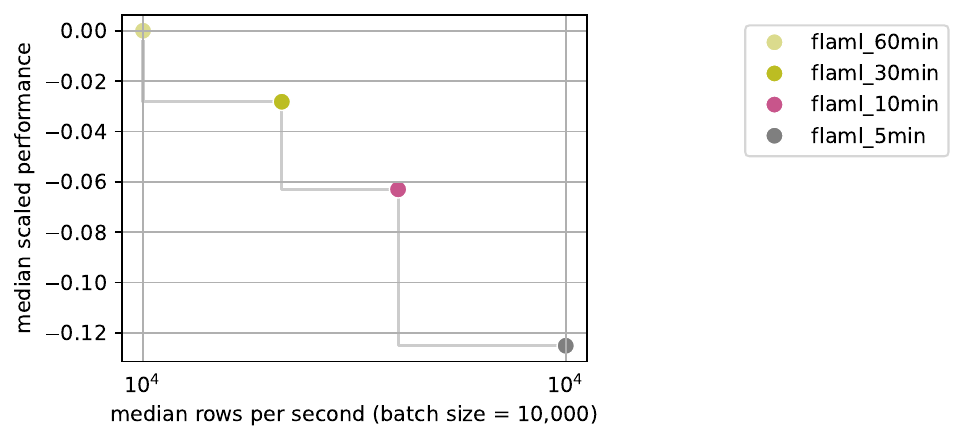}
\caption{flaml}
\end{subfigure}
\newline
\begin{subfigure}{0.31\textwidth}
\includegraphics[width=\linewidth]{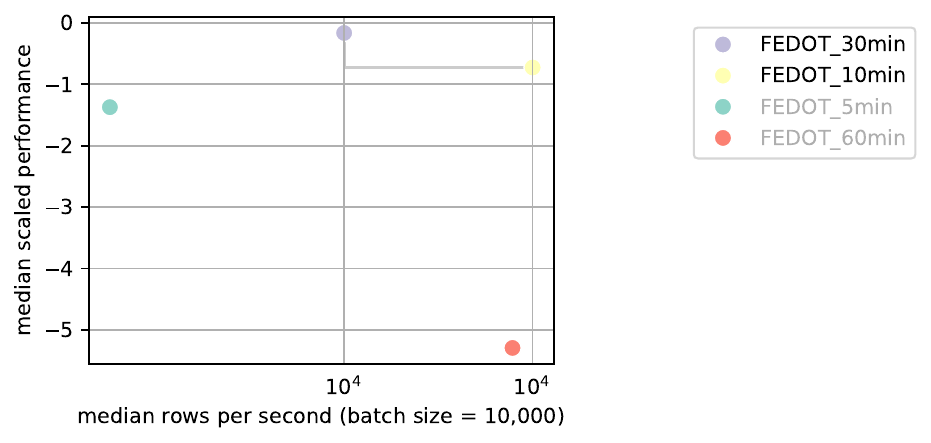}
\caption{FEDOT}
\end{subfigure}
\begin{subfigure}{0.31\textwidth}
\includegraphics[width=\linewidth]{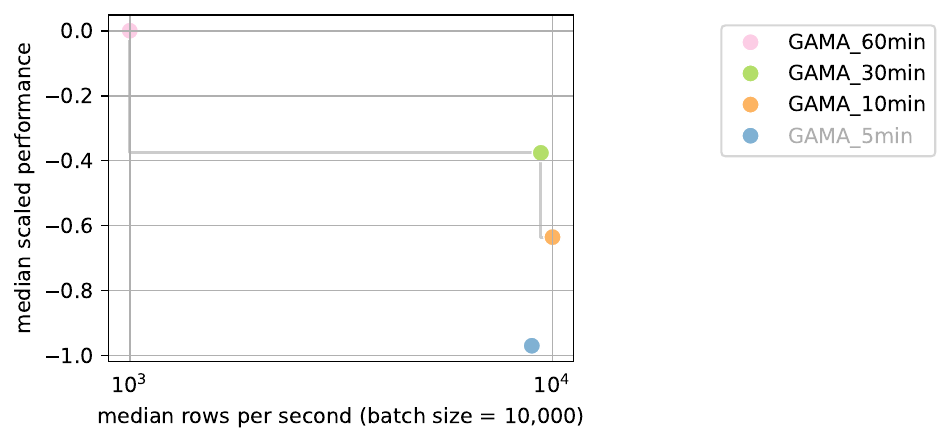}
\caption{GAMA(B)}
\end{subfigure}
\begin{subfigure}{0.31\textwidth}
\includegraphics[width=\linewidth]{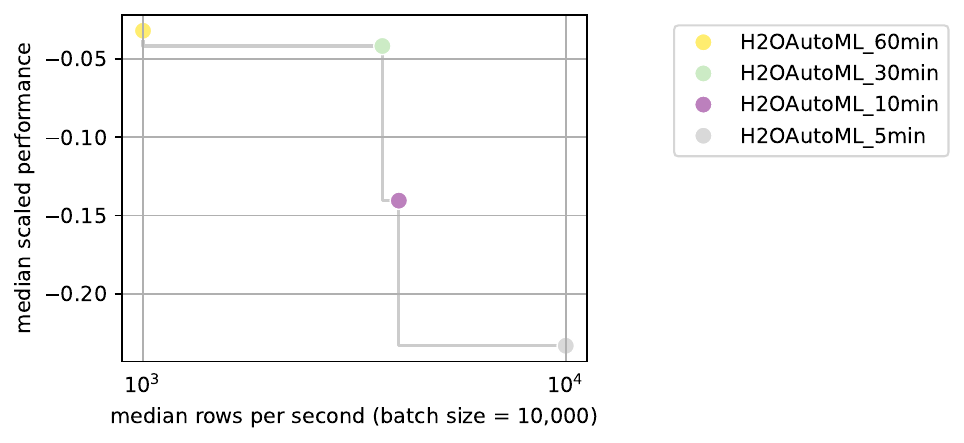}
\caption{H2OAutoML}
\end{subfigure}
\newline
\begin{subfigure}{0.31\textwidth}
\includegraphics[width=\linewidth]{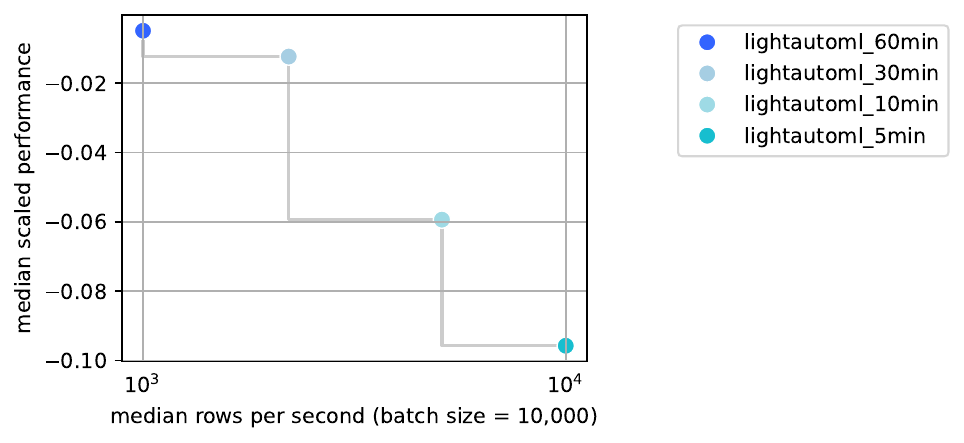}
\caption{lightautoml}
\end{subfigure}
\begin{subfigure}{0.31\textwidth}
\includegraphics[width=\linewidth]{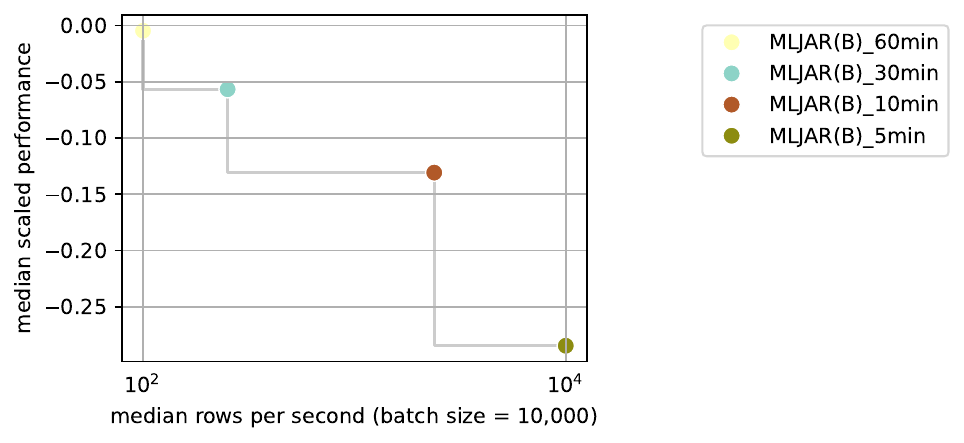}
\caption{MLJAR(B)}
\end{subfigure}
\begin{subfigure}{0.31\textwidth}
\includegraphics[width=\linewidth]{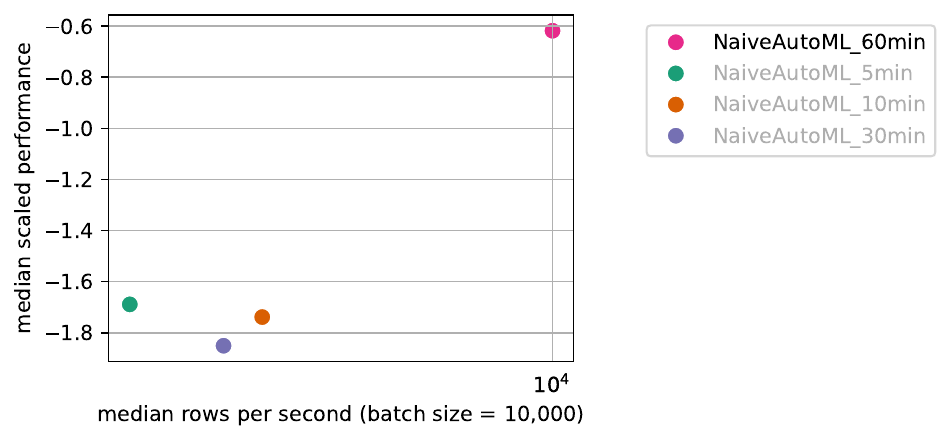}
\caption{NaiveAutoML}
\end{subfigure}
\newline
\begin{subfigure}{0.31\textwidth}
\includegraphics[width=\linewidth]{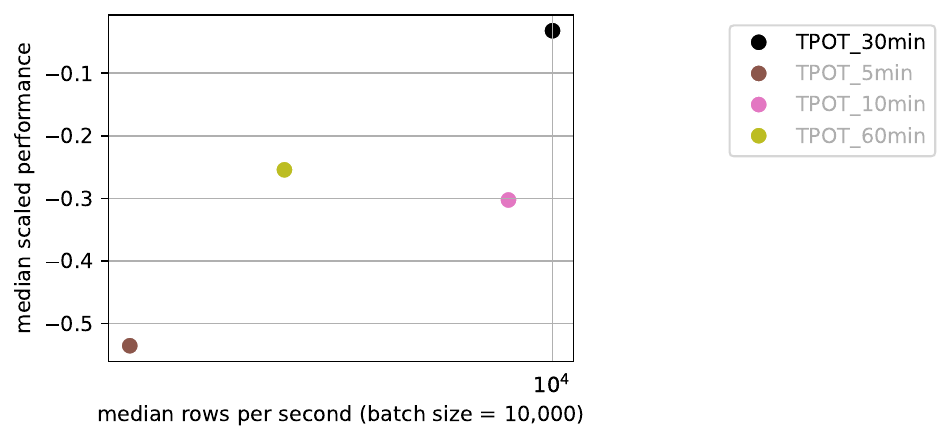}
\caption{TPOT}
\end{subfigure}

\end{center}
\caption{Pareto Frontier of framework evaluated by time constraints. The plots show the performance values are scaled from the random forest (-1) to best observed (0) versus the inference speed in median per second with a batch size of 10,000. \label{fig:appendix_pf_perf_inference_times}}
\end{figure}

Figure \ref{fig:appendix_pf_perf_inference_times} shows the Pareto Frontier per framework evaluated by time constraints. It can be observed that most of the frameworks follow the same pattern: the higher the inference, the lower the performance. However, there are some exceptions, as in the case of $\mathtt{MLJAR(B)}$, $\mathtt{NaiveAutoML}$ and $\mathtt{TPOT}$.

\clearpage
\subsection{Time constraint evaluations and early stopping} \label{cross:appendix_time_early}

\begin{figure}[h]
\begin{center}
\includegraphics[width=\linewidth]{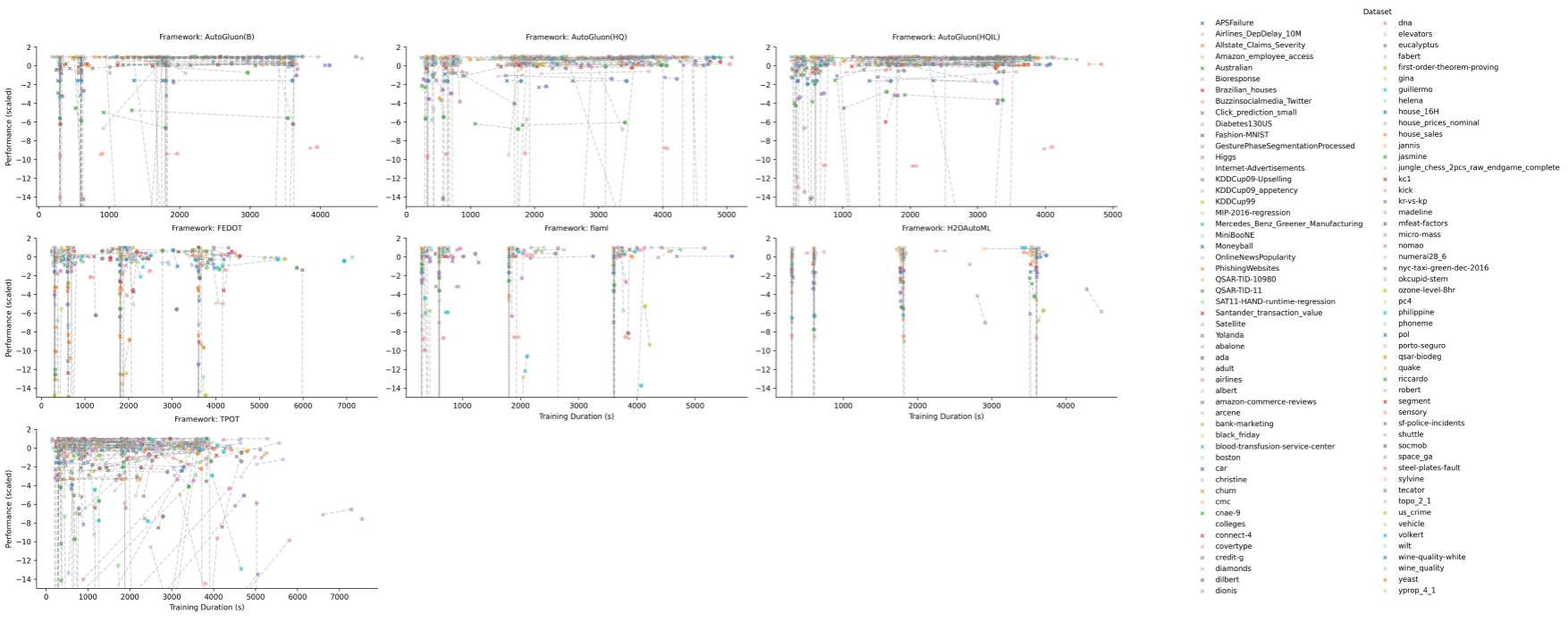}
\caption{Trade-offs between time savings and performance changes when applying early stopping by framework and tasks. Circles (O) indicate the results without early stopping (longer runtime, original performance), while crosses (X) represent the results with early stopping (shorter runtime, adjusted performance). \label{fig:convergence_analysis}}
\end{center}
\end{figure}

To show whether early stopping accelerates convergence without sacrificing performance, Figure \ref{fig:convergence_analysis} provides a framework-specific visualization of the trade-offs between time savings and performance changes when applying early stopping. Each subplot corresponds to a specific framework and displays all its evaluated tasks. Paired points represent the performance and runtime for each dataset: circles (O) indicate the results without early stopping (longer runtime, original performance), while crosses (X) represent the results with early stopping (shorter runtime, adjusted performance). The dashed lines connecting paired points illustrate the trade-offs for individual datasets, where the direction and magnitude of change can be observed. Unique dataset colours allow easy tracking of individual task behaviour across the subplots. Key interpretations include: negative \textit{regret} (performance gain), where X is higher than O on the y-axis; \textit{time saved}, where X is leftward of O on the x-axis; and diminishing returns, where X achieves comparable or better performance than o but with significantly reduced runtime.

It can be observed that $\mathtt{flaml}$ and $\mathtt{H2OAutoML}$ are highly conservative in terms of both time saved and performance. In contrast, $\mathtt{TPOT}$ often exceeds its time constraints, which explains its high number of tasks with significant time savings. $\mathtt{FEDOT}$ demonstrates a more reserved approach to saving time, often at the expense of performance. Meanwhile, $\mathtt{AutoGluon}$ achieves substantial time savings while maintaining comparable or even improved performance.

We investigated whether early stopping benefits larger datasets by testing the correlation between negative regret and dataset properties, such as the number of instances, number of classes, and feature-to-instance ratio, while controlling for the effect of time constraints. The analysis used results from several state-of-the-art AutoML frameworks across 104 tasks tested under four time constraints (5, 10, 30, and 60 minutes). Contrary to our expectation, the correlation between negative regret and dataset size (measured by the number of instances) was negligible, with a Pearson correlation coefficient of $r = 0.0076$, a confidence interval of $\left[ -0.04,  0.05 \right]$, and a $p-$value of 0.737. This lack of a significant relationship suggests that early stopping may not provide substantial advantages for larger datasets by preventing computationally expensive over-refinement. Instead, the findings highlight that the effect of early stopping might be more nuanced and dependent on other factors, such as dataset complexity or the AutoML framework's behaviour. When analyzing each framework independently (Table \ref{tab:partial_corr}) , it can be appreciated that most frameworks exhibit negligible or nonsignificant correlations between negative regret and dataset complexity, as measured by the number of instances. Notably, $\mathtt{AutoGluon(HQIL)}$ stands out with a moderate positive correlation ($r=0.436$, CI95\%: [0.34,0.53], $p<$0.001), suggesting that for this configuration, larger datasets might benefit more from early stopping by avoiding over-refinement. In contrast, other frameworks, such as FEDOT, flaml, H2OAutoML, and TPOT, show correlations close to zero, with wide confidence intervals that include both positive and negative values, indicating no clear relationship. These results highlight the variability in how different frameworks handle dataset complexity in relation to early stopping, suggesting that the benefits of early stopping may depend on framework-specific optimization strategies and capabilities.

\begin{table}[!h]
\centering
\caption{Partial correlation results for different frameworks between the dataset complexity and regret}
\label{tab:partial_corr}
\begin{tabular}{|l|l|l|l|}
\hline
\multicolumn{1}{|c|}{\textbf{Framework}} & \multicolumn{1}{c|}{\textit{r}} & \multicolumn{1}{c|}{CI95\%} & \multicolumn{1}{c|}{\textit{p-}val} \\ \hline
AutoGluon(B)                             & 0.0506                        & {[}-0.07, 0.17{]}                  & 0.3976                            \\ \hline
AutoGluon(HQ)                            & 0.0117                        & {[}-0.11, 0.13{]}                  & 0.8446                            \\ \hline
AutoGluon(HQIL)                          & 0.4358                        & {[}0.34, 0.53{]}                   & 1.8492e-14                        \\ \hline
FEDOT                                    & -0.0021                       & {[}-0.12, 0.11{]}                  & 0.9714                            \\ \hline
flaml                                    & 0.0167                        & {[}-0.1, 0.13{]}                   & 0.7801                            \\ \hline
H2OAutoML                                & 0.024                         & {[}-0.09, 0.14{]}                  & 0.6765                            \\ \hline
TPOT                                     & 0.0204                        & {[}-0.1, 0.14{]}                   & 0.7323                            \\ \hline
\end{tabular}
\end{table}

In addition, to assess the statistical significance of negative regret, we performed a one-sample t-test for regression, binary classification, and multiclass classification tasks separately, after cleaning the data to handle missing and infinite values. The results showed variability in mean regret across task types: for regression tasks, the mean regret was 777.507 ($p =$ 0.145); for binary classification tasks, the mean regret was -244,612.4 ($p =$ 0.115); and for multiclass classification tasks, the mean regret was 7,316.74 ($p =$ 0.586). In all cases, the $p-$values were above the conventional threshold for significance, indicating that we fail to reject the null hypothesis that the mean regret is equal to zero.

These findings suggest that while binary classification tasks show the largest negative regret, this difference is not statistically significant and may be influenced by factors such as dataset variability or the presence of outliers. Notably, the large variability in mean regret values across task types could stem from differences in the performance metric scales (e.g., RMSE for regression vs. accuracy for classification) or extreme values in specific datasets.

\begin{figure}[h]
\begin{center}

\includegraphics[width=\linewidth]{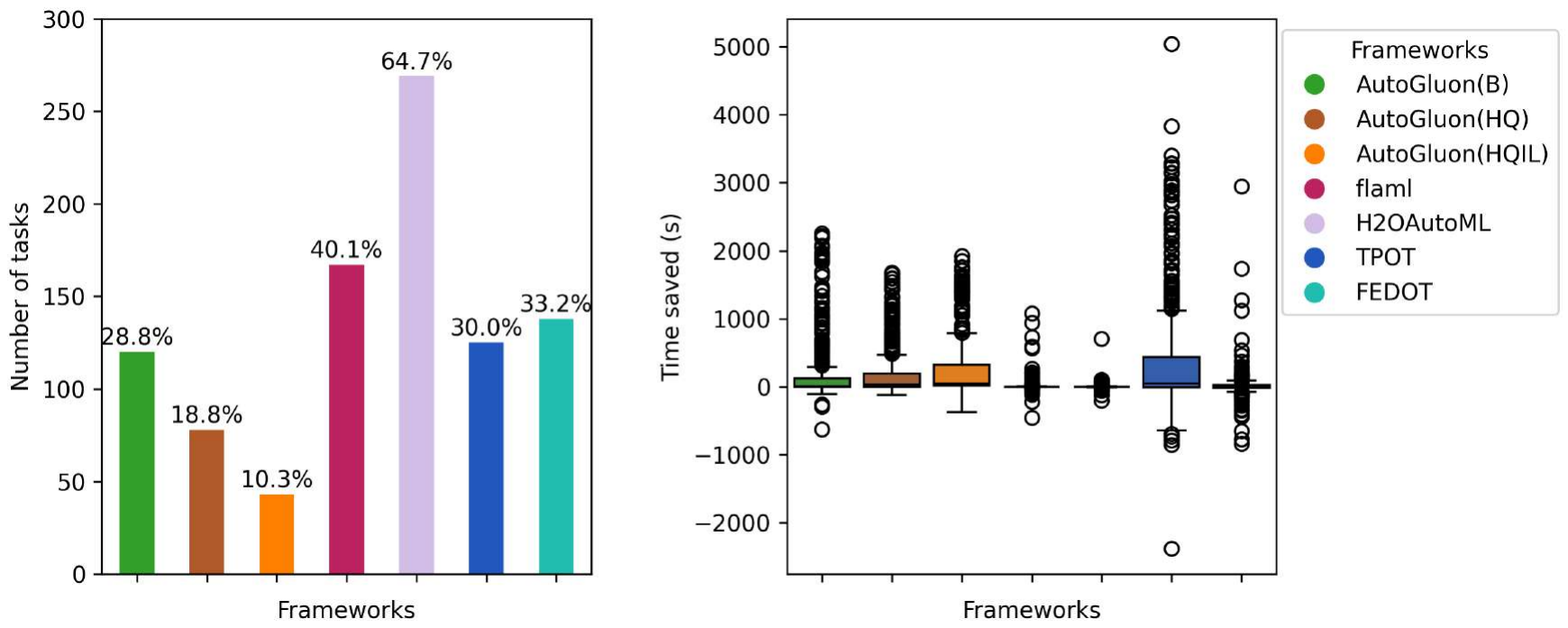}
\end{center}
\caption{Percentage of non-time saved tasks per frameworks among all constraints and distribution of \textit{time saved}}
\label{fig:negative_time_saved}
\end{figure}

Focus on \textit{time saved} (in this case we did not scale the time to have a different perspective), out of 416 tasks per each one (104 datasets $\times$ 4 time constraints). Figure \ref{fig:negative_time_saved} shows the percentage of non-time saved tasks with early stopping that took at least 1 second longer compared to the standard configuration. Early stopping saved time of up 89\% of tasks for $\mathtt{AutoGluon(HQIL)}$ and a minimum of 36\% for $\mathtt{H2OautoML}$. When looking into the distribution, $\mathtt{AutoGluon}$ saves normally around 500 seconds as an average upper limit. $\mathtt{flaml}$ and $\mathtt{H2OAutoML}$ demonstrated a minimum range between 0 and not more than 25 seconds, and $\mathtt{FEDOT}$ up to 90 seconds. $\mathtt{TPOT}$ has an upper limit above 1000 seconds, showing also the highest range of outliers. 

In the next part, first, we introduce the results of the frameworks evaluated with and without early stopping, which are $\mathtt{AutoGluon}$ (in four setups) $\mathtt{flaml}$, $\mathtt{H2OAutoML}$, $\mathtt{TPOT}$ and $\mathtt{FEDOT}$. After this, we present the results of evaluating such configurations against the rest of the frameworks regardless of their capability of early stopping. All the frameworks that contain the string ``$\mathtt{\_E}$'' mean the early stopping feature is turned on.

\subsubsection{Performance and inference time}

\begin{figure}[h!]
\begin{center}

\begin{subfigure}{0.45\textwidth}
\includegraphics[width=\linewidth]{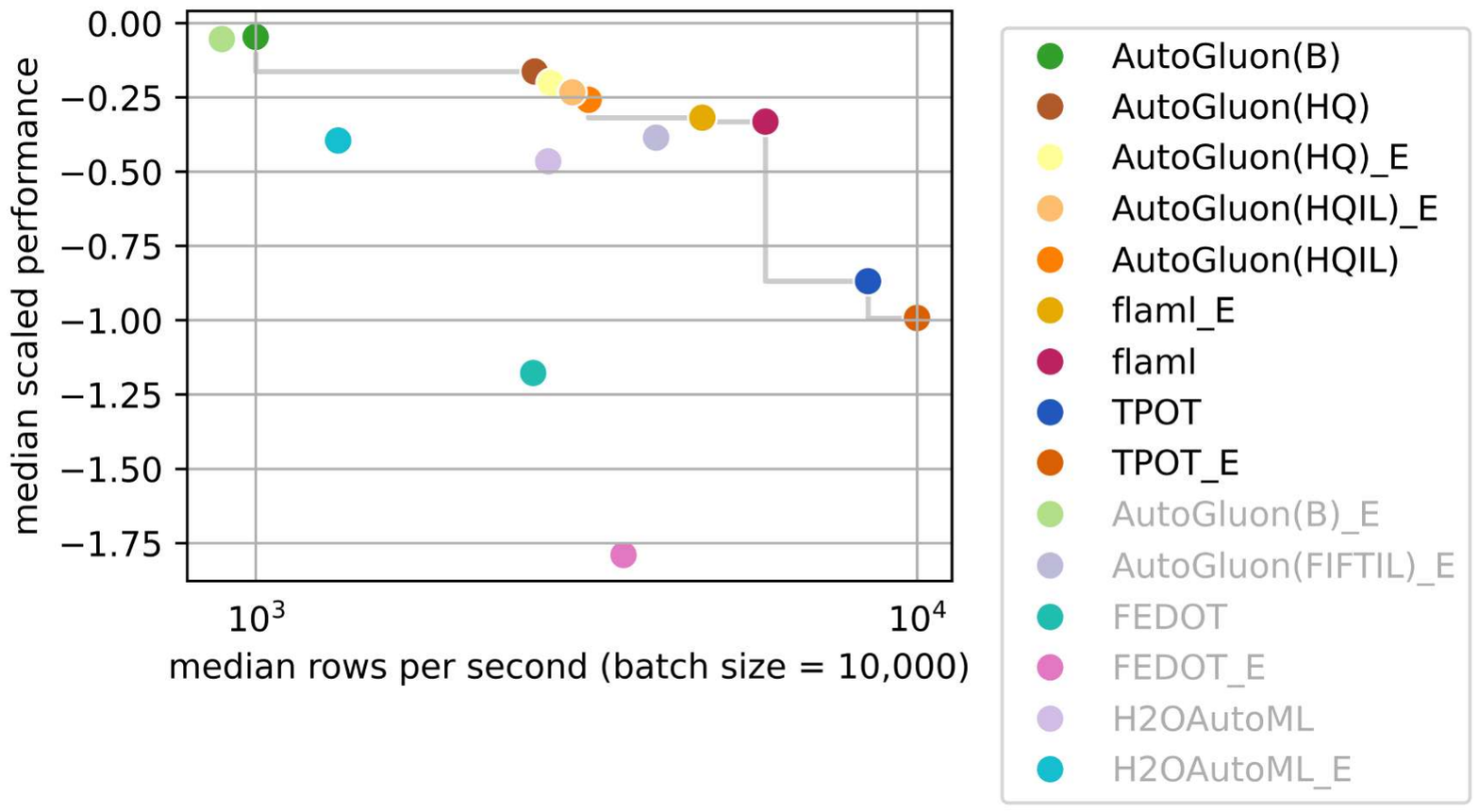}
\caption{5 minutes}
\end{subfigure}
\begin{subfigure}{0.45\textwidth}
\includegraphics[width=\linewidth]{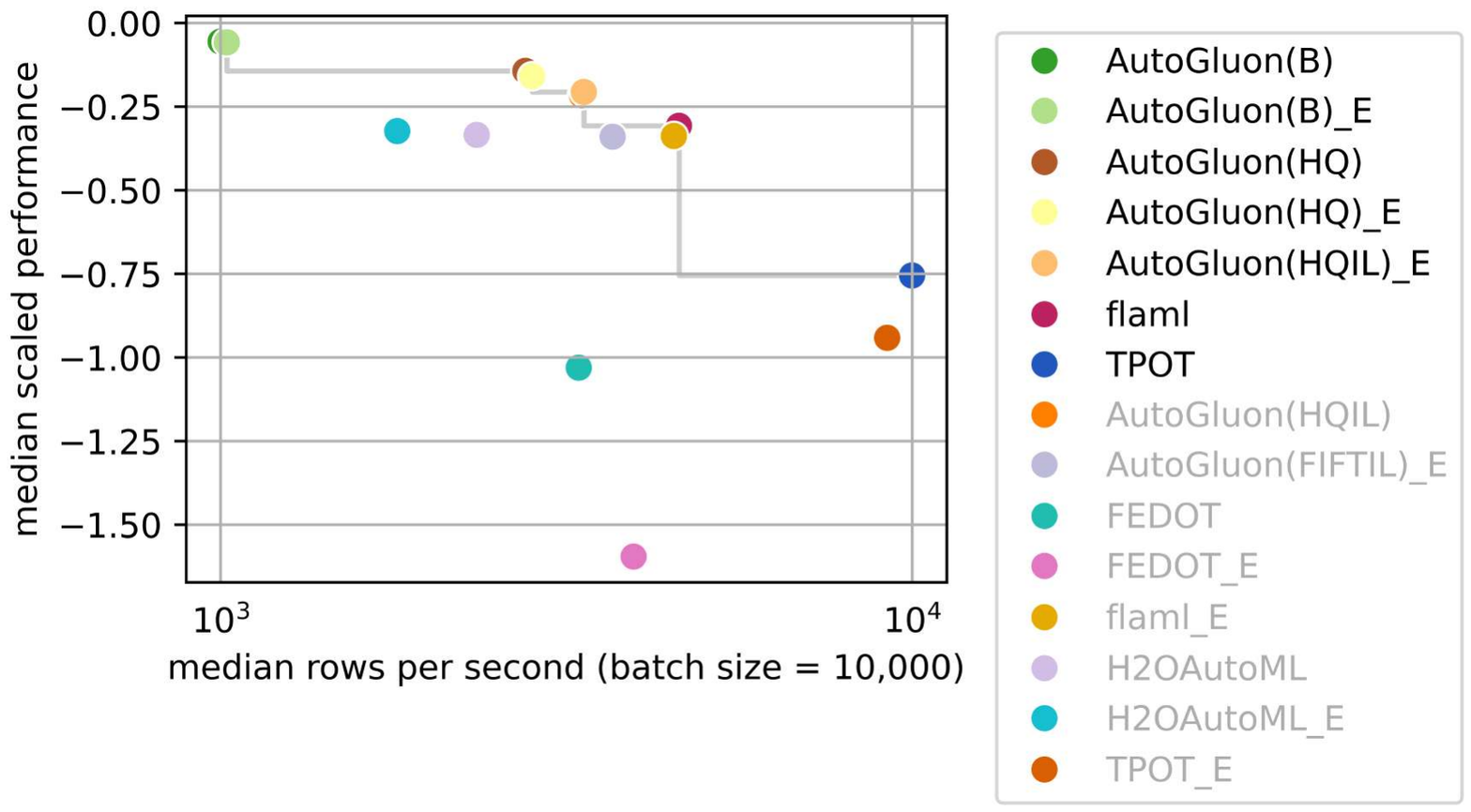}
\caption{10 minutes}
\end{subfigure}
\newline
\begin{subfigure}{0.45\textwidth}
\includegraphics[width=\linewidth]{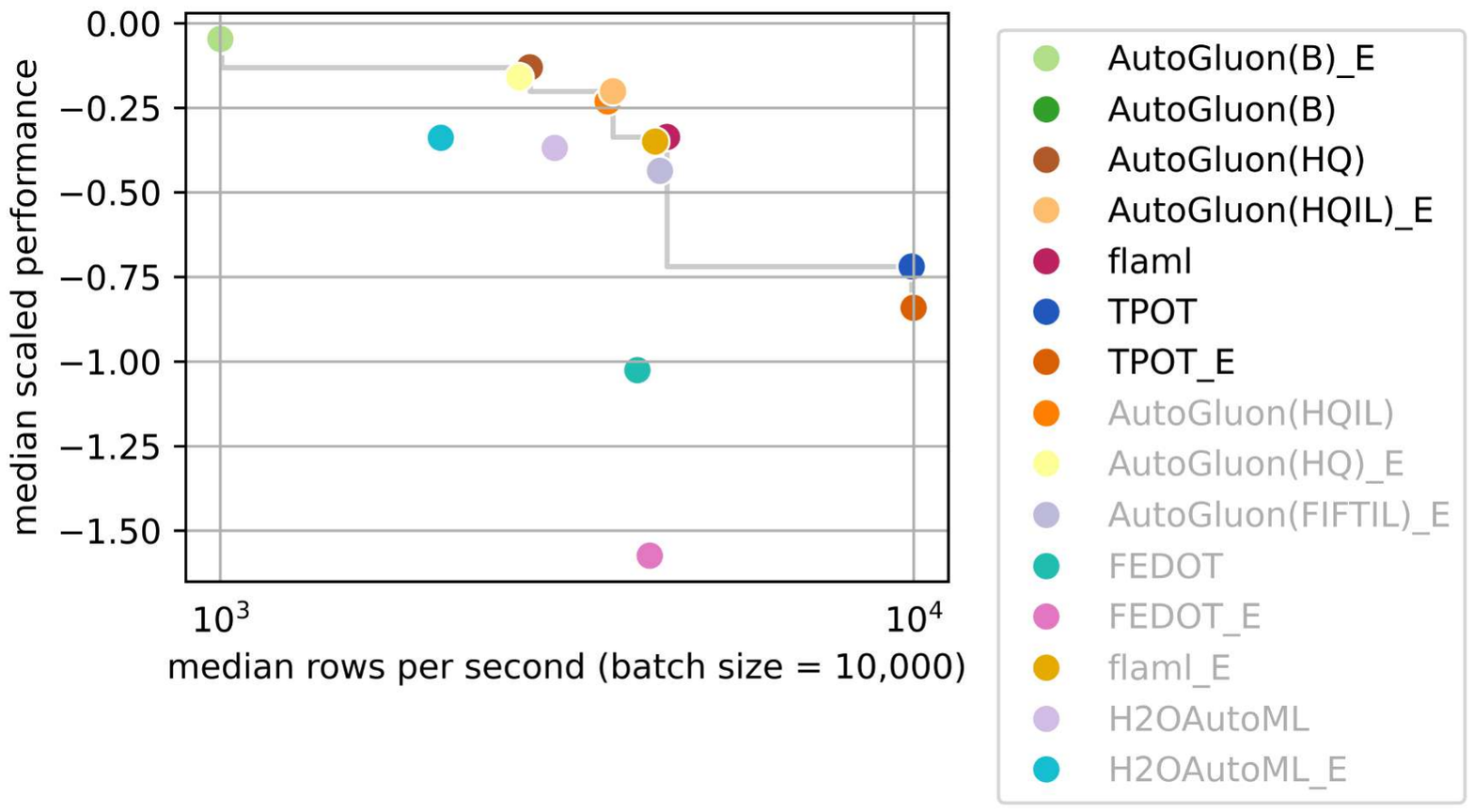}
\caption{30 minutes}
\end{subfigure}
\begin{subfigure}{0.45\textwidth}
\includegraphics[width=\linewidth]{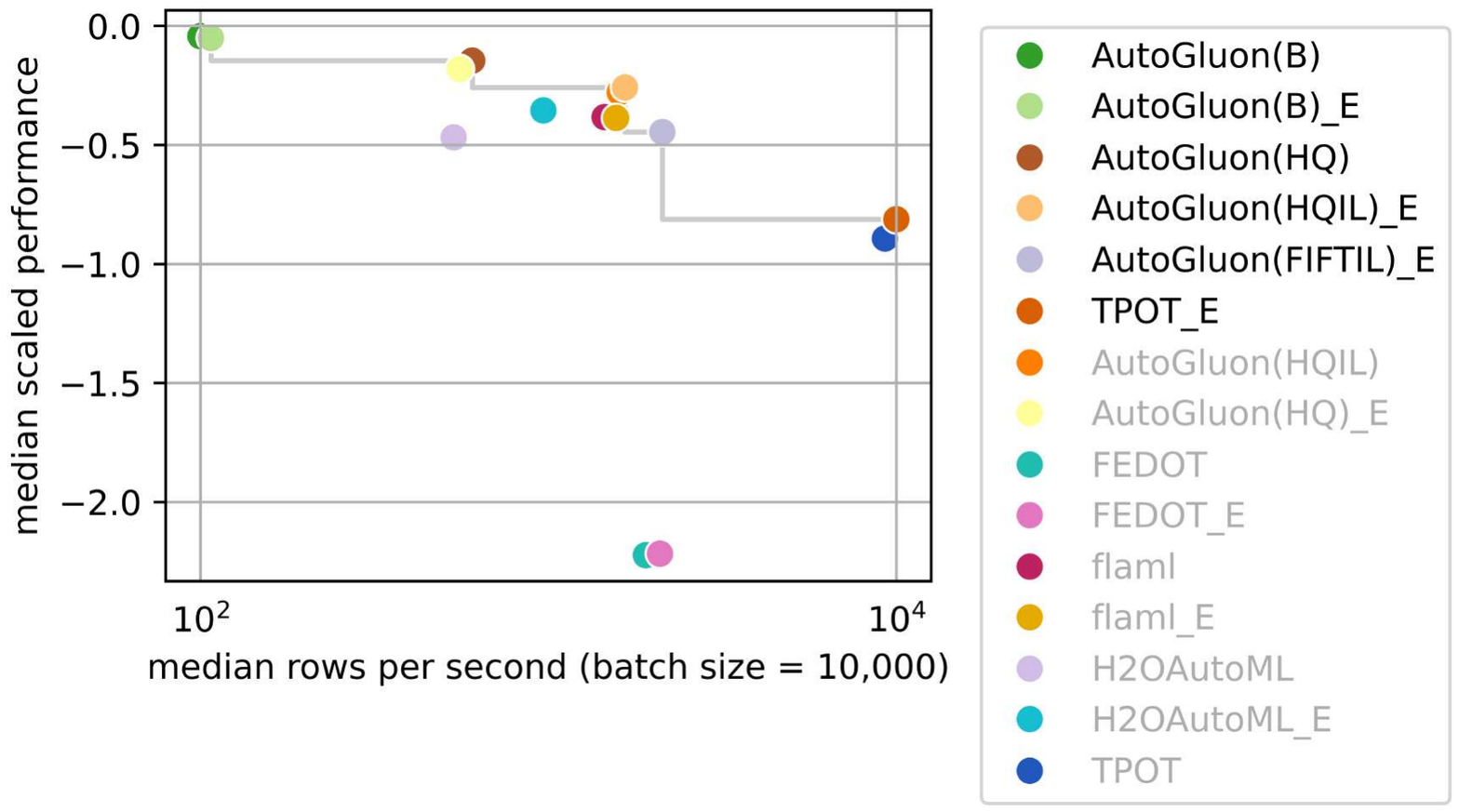}
\caption{60 minutes}
\end{subfigure}
\newline

\end{center}
\caption{Pareto Frontier of the frameworks with and without early stopping. The plots show the performance values scaled from the random forest (-1) to best observed (0) versus the inference speed in median per second with a batch size of 10,000. \label{fig:pf_early_perf_inference_times}}
\end{figure}

Firstly, we continue evaluating the scaled performance against the corresponding inference time, Figure \ref{fig:pf_early_perf_inference_times}. In the 5-minute evaluation (a), $\mathtt{AutoGluon(B)}$ shows robust performance, and $\mathtt{AutoGluon(B)\_E}$ has a slightly stronger performance. $\mathtt{AutoGluon(HQ)}$ and $\mathtt{AutoGluon(HQ)\_E}$ are the best positioned to maintain a favourable trade-off between inference speed and accuracy. It is evident that $\mathtt{flaml}$ is very competitive without sacrificing inference speed. Frameworks such as $\mathtt{FEDOT}$ and $\mathtt{H2OAutoML}$ struggle to perform efficiently in the 5-minute timeframe. 

As the time limit increases to 10 minutes (b), we observe that $\mathtt{AutoGluon(HQIL)}$ with and without early stopping showed again the best trade-off followed by $\mathtt{AutoGluon(HQ)}$. Consistently improve inference time relative to their counterparts without early stopping but sacrificing too much performance.

At 30 minutes (c), $\mathtt{flaml}$ and $\mathtt{AutoGluon}$  dominated the Pareto frontier with strong models that do not sacrifice inference time. In the 60-minute evaluation (d), $\mathtt{AutoGluon(HQIL)\_E}$ shows the best performance, which is stronger and even faster than the standard setup, indicating that early stopping might improve its ability to balance speed and accuracy. 

\begin{figure}[h]
\begin{center}
\includegraphics[width=0.7\linewidth]{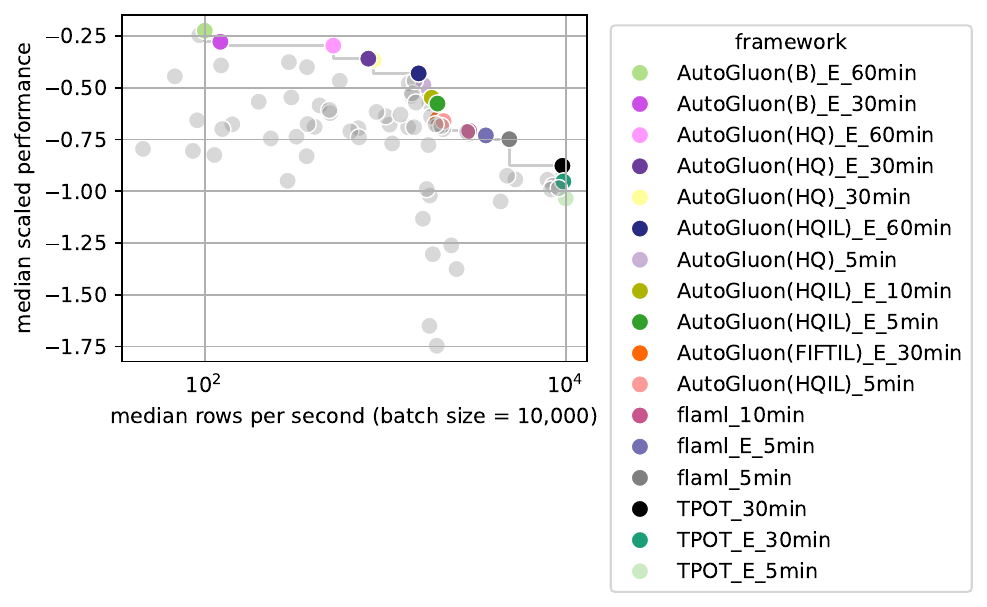}

\end{center}
\caption{Pareto Frontier of the frameworks with and without early stopping. All time constraints are joined. The plots show the performance values scaled from the random forest (-1) to best observed (0) versus the inference speed in median per second with a batch size of 10,000. 
Only the Pareto efficient methods are displayed in the legend. \label{fig:pf_early_perf_inference_times_general}}
\end{figure}

In the Figure \ref{fig:pf_early_perf_inference_times_general}, it is presented the Pareto Frontier of the frameworks with and without early stopping, joining all time constraints, the fastest models belong to $\mathtt{flaml}$ since $\mathtt{TPOT}$ is strengthened by the $\mathtt{CP}$. All models controlling the right upper are $\mathtt{AutoGluon}$ setups, especially $\mathtt{AutoGluon(HQ)}$ and $\mathtt{AutoGluon(HQIL)\_E}$ with only 10 minutes training time, even $\mathtt{AutoGluon(HQIL)}$ in 5 minutes showed one the strongest trade-off.

\subsubsection{Performance and training time}
\label{cross:appendix_performance_training_time}

\begin{figure}[h]
\centering

\begin{subfigure}{0.45\textwidth}
\includegraphics[width=\linewidth]{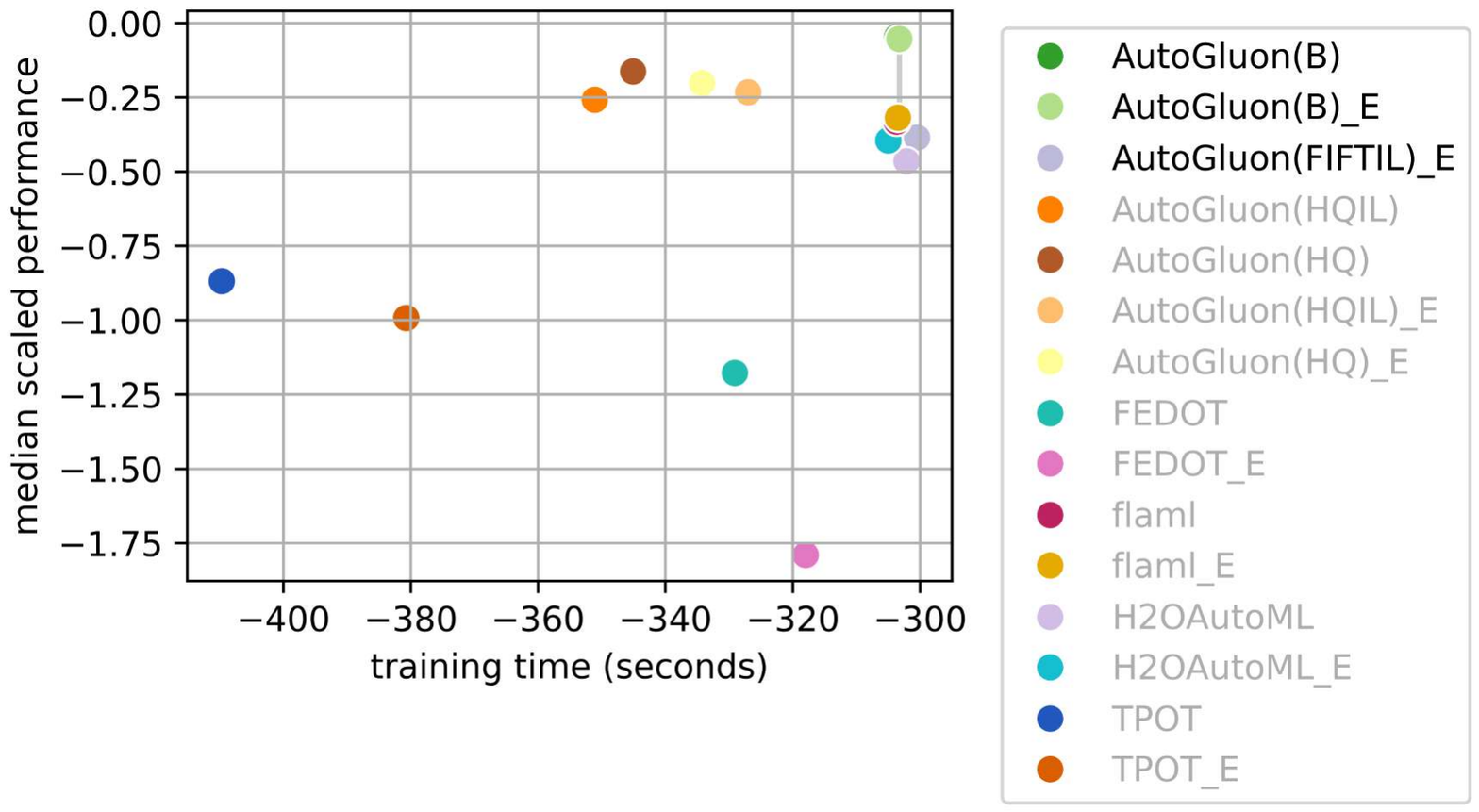}
\caption{5 minutes}
\end{subfigure}
\begin{subfigure}{0.45\textwidth}
\includegraphics[width=\linewidth]{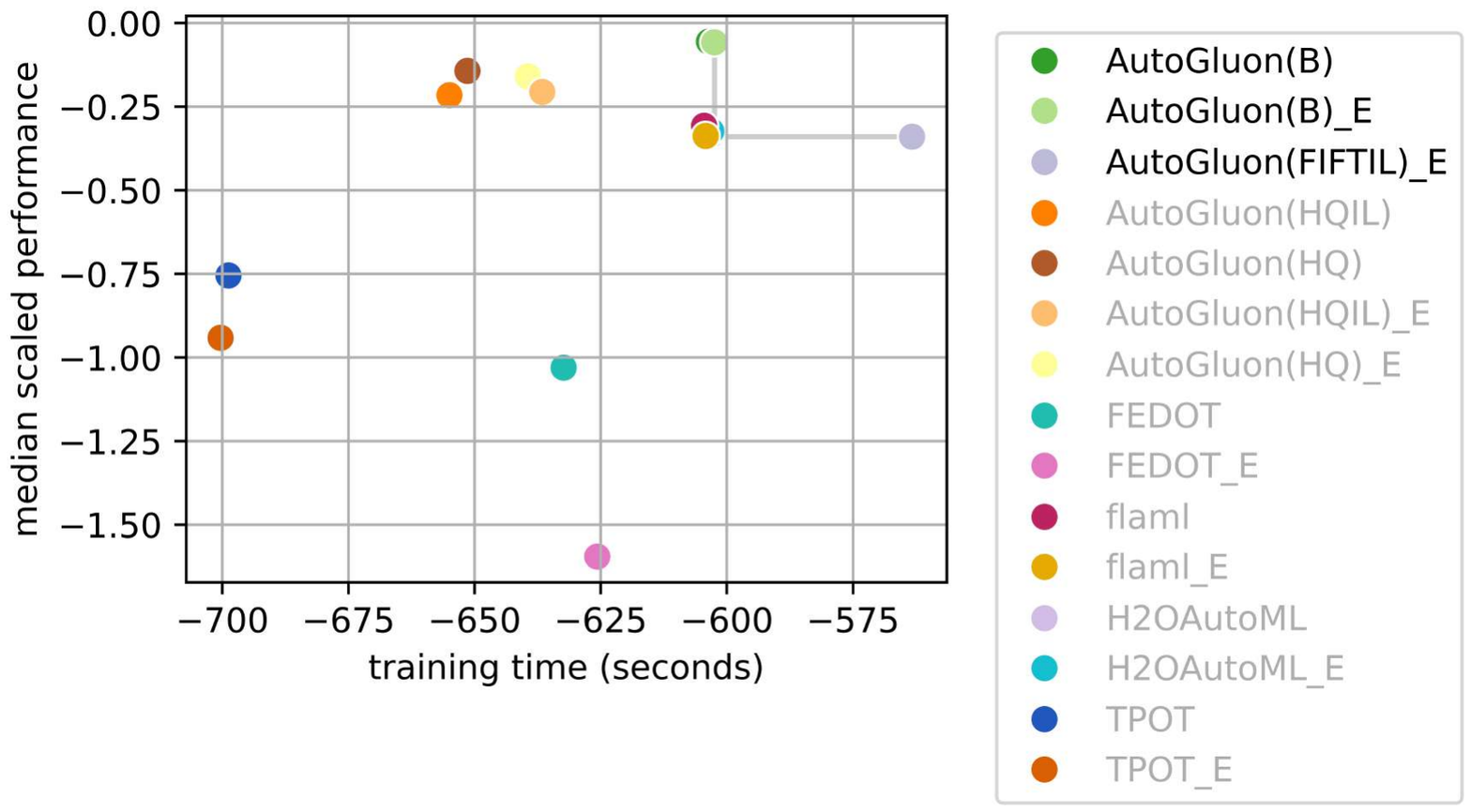}
\caption{10 minutes}
\end{subfigure}
\newline
\begin{subfigure}{0.45\textwidth}
\includegraphics[width=\linewidth]{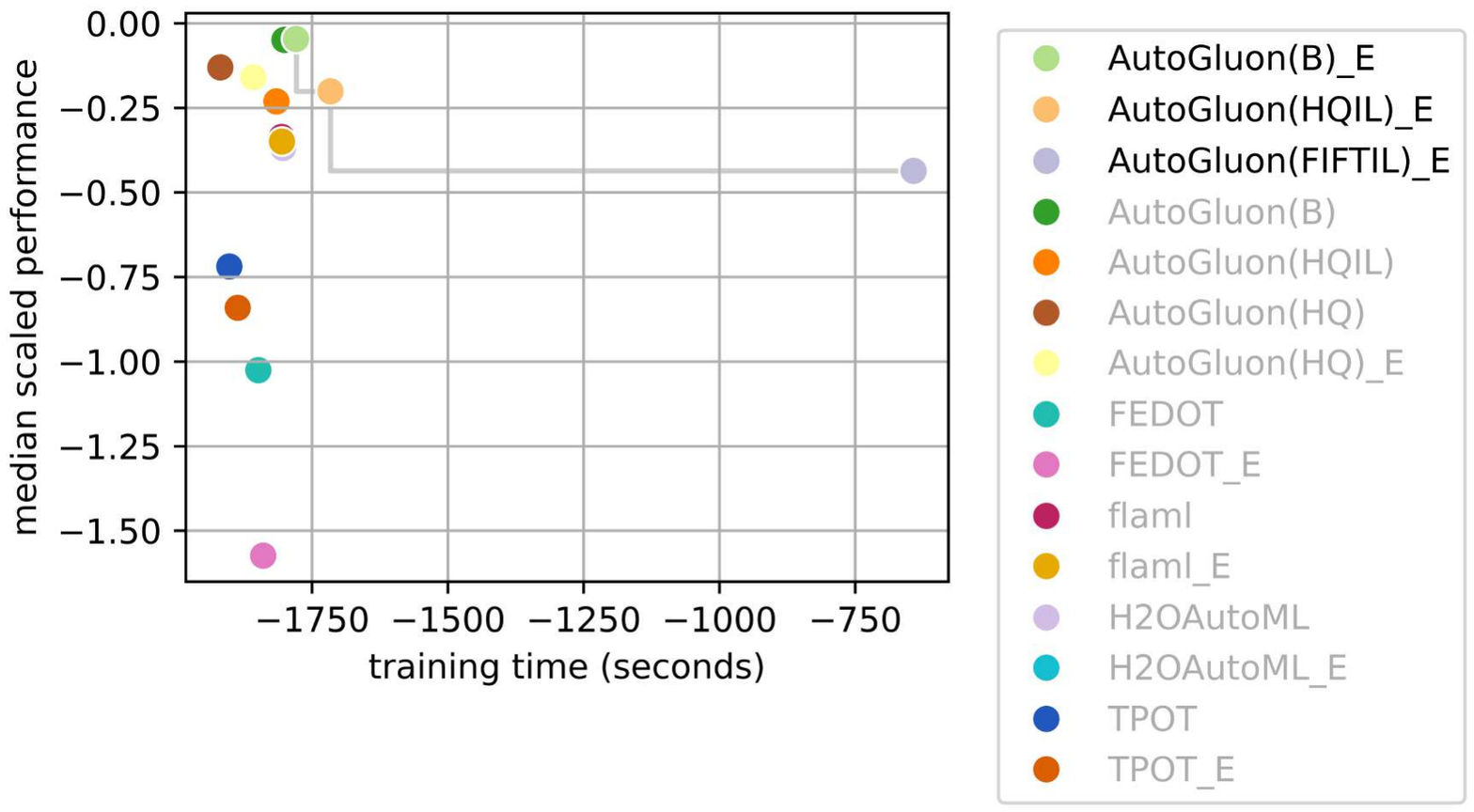}
\caption{30 minutes}
\end{subfigure}
\begin{subfigure}{0.45\textwidth}
\includegraphics[width=\linewidth]{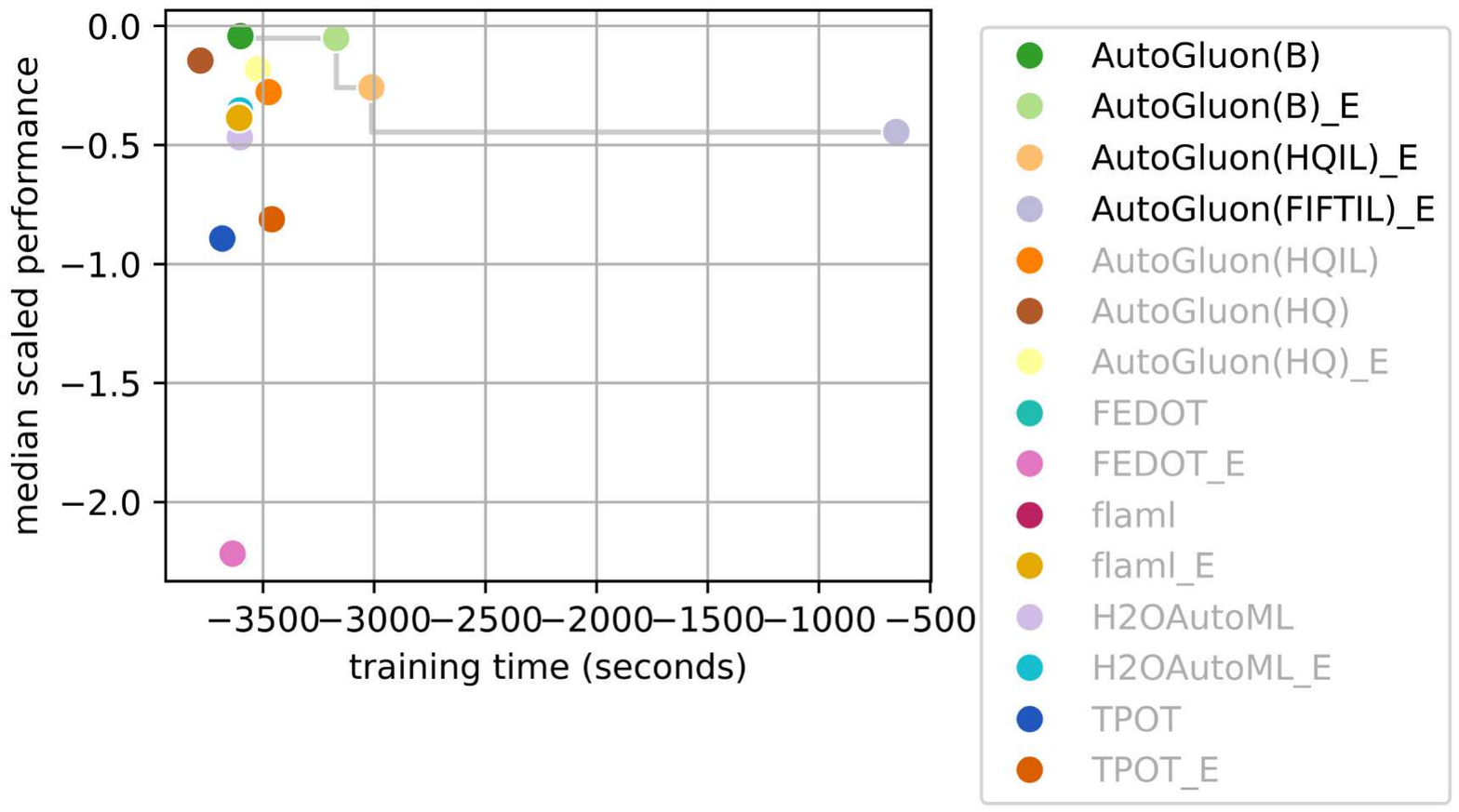}
\caption{60 minutes}
\end{subfigure}
\caption{Pareto Frontier of the frameworks with and without early stopping. The plots show the performance values scaled from the random forest (-1) to best observed (0) versus the training time. \label{fig:pf_early_perf_training_time}}
\end{figure}

In Figure \ref{fig:pf_early_perf_training_time}, we evaluate the trade-off between performance and training time. The y-axis represents the median scaled performance, where frameworks closer to 0 exhibit stronger performance, while the x-axis shows the training time, with points farther to the right indicating frameworks that required less time to achieve their results. As time constraints increase, the distribution of frameworks begins to shift, providing key insights into how much computational power is required to maintain competitive performance across different AutoML frameworks.

In the 5-minute and 10-minute scenarios (a) and (b), $\mathtt{AutoGluon(B)}$ setups demonstrate strong performance even within short time limits, but in 10 minutes setup, $\mathtt{AutoGluon(FIFTIL)}$ the early stopping capability is more visible since it did not require to use the whole time to achieve competitive results. Frameworks like $\mathtt{FEDOT}$ exhibit weaker results, particularly in the 5-minute case, suggesting that these frameworks may not be as efficient under tighter time constraints. 

As we extend the time budget to 30 minutes (c), more frameworks start converging toward better performance. We observe that frameworks such as $\mathtt{AutoGluon(B)\_E}$ and $\mathtt{AutoGluon(HQIL)\_E}$ continue to exhibit the best edge in terms of both performance and reduced training time, the early stopping of $\mathtt{AutoGluon(FIFTIL)}$ is even clearer, it stopped slightly after 10 minutes for a 30 minutes configuration and still recording stronger results than $\mathtt{flaml}$, $\mathtt{TPOT}$ and $\mathtt{FEDOT}$.

In the 60-minute scenario (d), $\mathtt{AutoGluon(B)}$ and $\mathtt{AutoGluon(B)\_E}$ showed the best performance training everything in 1 hour or less but $\mathtt{AutoGluon(FIFTIL)}$ is even faster than earlier, it did not use half of the time provided it and still is almost as competitive as $\mathtt{H2OAutoML}$ and stronger than $\mathtt{TPOT}$ and $\mathtt{FEDOT}$. This shows that certain frameworks benefit from early stopping, making them ideal choices when computational resources are limited, as they can achieve comparable results without fully utilizing the allocated time.

\begin{figure}[h]
\begin{center}
\includegraphics[width=0.7\linewidth]{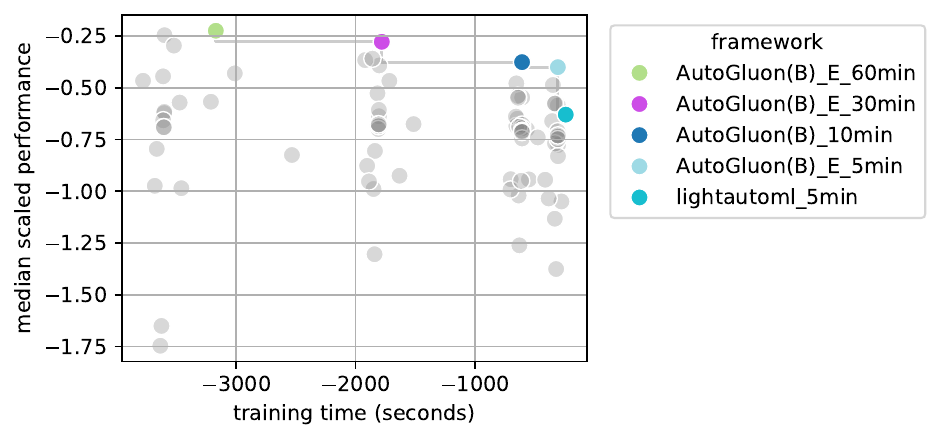}

\end{center}
\caption{Pareto Frontier of all the AutoML systems with and without early stopping. The plots show the performance values scaled from the random forest (-1) to best observed (0) versus the training time. Only the Pareto efficient methods are displayed in the legend.  \label{fig:pf_early_perf_training_time_general}}
\end{figure}

Figure \ref{fig:pf_early_perf_training_time_general} presents the Pareto Frontier of the frameworks independently if early stopping is available or not; the goal is to join all time constraints across all setups. The plot can be seen split into three vertical clusters, such as the time constraints, but 5 and 10 minutes function as a single cluster since its difference is smaller. In all four scenarios, $\mathtt{AutoGluon(B)}$ with and without not early stopping shows its power as the framework with the best performance and training time and $\mathtt{lightautoml}$ shows potential in training time while keeping competitive performance.

\subsubsection{Training time and inference time}

\begin{figure}[h]
\begin{center}

\begin{subfigure}{0.45\textwidth}
\includegraphics[width=\linewidth]{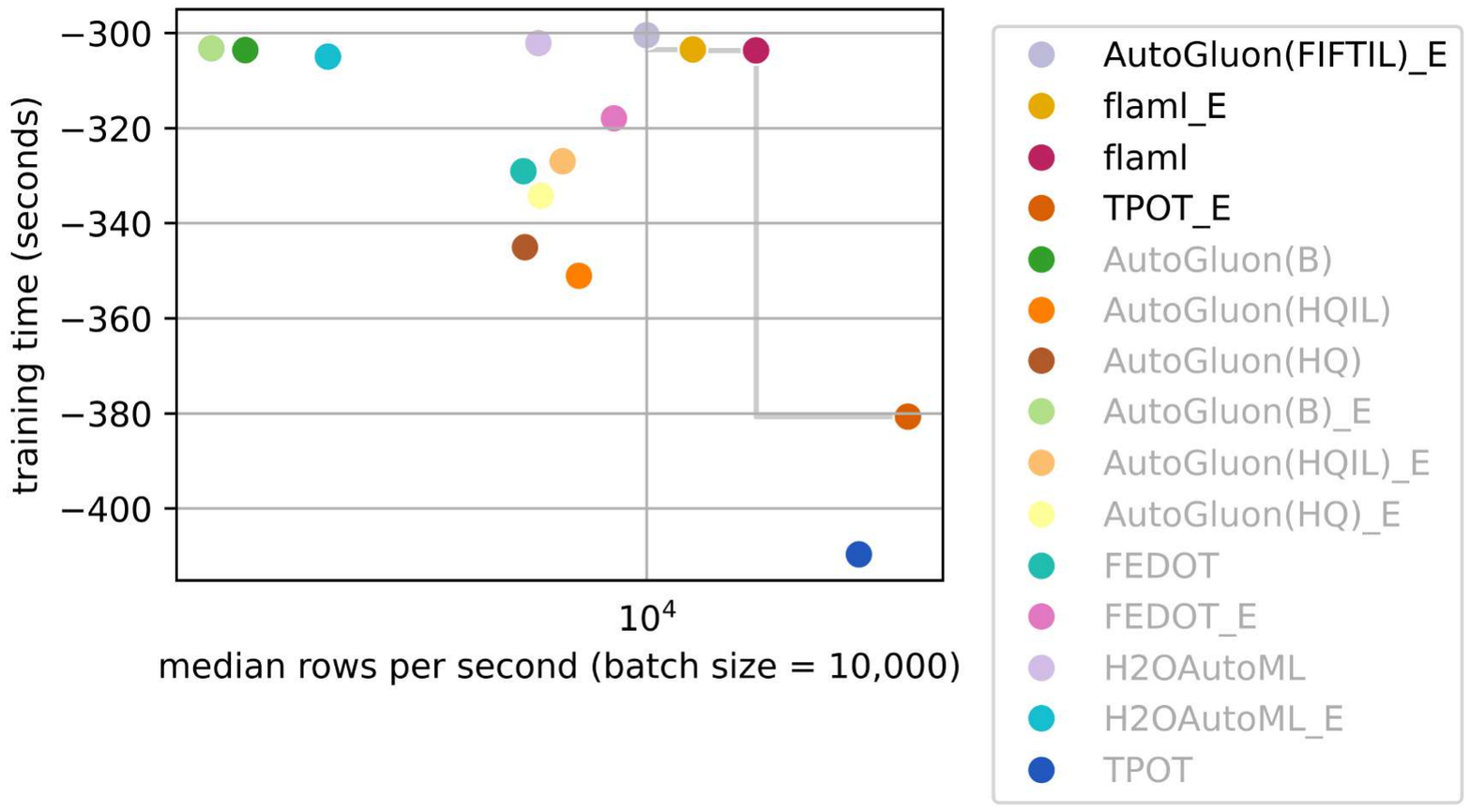}
\caption{5 minutes}
\end{subfigure}
\begin{subfigure}{0.45\textwidth}
\includegraphics[width=\linewidth]{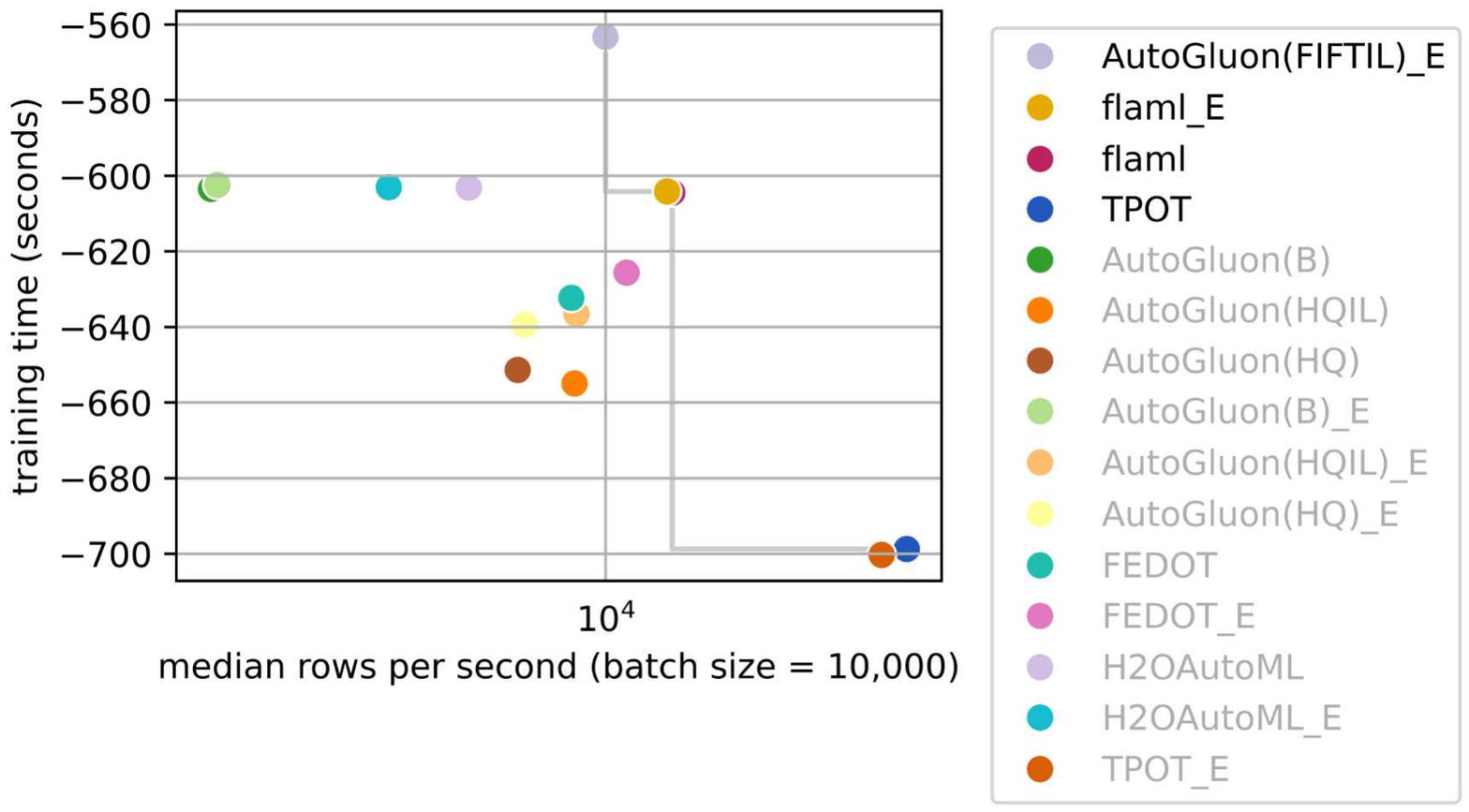}
\caption{10 minutes}
\end{subfigure}
\newline
\begin{subfigure}{0.45\textwidth}
\includegraphics[width=\linewidth]{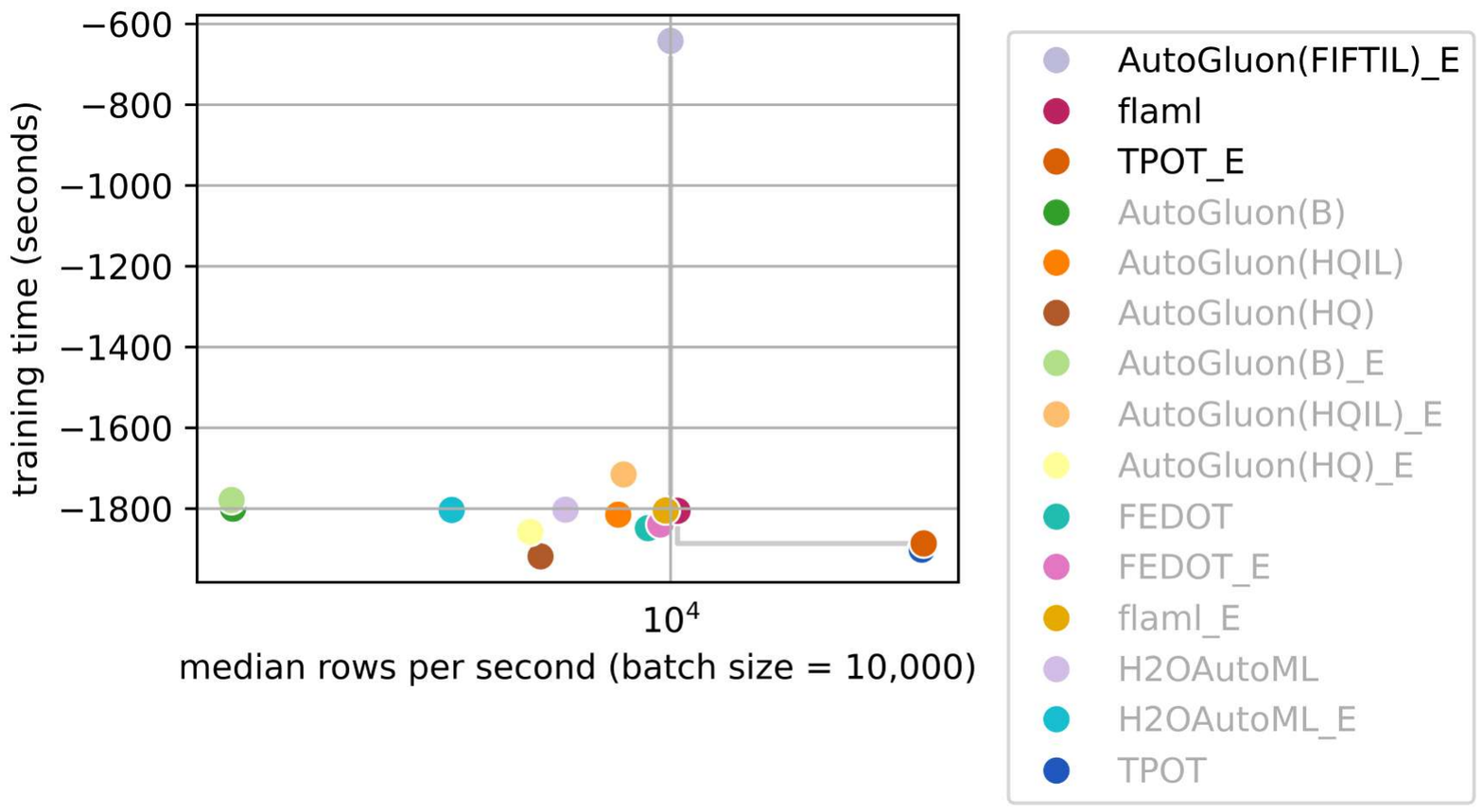}
\caption{30 minutes}
\end{subfigure}
\begin{subfigure}{0.45\textwidth}
\includegraphics[width=\linewidth]{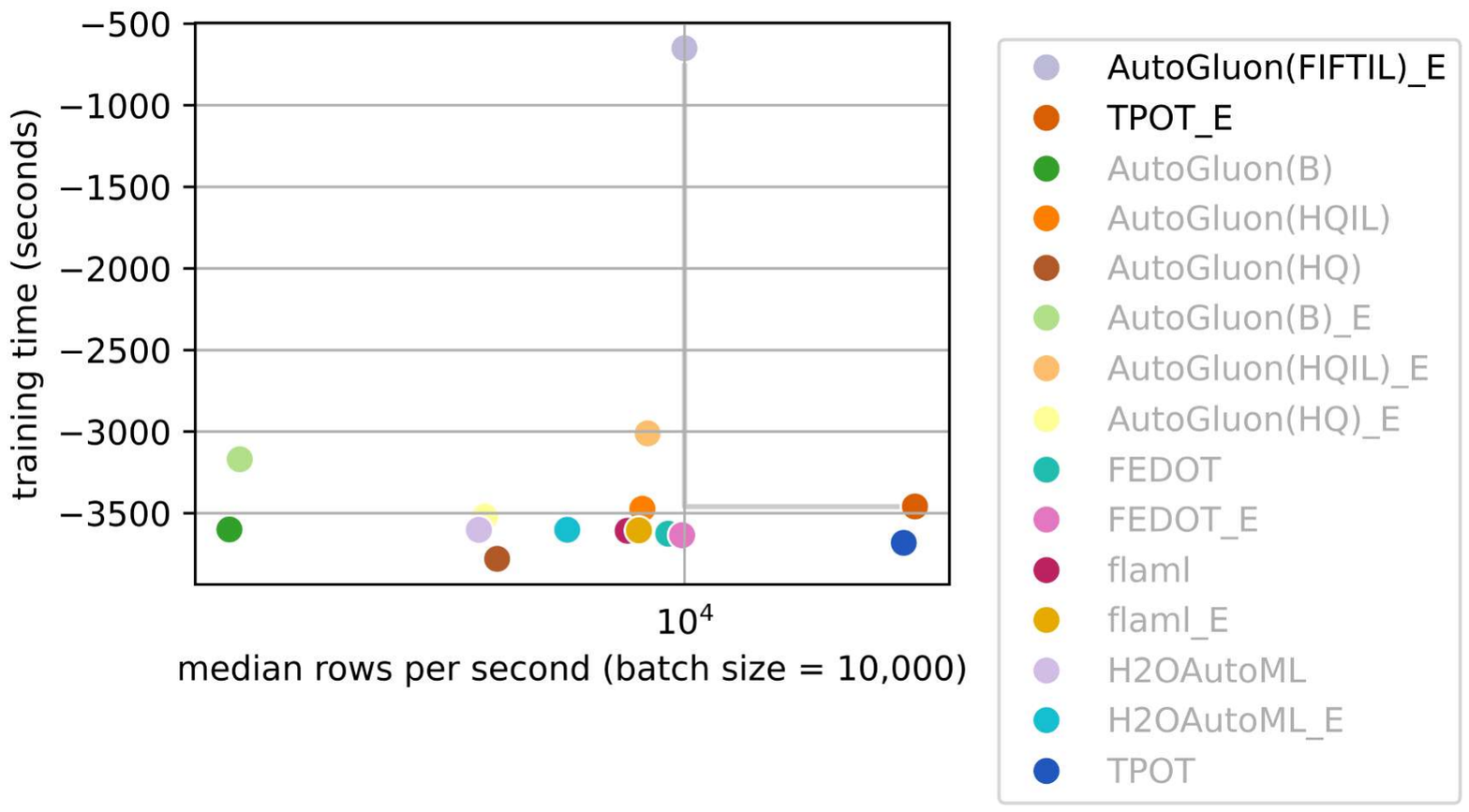}
\caption{60 minutes}
\end{subfigure}
\newline

\end{center}
\caption{Pareto Frontier of the frameworks with and without early stopping. All time constraints are joined. The plots show the inference speed in median per second with a batch size of 10,000 versus the training time. \label{fig:pf_early_inference_training_time}}
\end{figure}

Additionally, in Figure \ref{fig:pf_early_inference_training_time}, we evaluate the inference time, x-axis, where the higher and more to the right, the better, and the y-axis denotes the negative training time (the closer to 0, the faster). In the 5-minute evaluation (a), $\mathtt{AutoGluon(FIFITL)\_E}$ stands out for its efficiency in training, showing an acceptable trade-off with reduced inference speeds. $\mathtt{flaml\_E}$ and $\mathtt{flaml}$ exhibit solid and the best performance in this case.

About the 10-minute evaluation (b), we observe that $\mathtt{flaml\_E}$ continues to demonstrate a favourable balance between training and inference. In 30 minutes (c), the relative trade-offs become more pronounced only $\mathtt{AutoGluon(FIFITL)\_E}$ demonstrate the fastest training times, using $\sim$10 minutes and still being faster or as fast as all the frameworks in inference.

Finally, at the 60-minute evaluation (d), we observe that the differences between frameworks become less pronounced, $\mathtt{TPOT\_E}$ maintains its ``relatively'' fast inference speed but incurs a high training cost. Meanwhile, $\mathtt{AutoGluon(FIFITL)\_E}$ emerges as the strongest in terms of training time and inference time, not needing more than 10 minutes for a 60-minute task.

\clearpage
\subsection{Extra analysis performance and inference time per AutoML system}

\begin{figure}[h]
\begin{center}

\begin{subfigure}{0.31\textwidth}
\includegraphics[width=\linewidth]{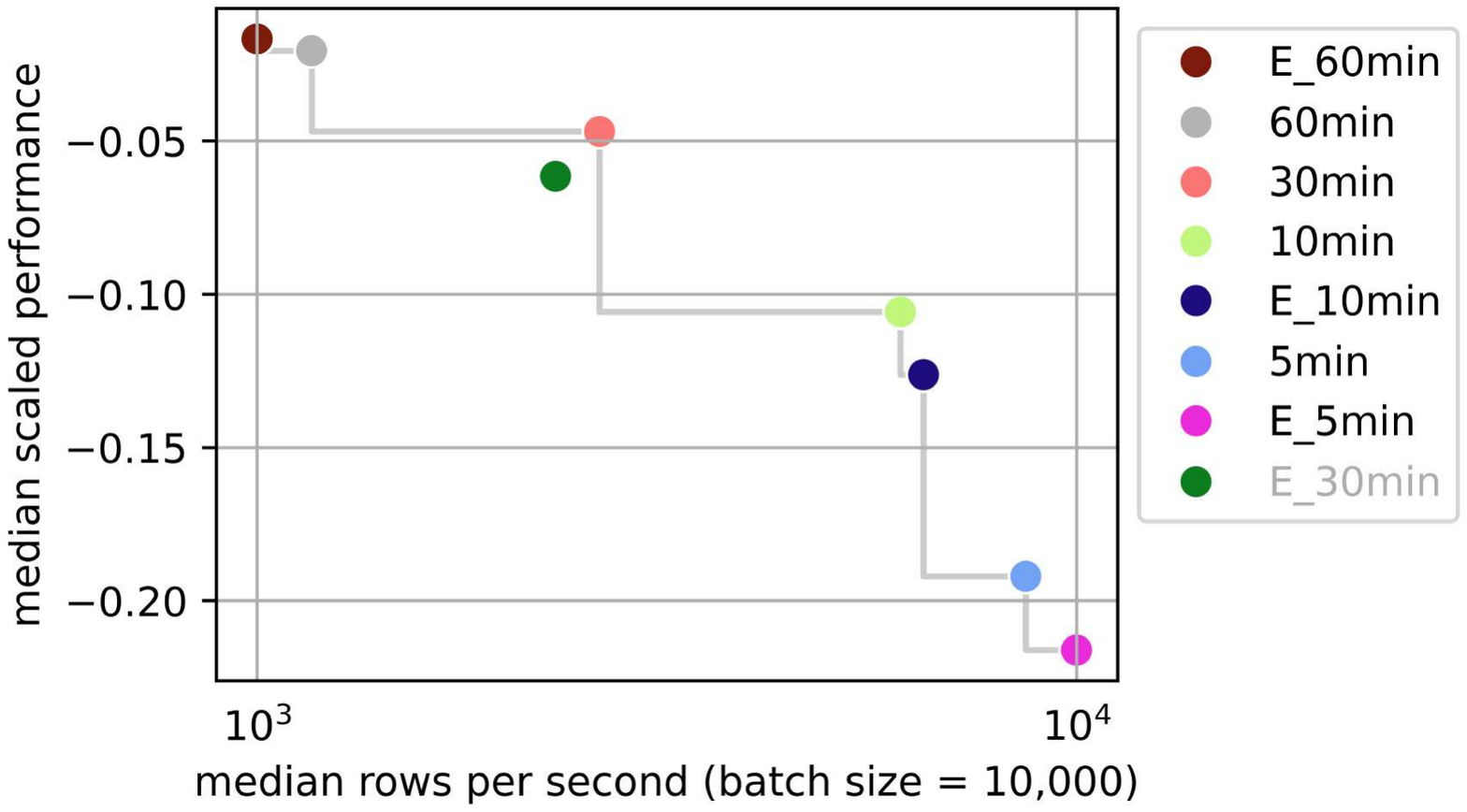}
\caption{\tiny AutoGluon(HQ)}
\end{subfigure}
\begin{subfigure}{0.31\textwidth}
\includegraphics[width=\linewidth]{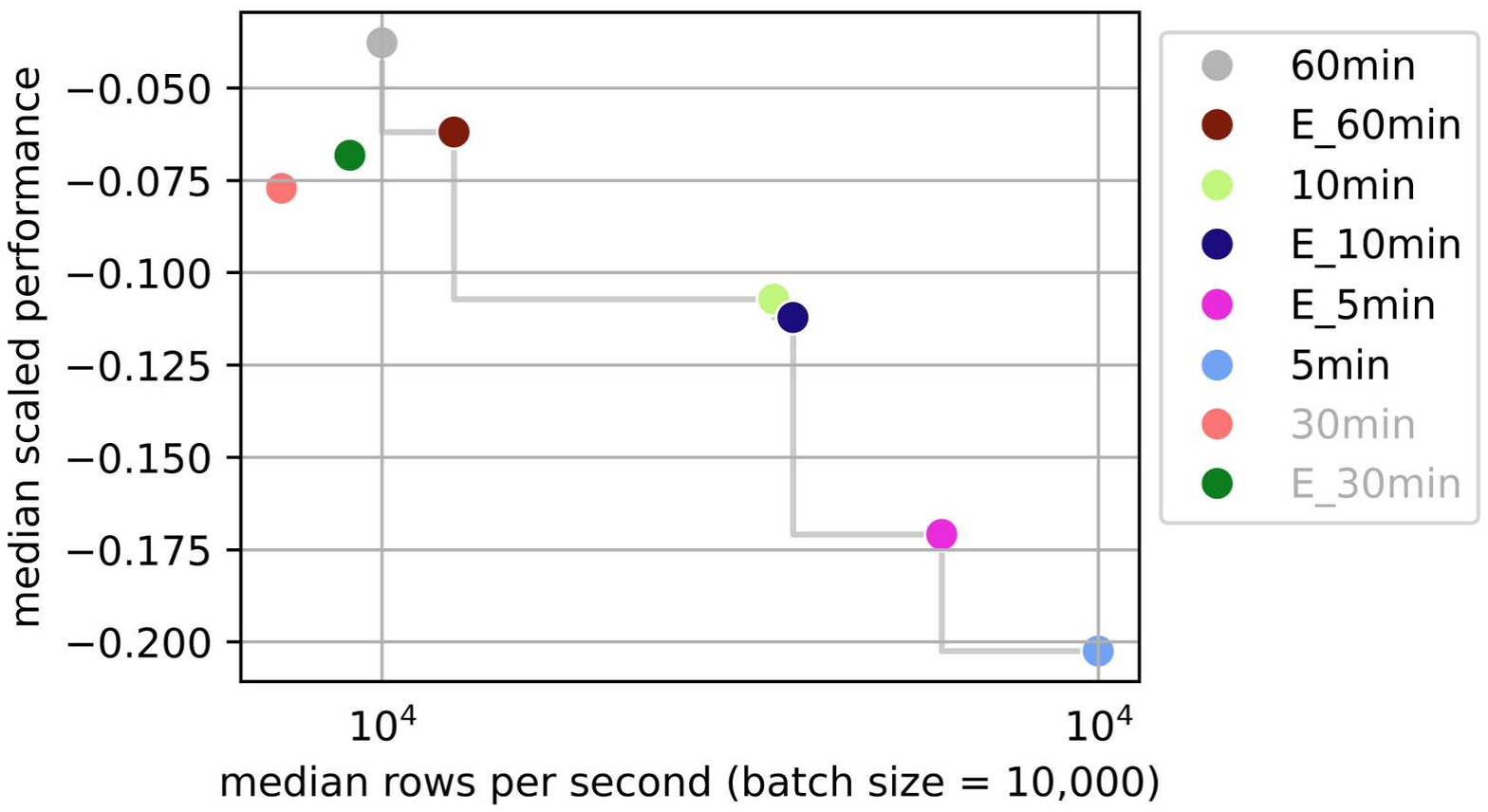}
\caption{\tiny AutoGluon(HQIL)}
\end{subfigure}
\begin{subfigure}{0.31\textwidth}
\includegraphics[width=\linewidth]{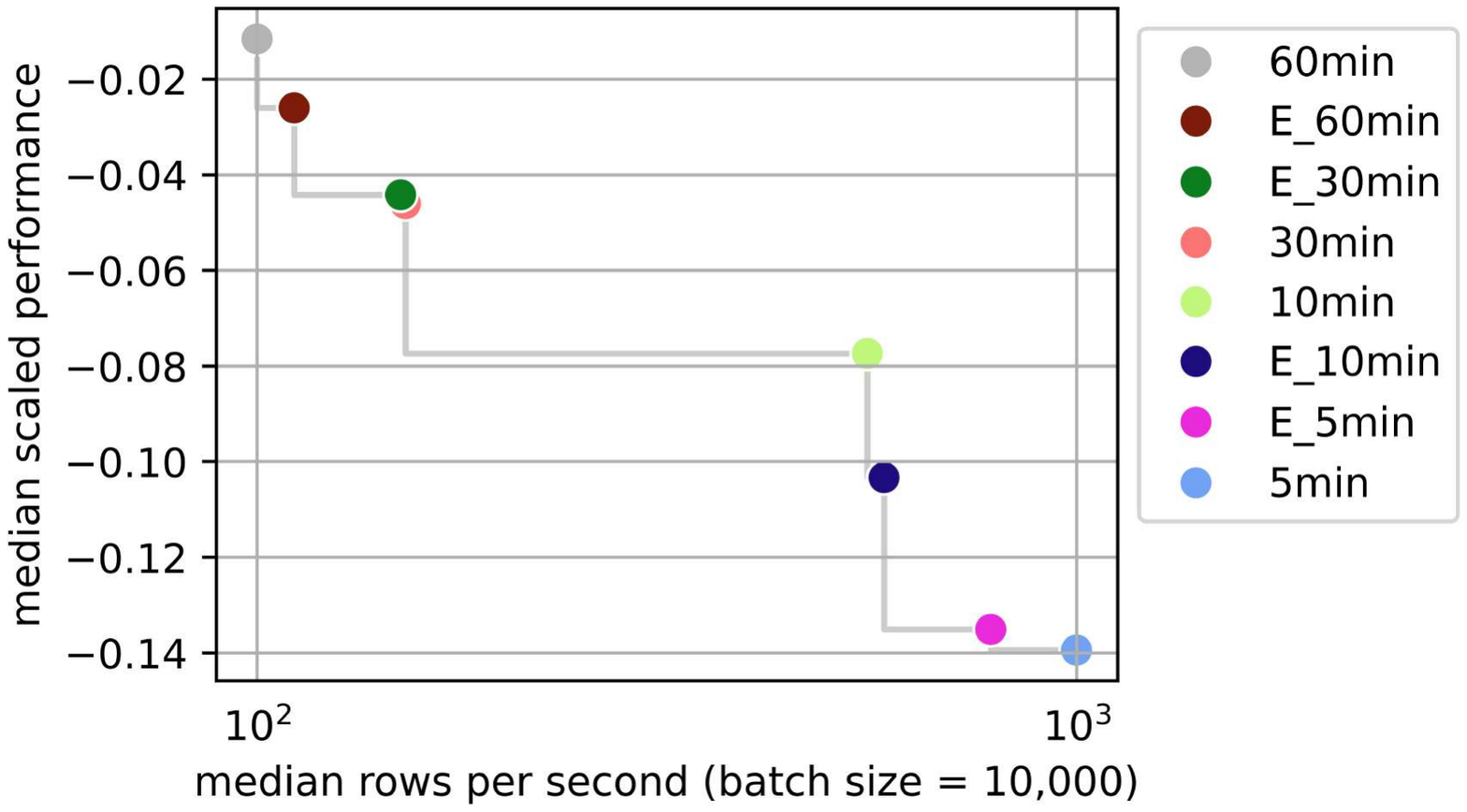}
\caption{\tiny AutoGluon(B)}
\end{subfigure}
\newline
\begin{subfigure}{0.31\textwidth}
\includegraphics[width=\linewidth]{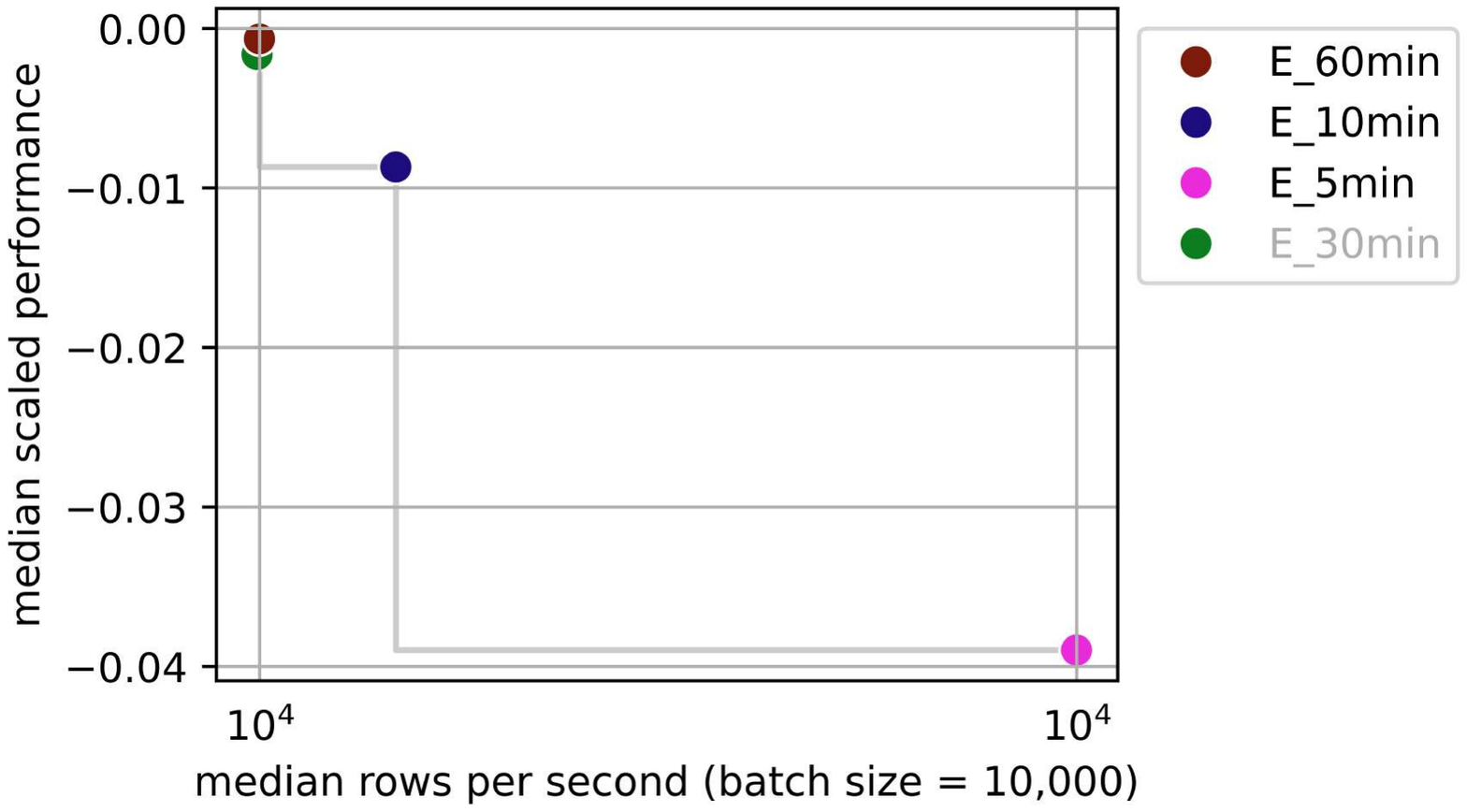}
\caption{\tiny AutoGluon(FIFTIL)}
\end{subfigure}
\begin{subfigure}{0.31\textwidth}
\includegraphics[width=\linewidth]{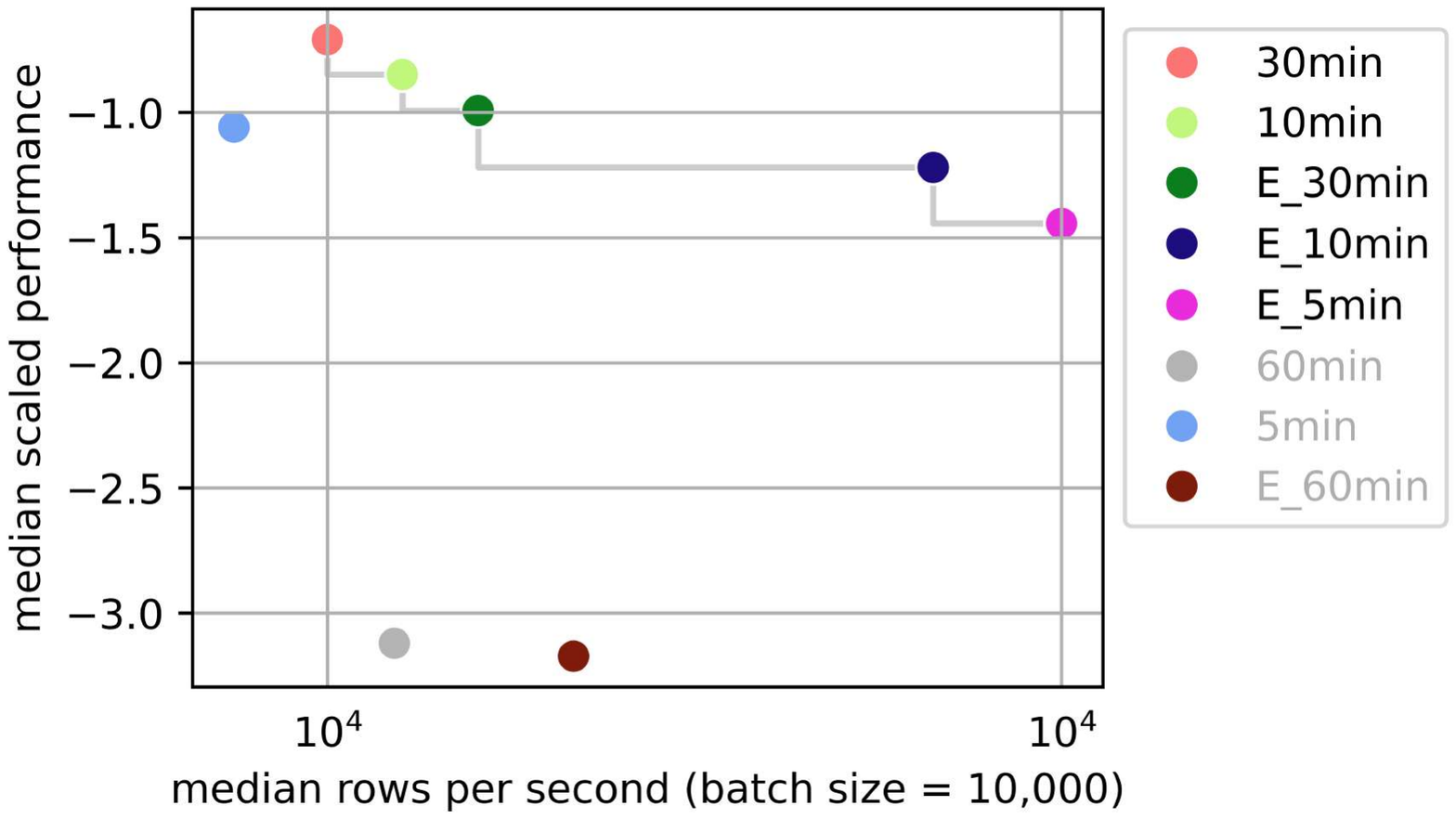}
\caption{\tiny FEDOT}
\end{subfigure}
\begin{subfigure}{0.31\textwidth}
\includegraphics[width=\linewidth]{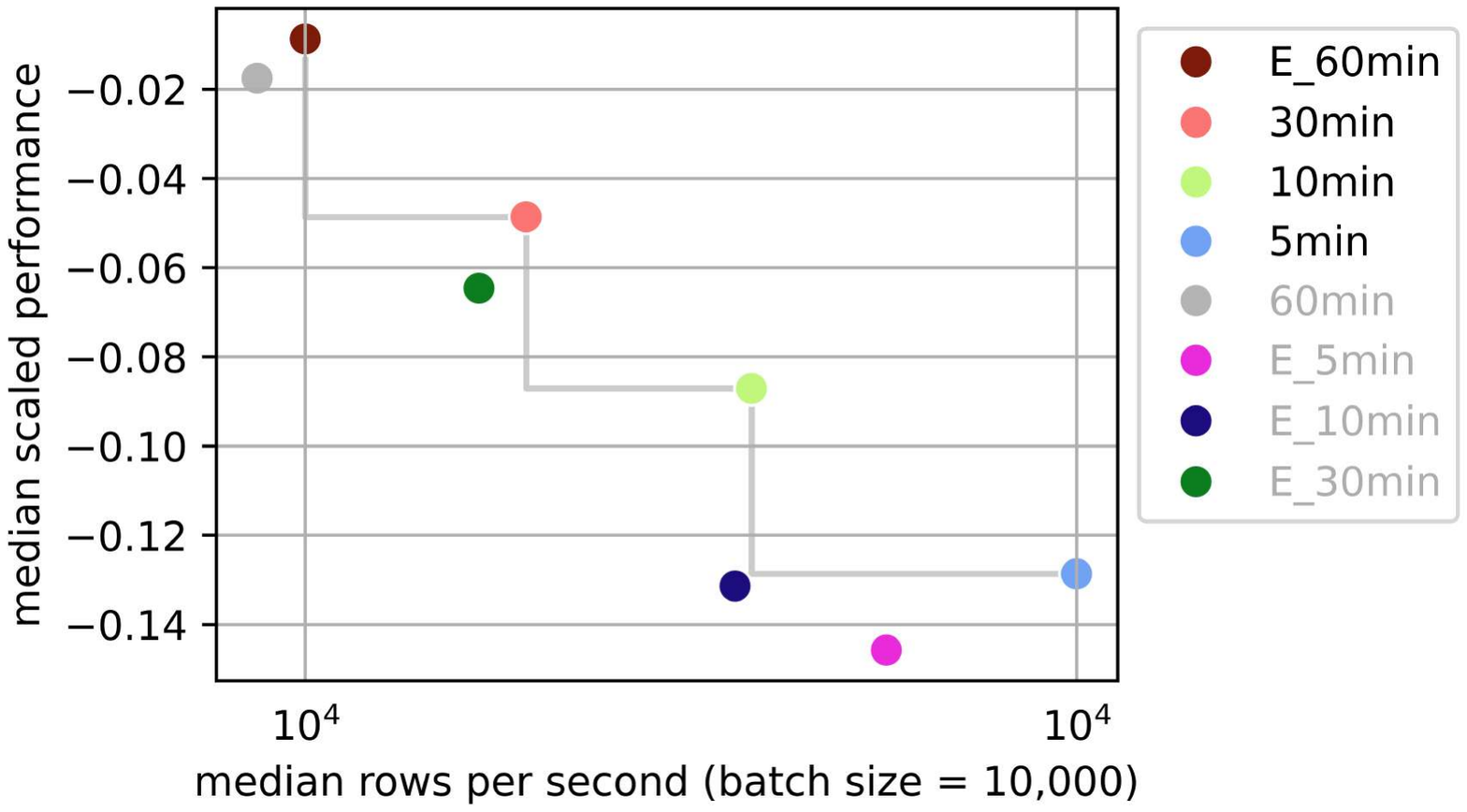}
\caption{\tiny flaml}
\end{subfigure}
\newline
\begin{subfigure}{0.31\textwidth}
\includegraphics[width=\linewidth]{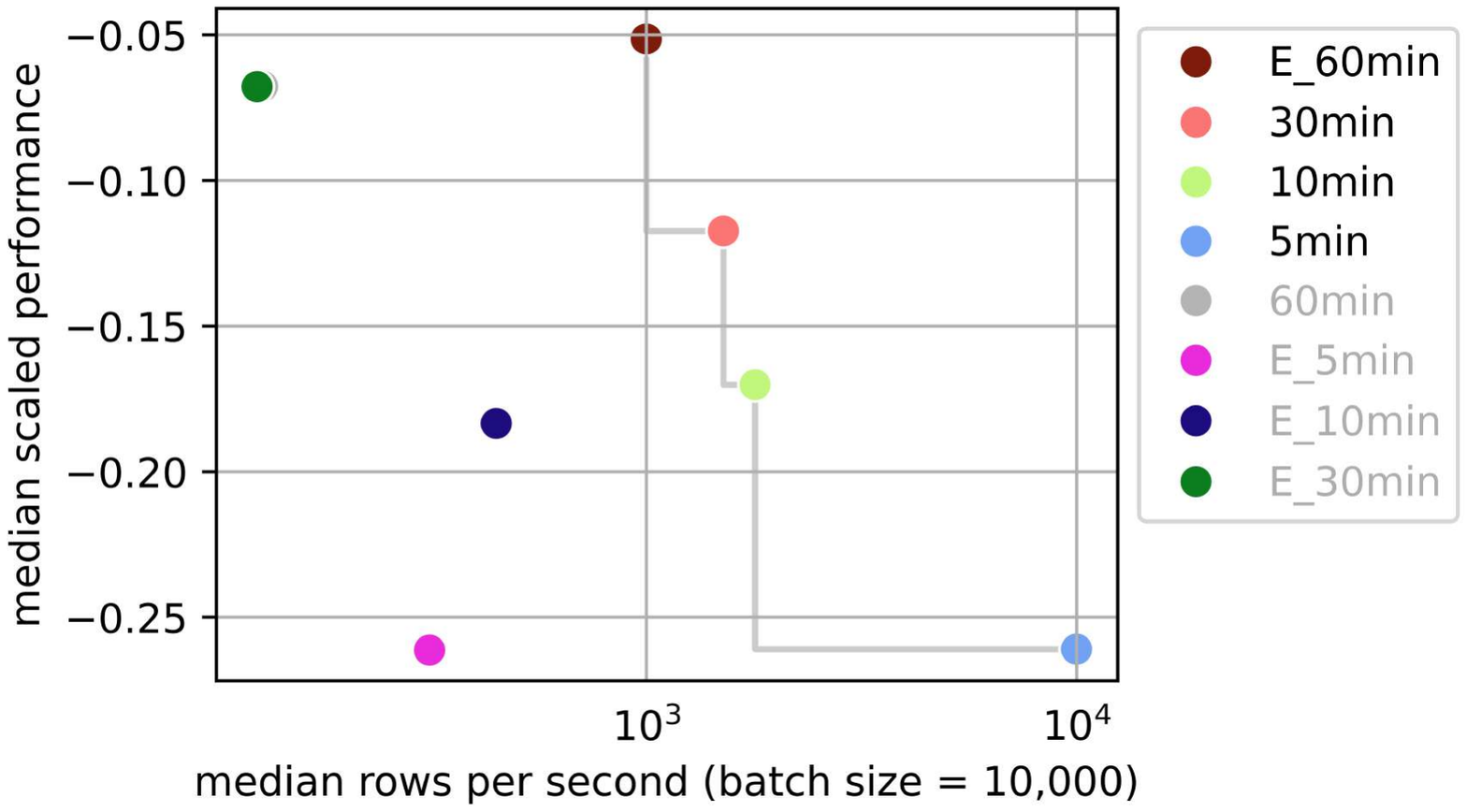}
\caption{\tiny H2OAutoML}
\end{subfigure}
\begin{subfigure}{0.31\textwidth}
\includegraphics[width=\linewidth]{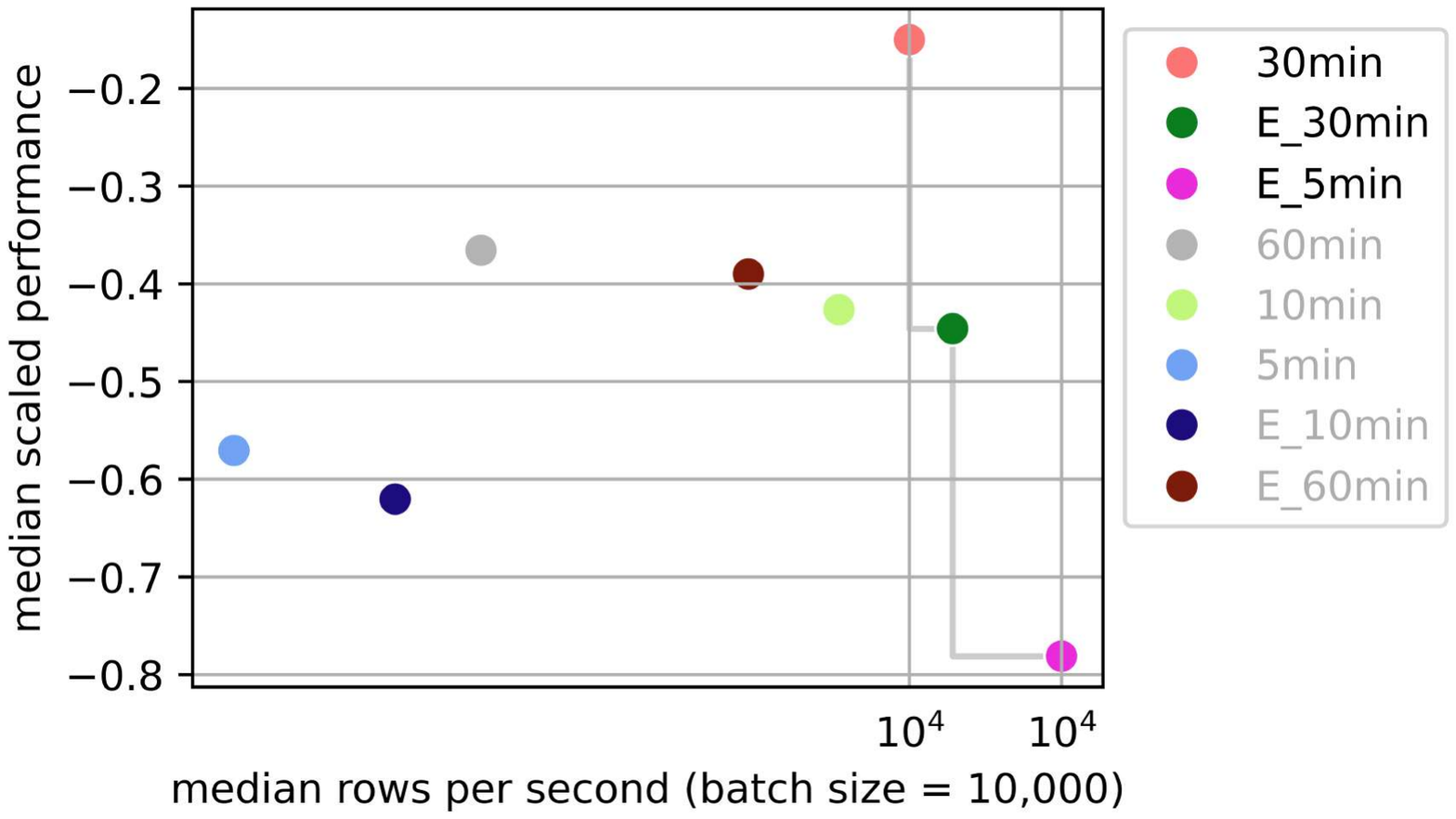}
\caption{\tiny TPOT}
\end{subfigure}

\end{center}
\caption{Pareto Frontier of the frameworks with and without early stopping. The plots show the performance values scaled from the random forest (-1) to best observed (0) versus the inference speed in median per second with a batch size of 10,000. \label{fig:appendix_pf_early_perf_inference_times}}
\end{figure}

Figure \ref{fig:appendix_pf_early_perf_inference_times} compares performance versus inference time for the frameworks that allow early stopping. It can be seen that those capable of achieving faster inference times typically do so at the cost of performance. This trade-off suggests that optimizing for quick predictions often requires sacrificing model accuracy or generalization. Frameworks that prioritize fast inference are likely to reduce complexity in their model architectures or opt for less resource-intensive algorithms, which could lead to lower overall performance in terms of predictive power. This trade-off is particularly critical when frameworks operate under stringent time constraints, as some prioritize rapid decision-making over model quality to meet real-time needs.

\begin{figure}[h]
\begin{center}

\begin{subfigure}{0.31\textwidth}
\includegraphics[width=\linewidth]{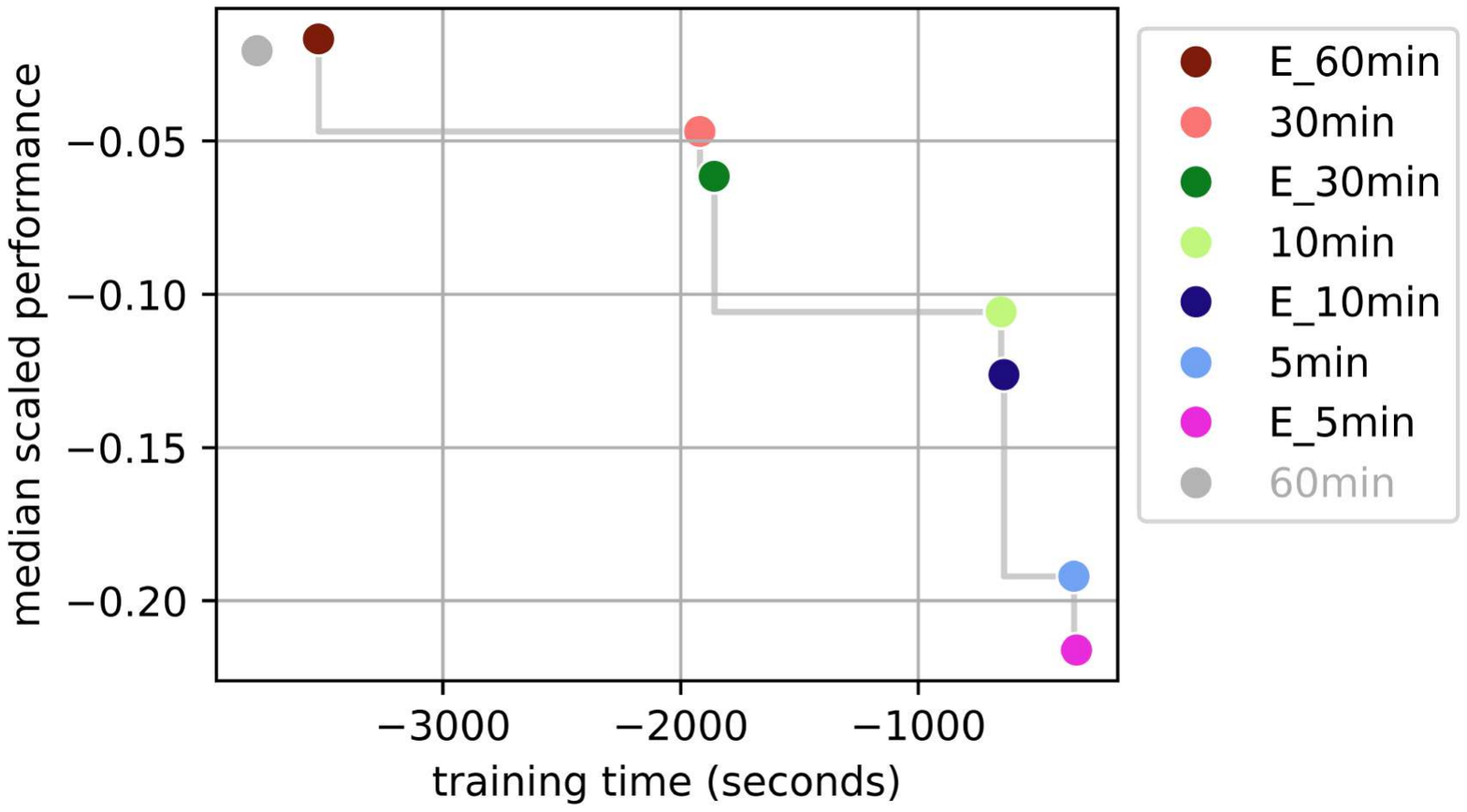}
\caption{\tiny AutoGluon(HQ)}
\end{subfigure}
\begin{subfigure}{0.31\textwidth}
\includegraphics[width=\linewidth]{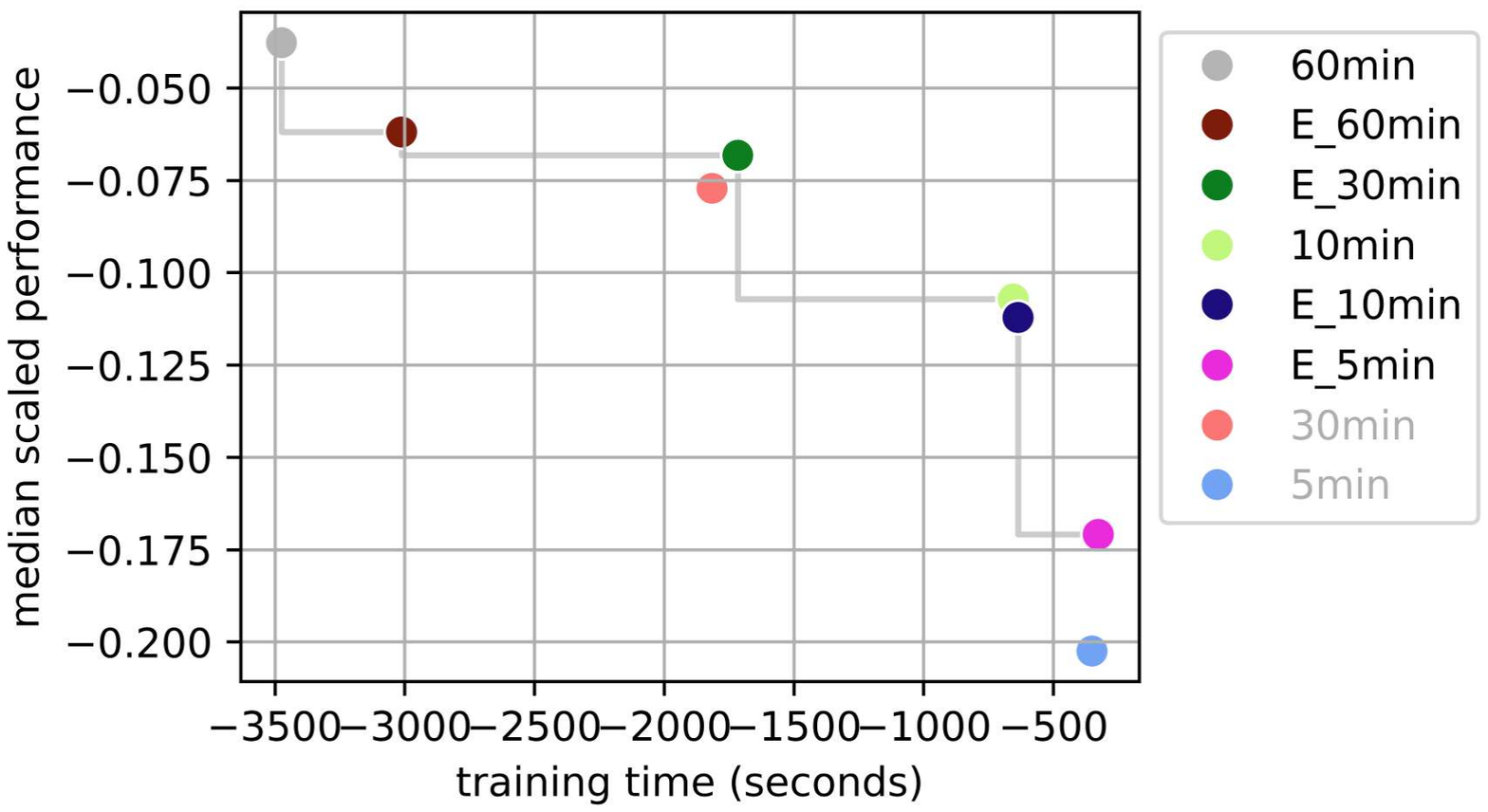}
\caption{\tiny AutoGluon(HQIL)}
\end{subfigure}
\begin{subfigure}{0.31\textwidth}
\includegraphics[width=\linewidth]{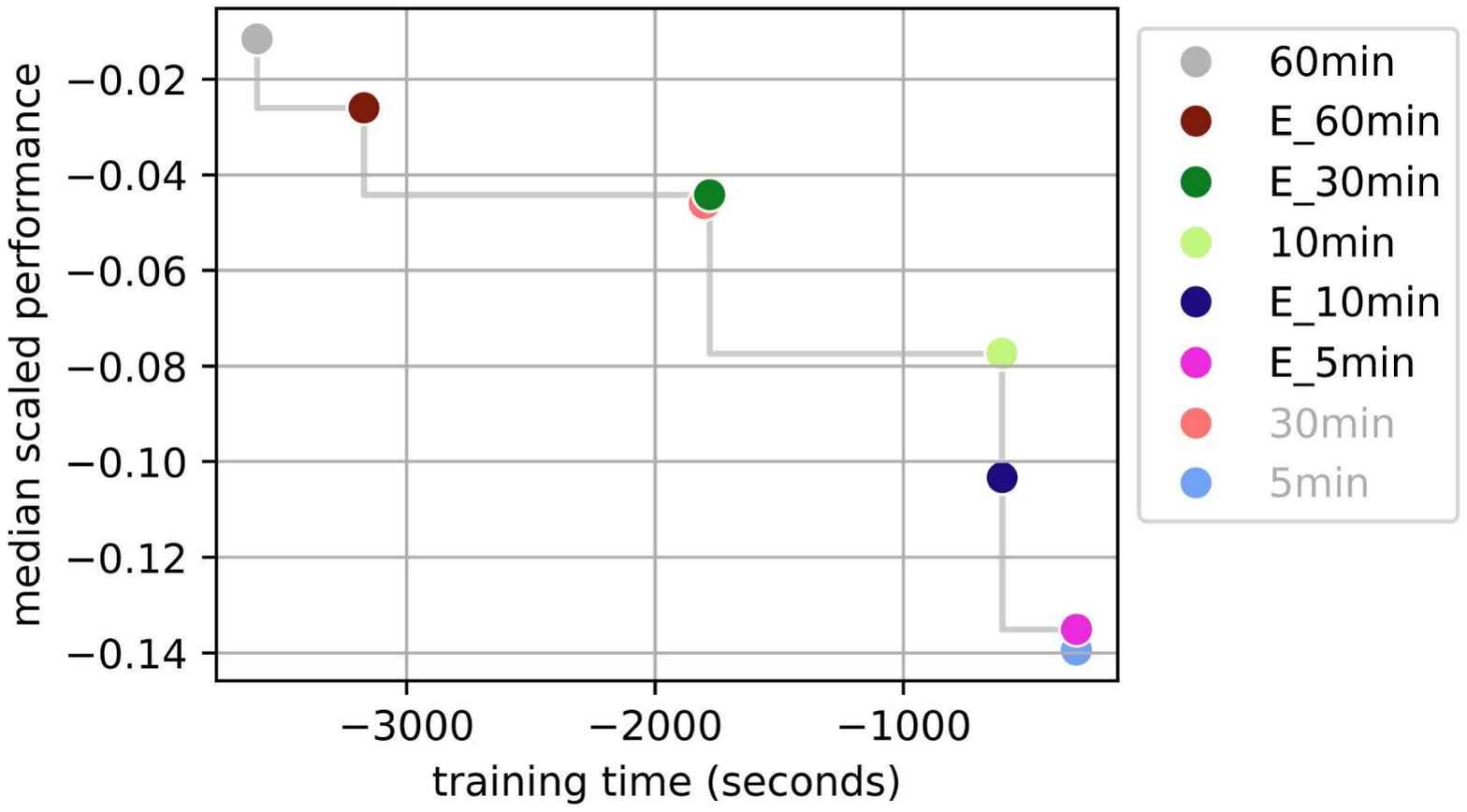}
\caption{\tiny AutoGluon(B)}
\end{subfigure}
\newline
\begin{subfigure}{0.31\textwidth}
\includegraphics[width=\linewidth]{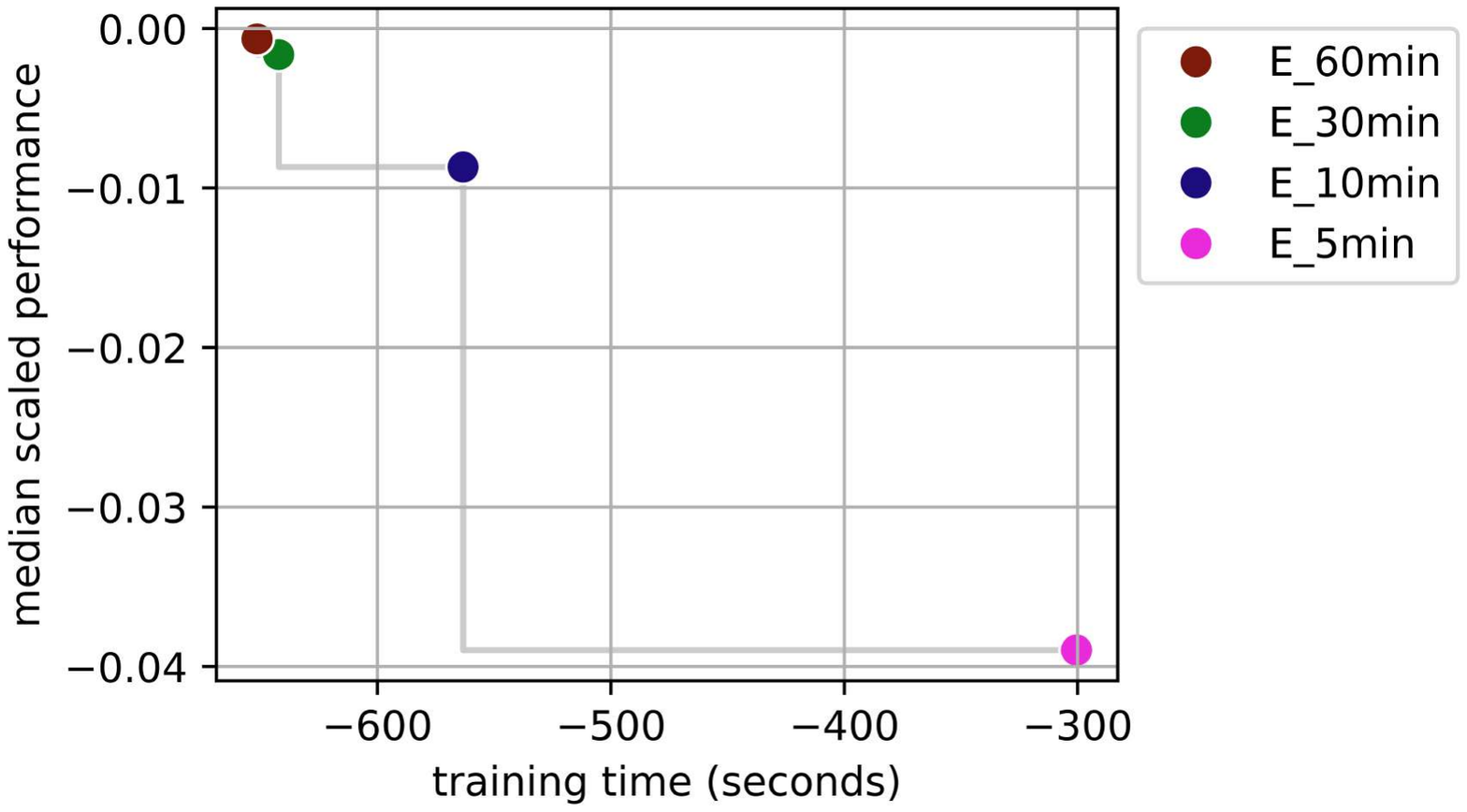}
\caption{\tiny AutoGluon(FIFTIL)}
\end{subfigure}
\begin{subfigure}{0.31\textwidth}
\includegraphics[width=\linewidth]{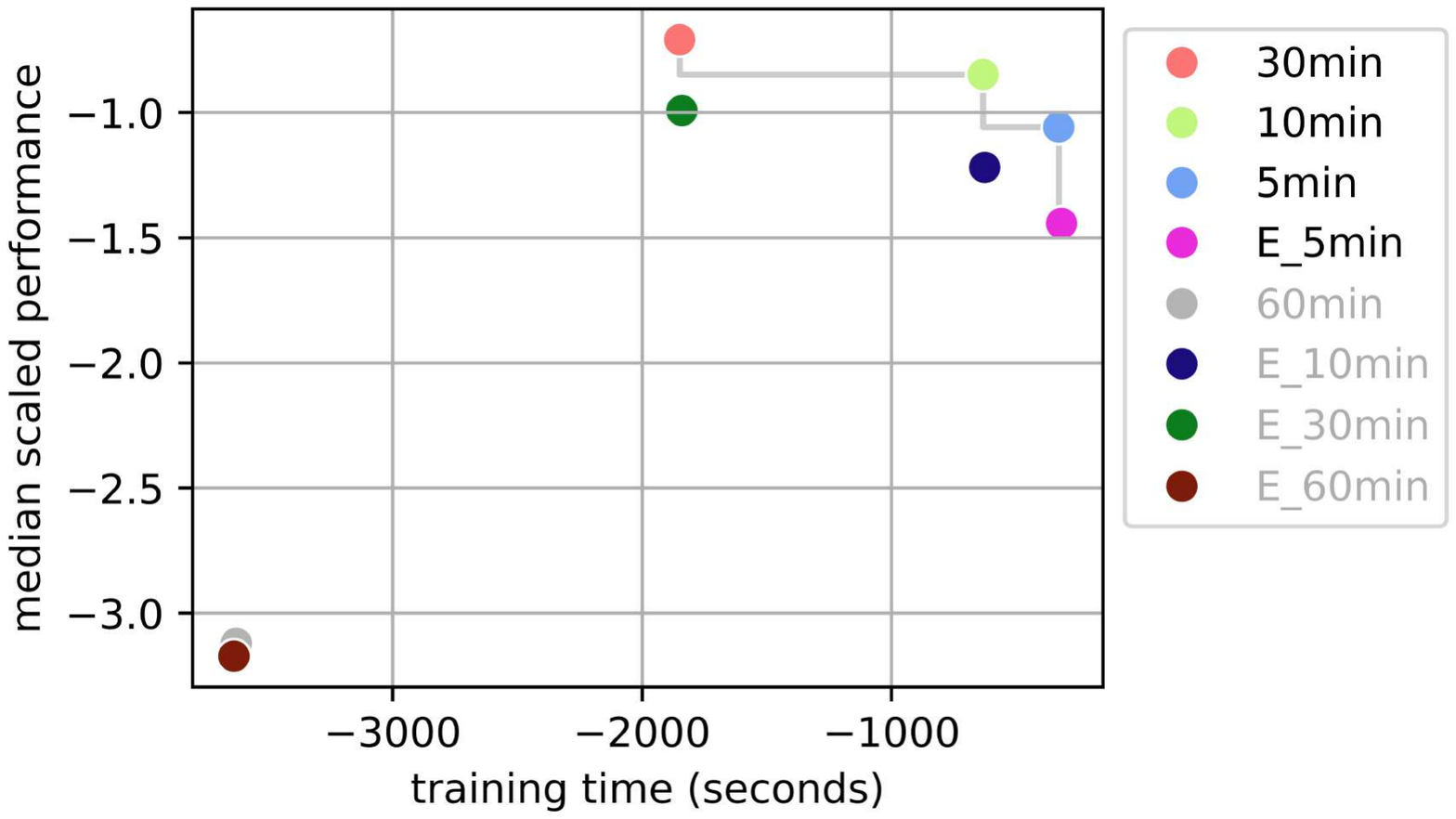}
\caption{\tiny FEDOT}
\end{subfigure}
\begin{subfigure}{0.31\textwidth}
\includegraphics[width=\linewidth]{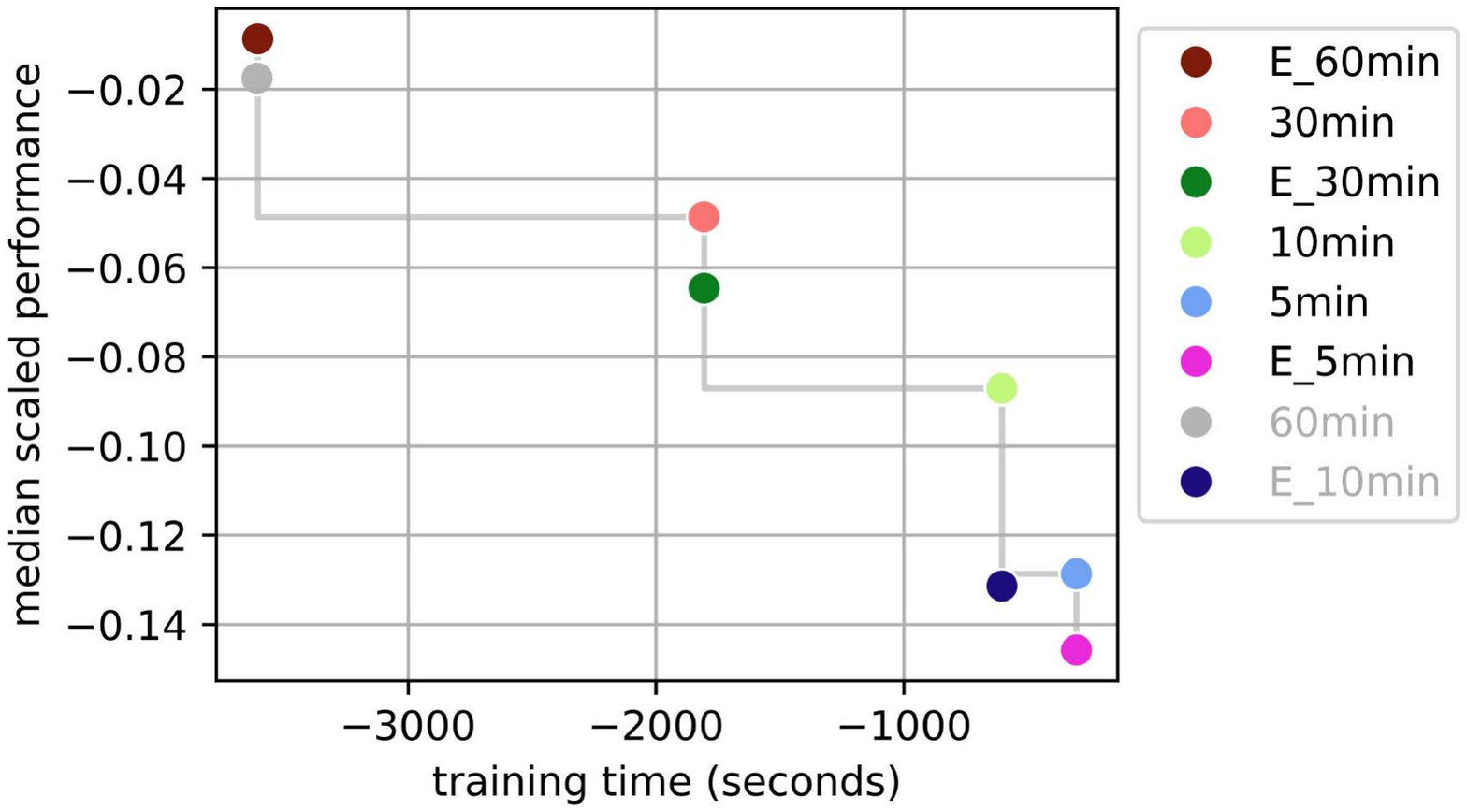}
\caption{\tiny flaml}
\end{subfigure}
\newline
\begin{subfigure}{0.31\textwidth}
\includegraphics[width=\linewidth]{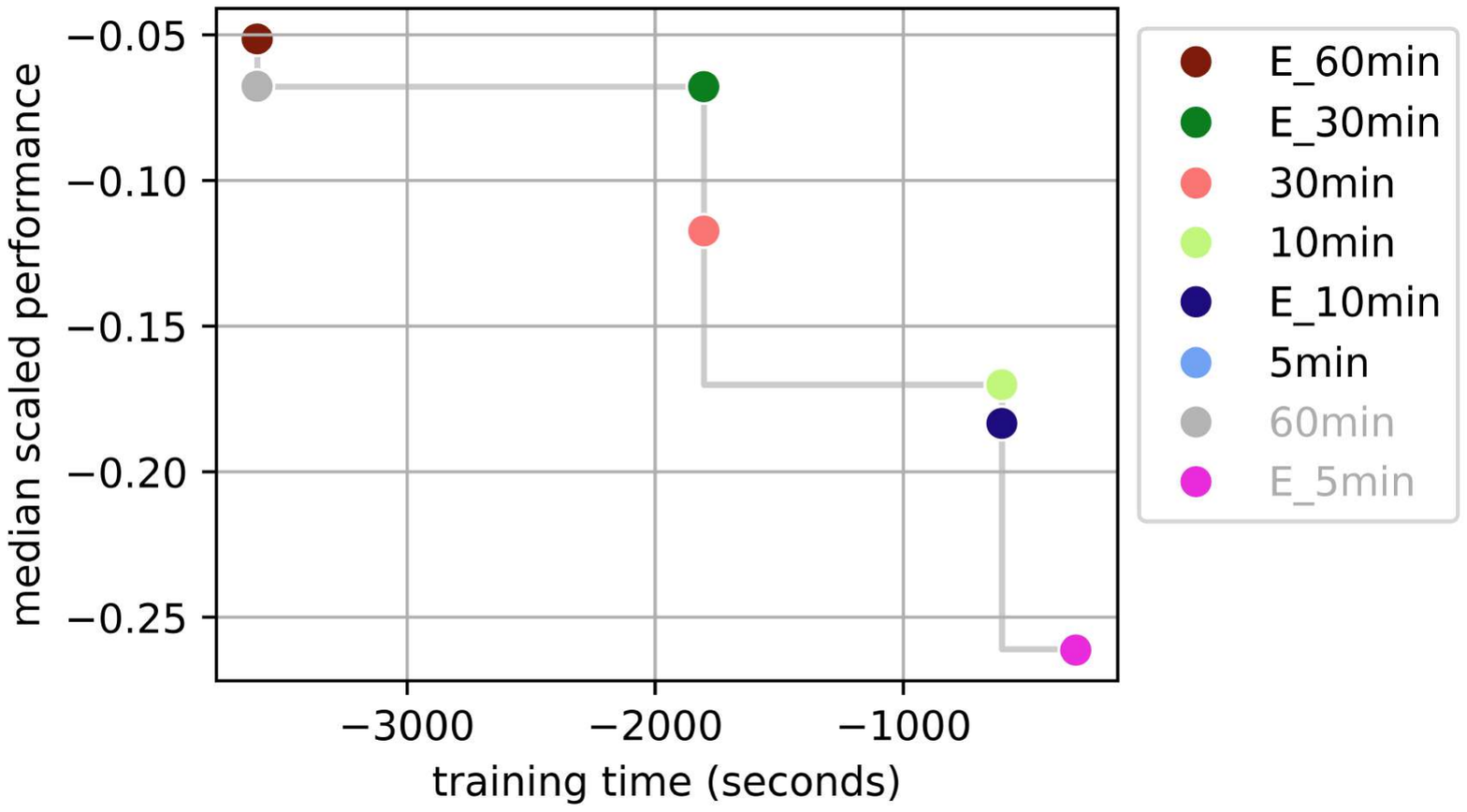}
\caption{\tiny H2OAutoML}
\end{subfigure}
\begin{subfigure}{0.31\textwidth}
\includegraphics[width=\linewidth]{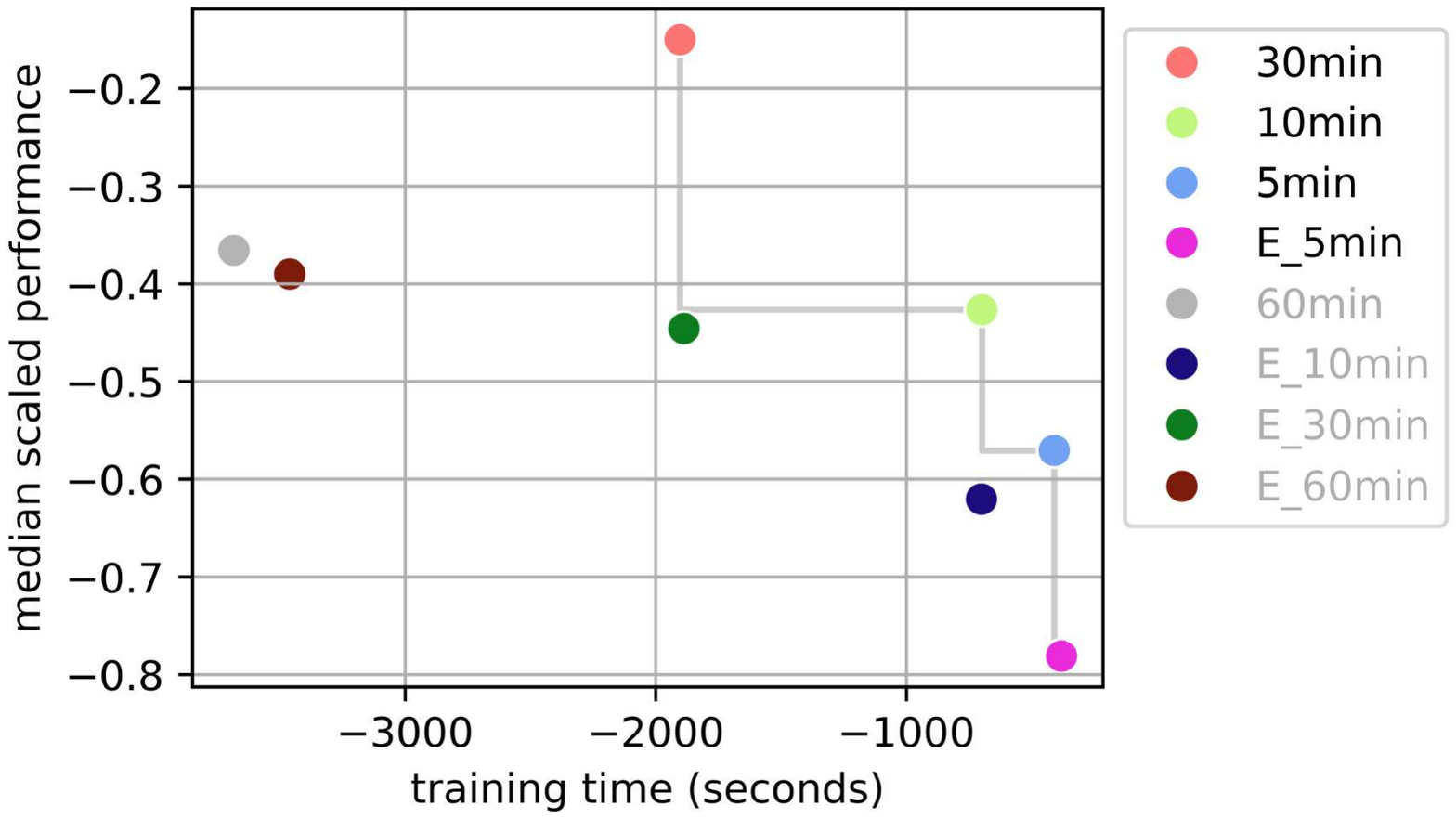}
\caption{\tiny TPOT}
\end{subfigure}

\end{center}
\caption{Pareto Frontier of the frameworks with and without early stopping. The plots show the performance values scaled from the random forest (-1) to best observed (0) versus the training time. \label{fig:appendix_pf_early_perf_training_time}}
\end{figure}

Similarly, Figure \ref{fig:appendix_pf_early_perf_training_time} examines the trade-off between performance and training time. Here, the trend indicates that frameworks that complete their training faster tend to sacrifice inference efficiency, meaning that while the training process is expedited, the model’s ability to make predictions quickly is diminished. This trade-off can occur because faster training frameworks often cut down on resource-intensive steps, such as hyperparameter tuning or complex feature engineering. However, while this results in shorter training durations, the model might be less optimized for real-time inference. Additionally, frameworks with early stopping sometimes exhibit a slight improvement in training time.

\begin{figure}[h]
\begin{center}

\begin{subfigure}{0.31\textwidth}
\includegraphics[width=\linewidth]{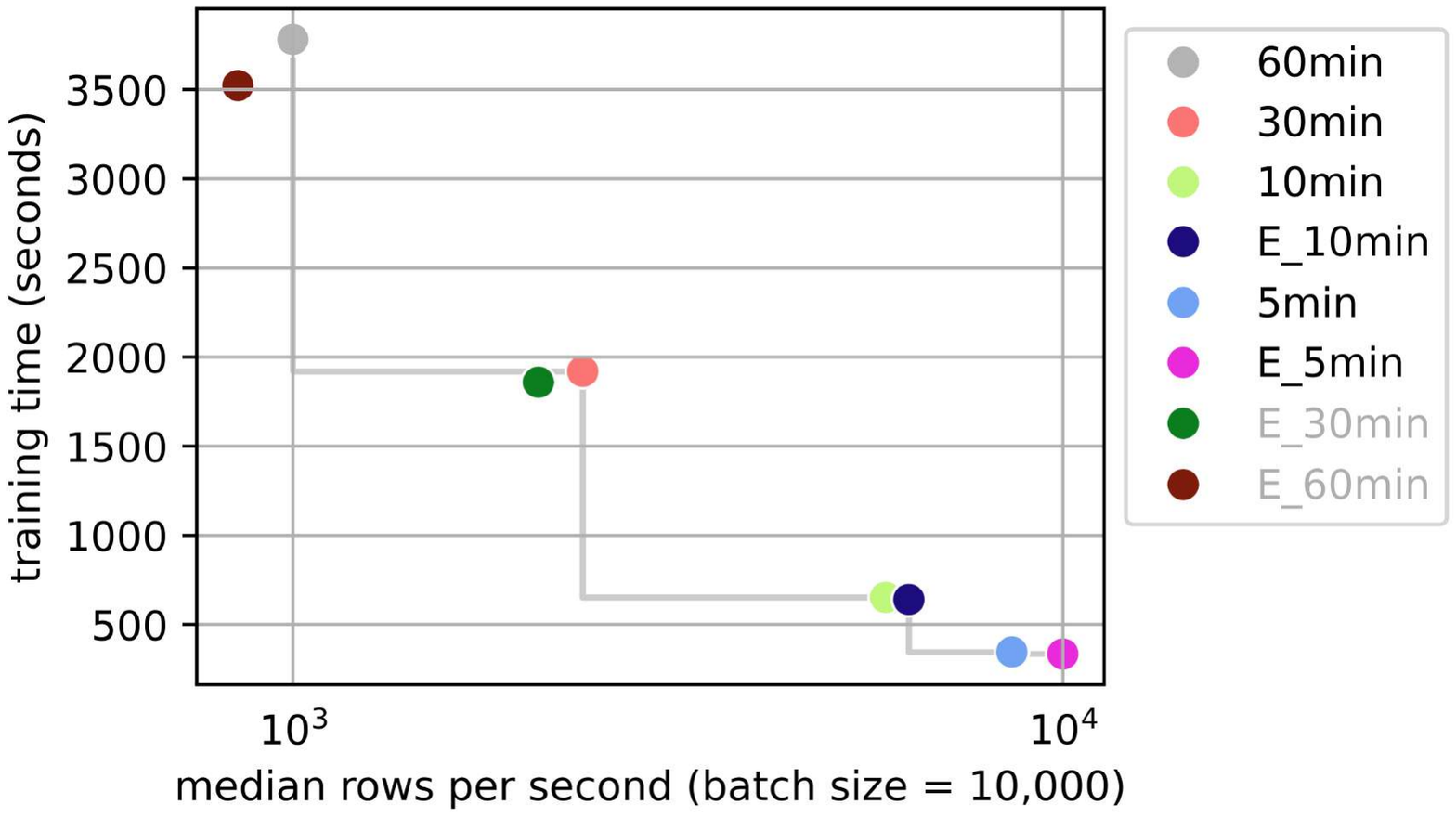}
\caption{\tiny AutoGluon(HQ)}
\end{subfigure}
\begin{subfigure}{0.31\textwidth}
\includegraphics[width=\linewidth]{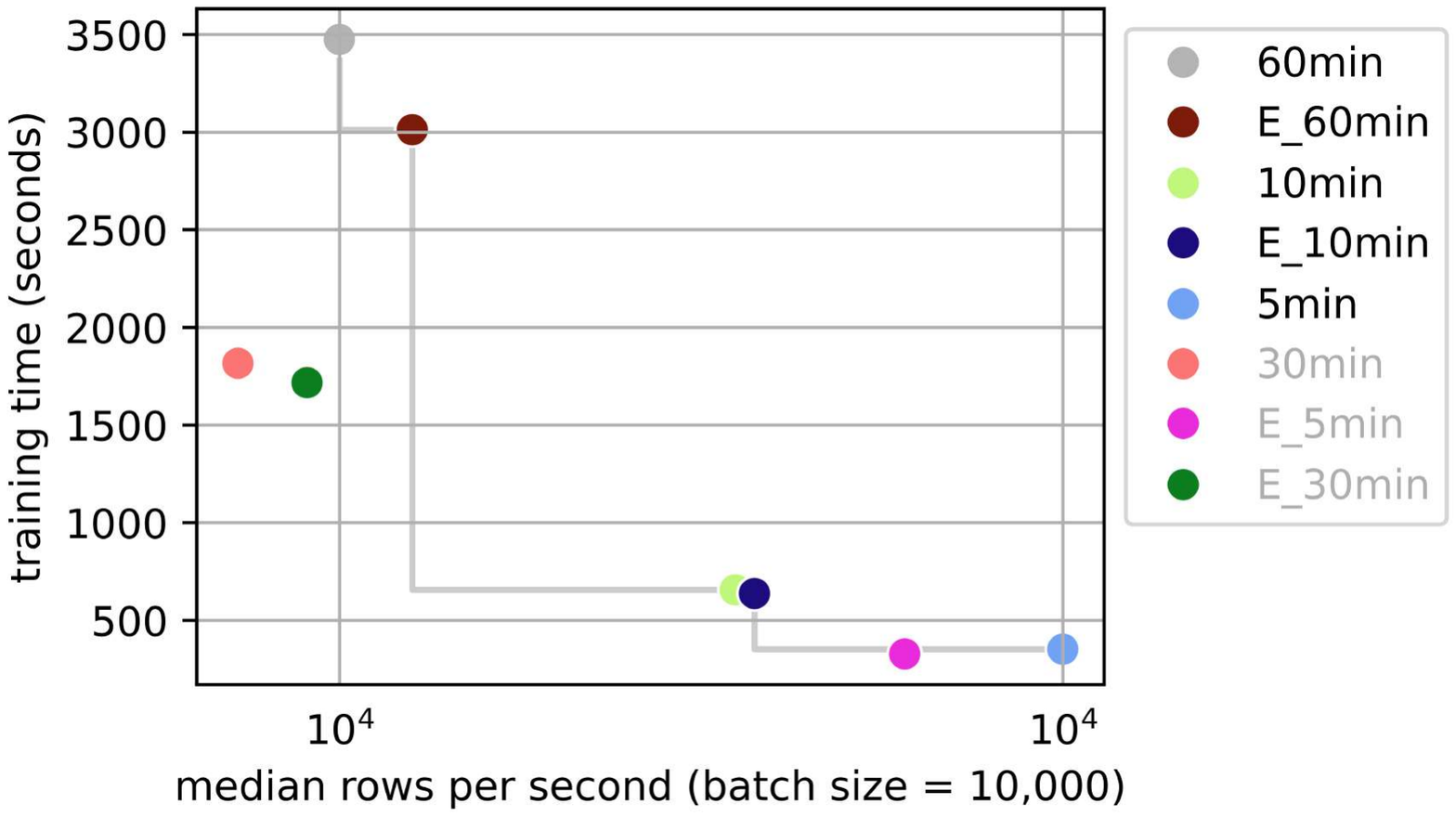}
\caption{\tiny AutoGluon(HQIL)}
\end{subfigure}
\begin{subfigure}{0.31\textwidth}
\includegraphics[width=\linewidth]{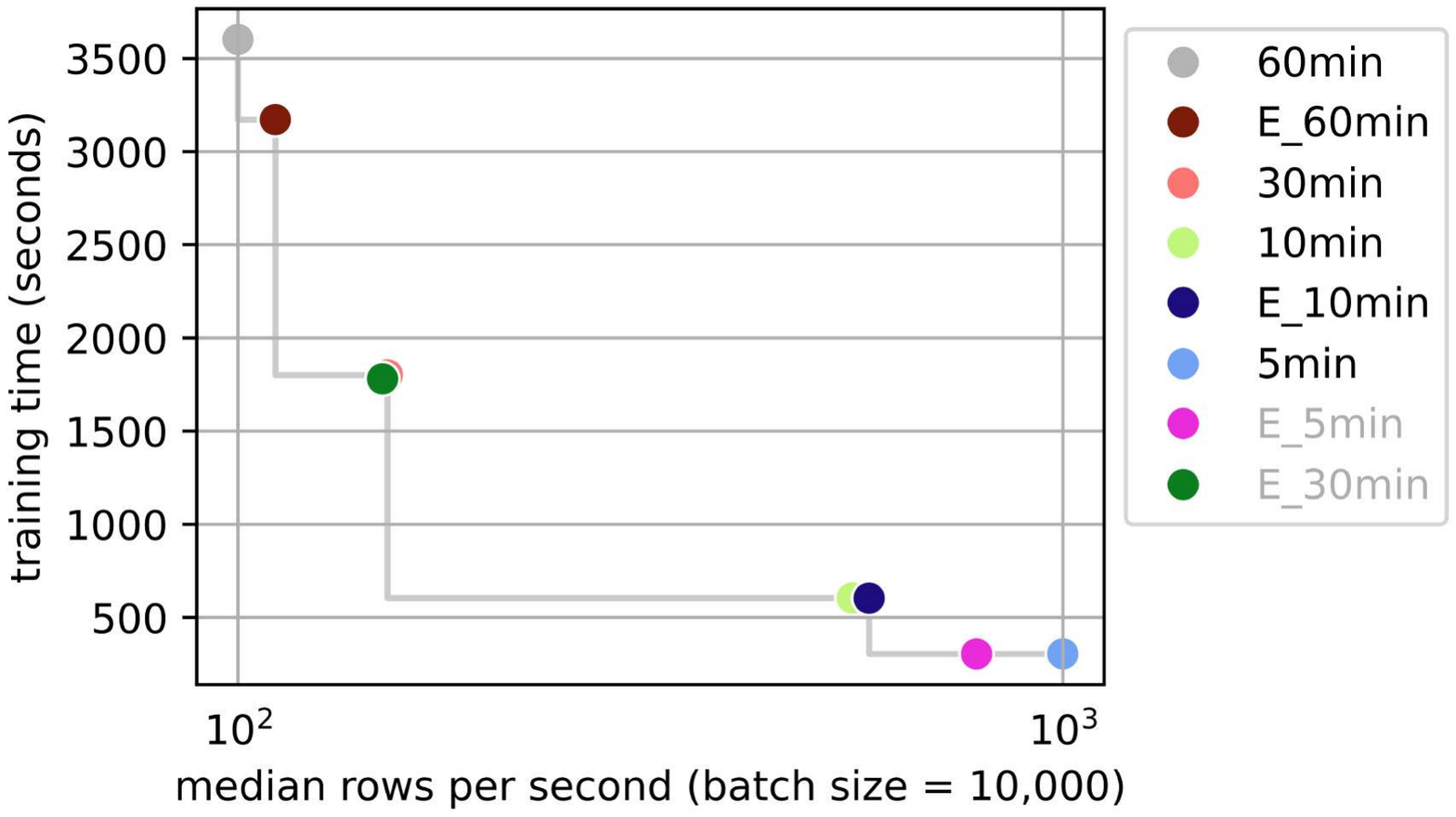}
\caption{\tiny AutoGluon(B)}
\end{subfigure}
\newline
\begin{subfigure}{0.31\textwidth}
\includegraphics[width=\linewidth]{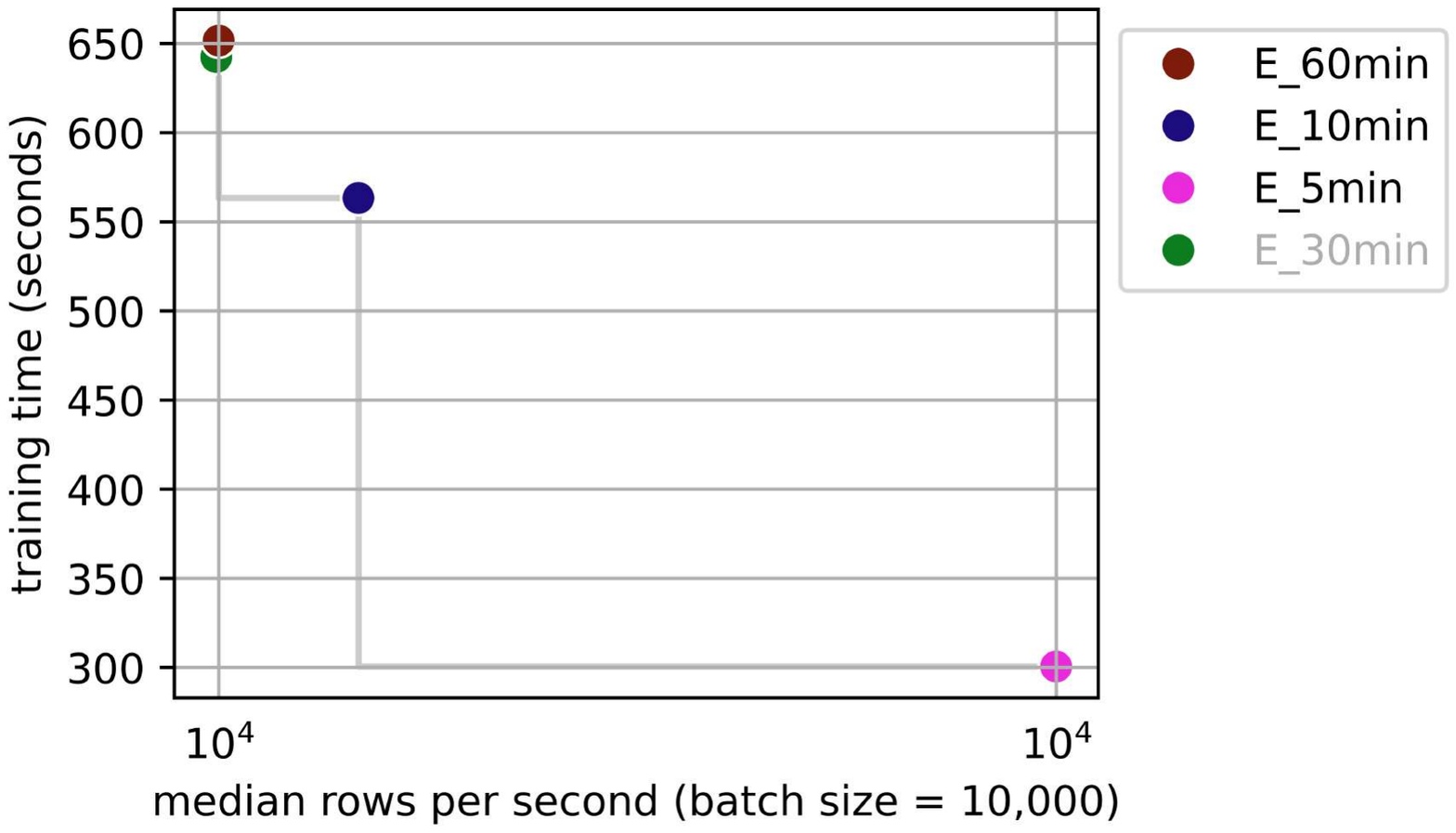}
\caption{\tiny AutoGluon(FIFTIL)}
\end{subfigure}
\begin{subfigure}{0.31\textwidth}
\includegraphics[width=\linewidth]{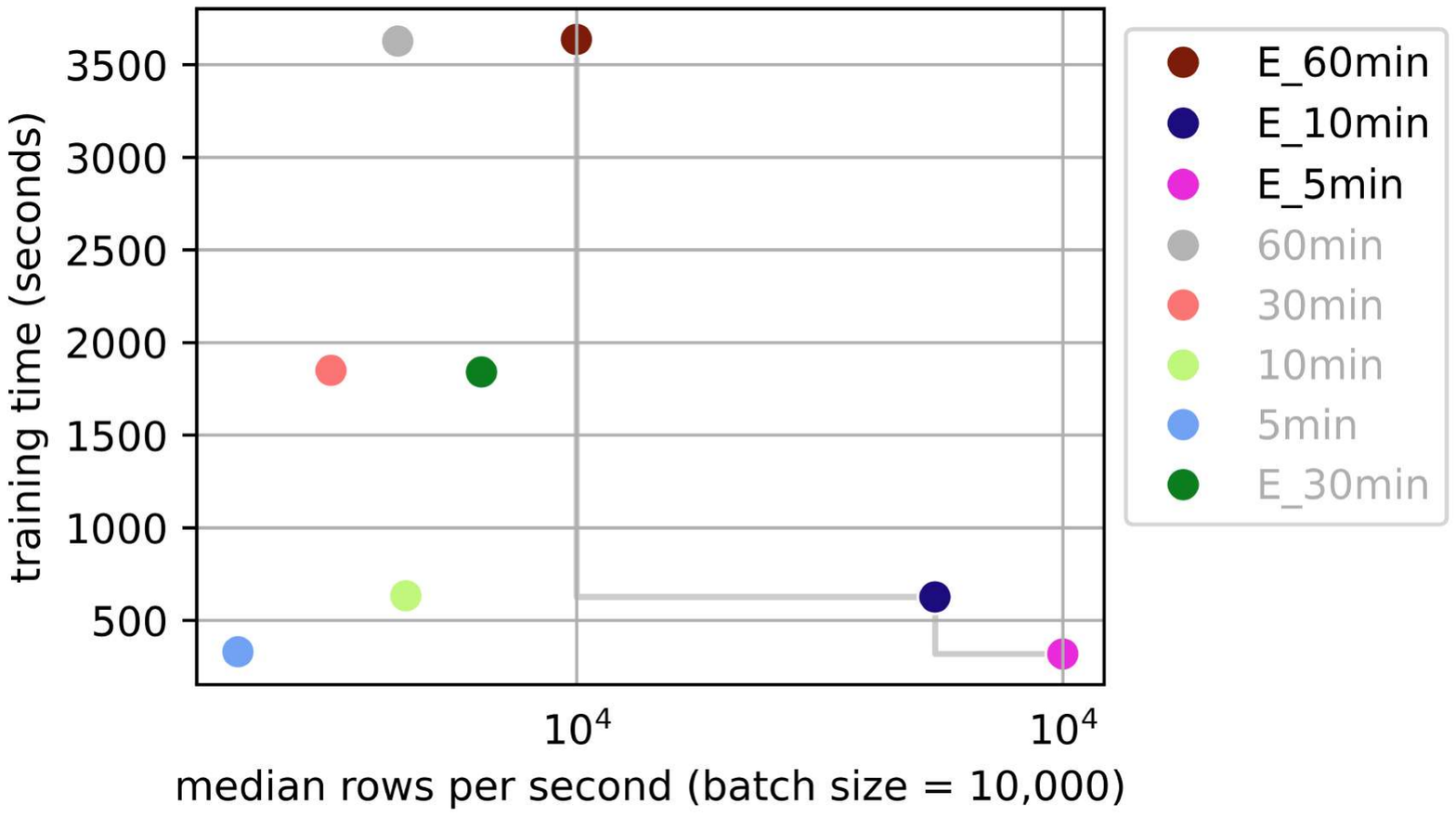}
\caption{\tiny FEDOT}
\end{subfigure}
\begin{subfigure}{0.31\textwidth}
\includegraphics[width=\linewidth]{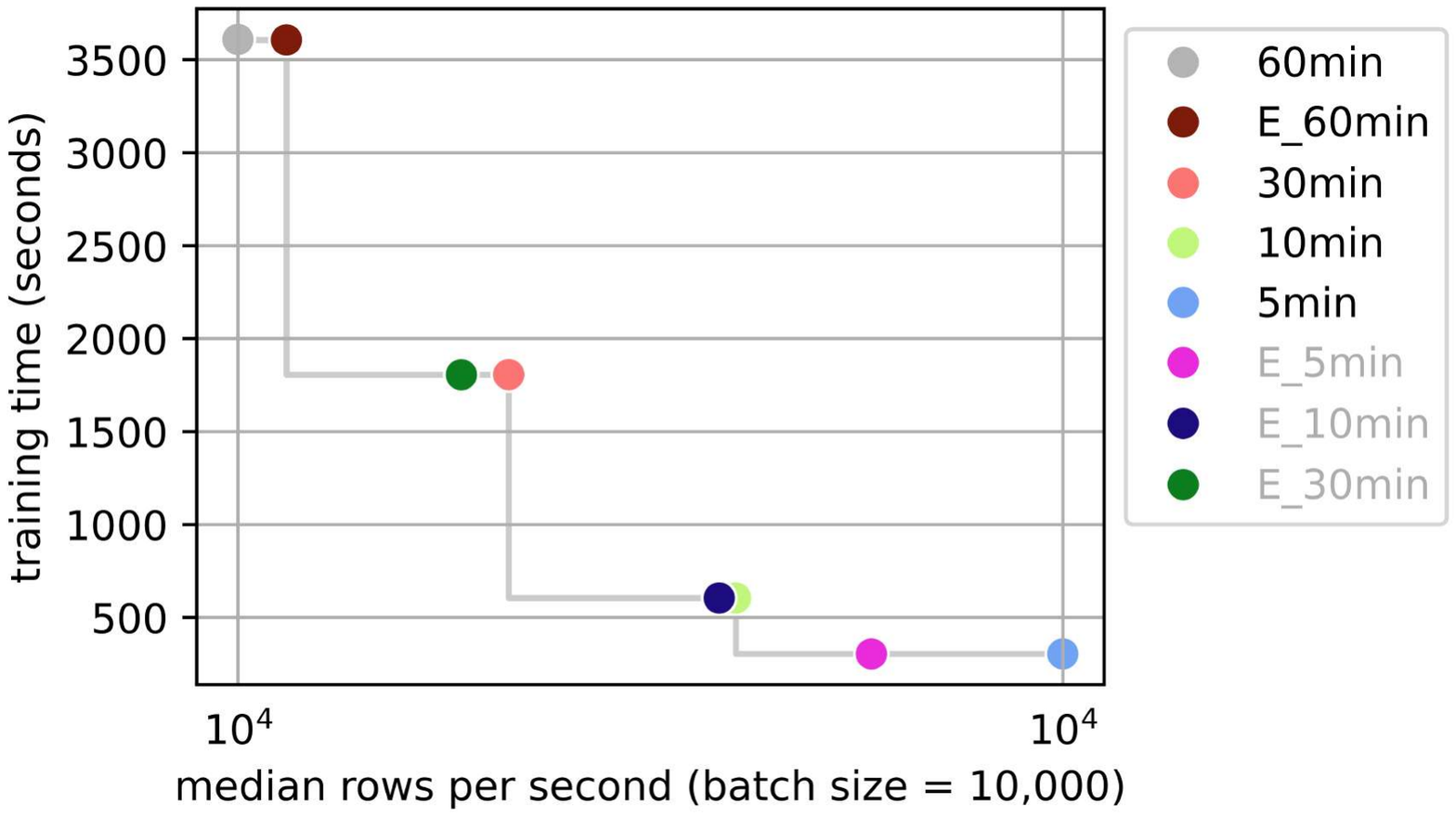}
\caption{\tiny flaml}
\end{subfigure}
\newline
\begin{subfigure}{0.31\textwidth}
\includegraphics[width=\linewidth]{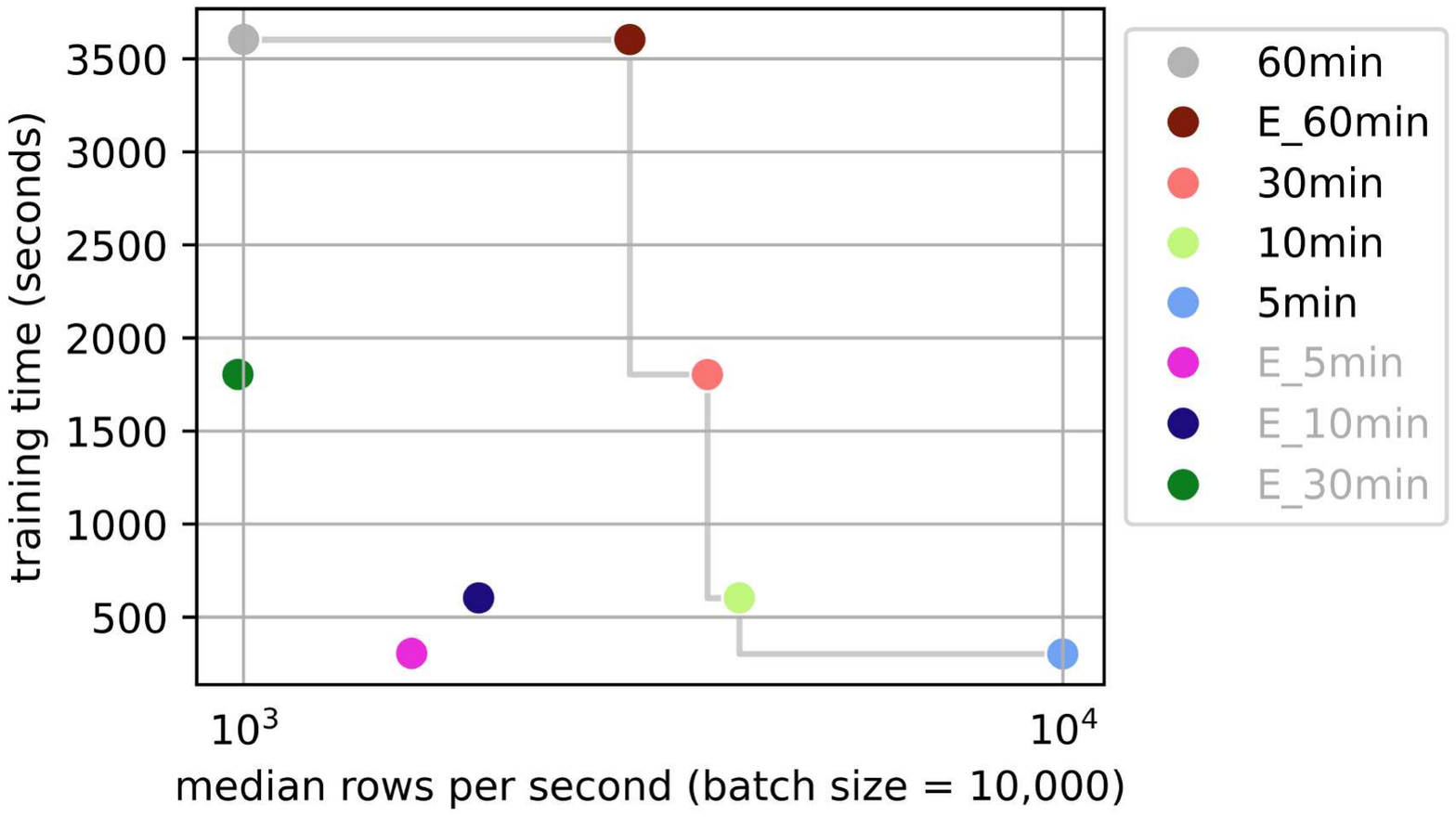}
\caption{\tiny H2OAutoML}
\end{subfigure}
\begin{subfigure}{0.31\textwidth}
\includegraphics[width=\linewidth]{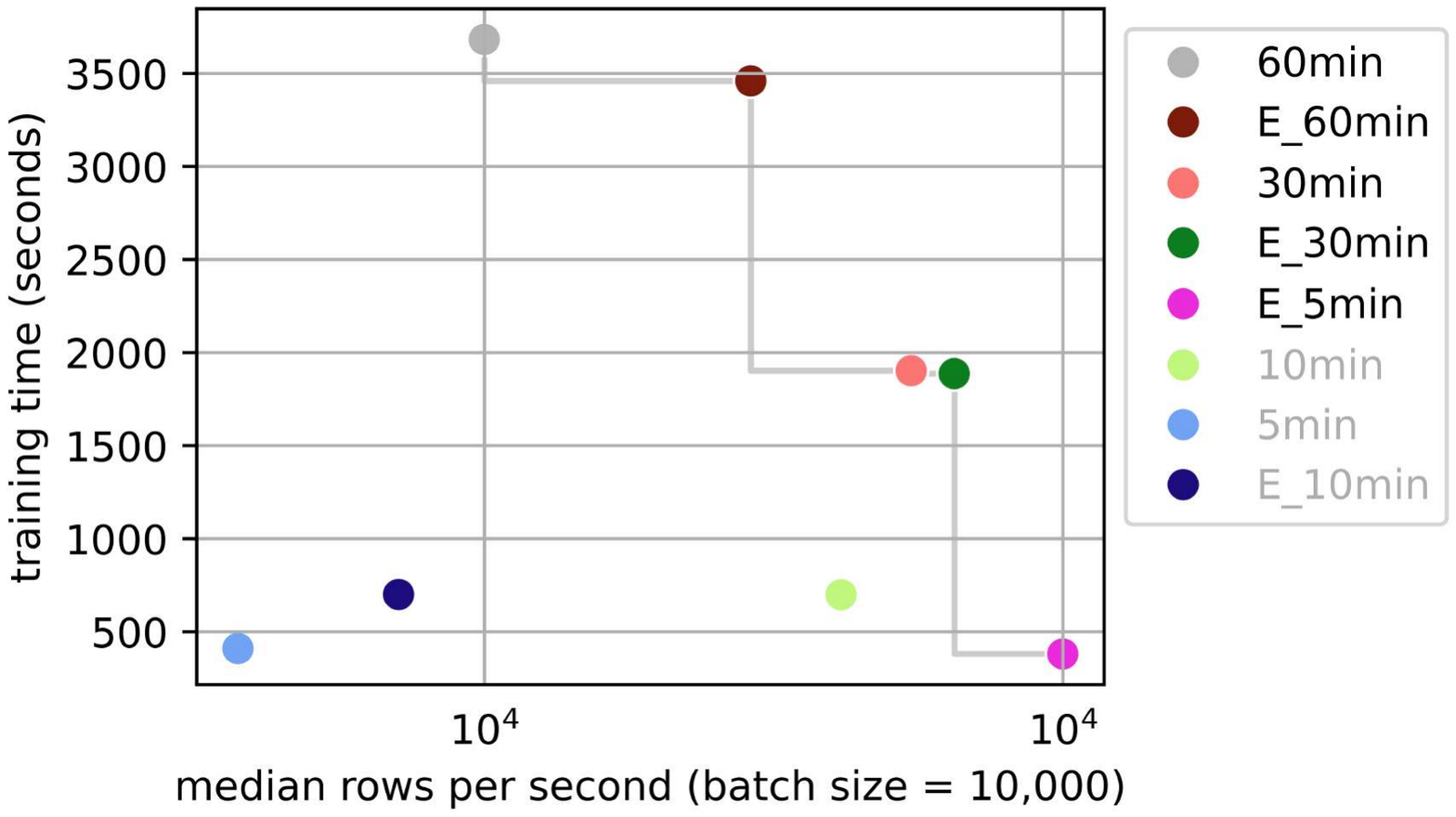}
\caption{\tiny TPOT}
\end{subfigure}

\end{center}
\caption{Pareto Frontier of the frameworks with and without early stopping. All time constraints are joined. The plots show the inference speed in median per second with a batch size of 10,000 versus the training time. \label{fig:appendix_pf_early_inference_training_time}}
\end{figure}

Finally, Figure \ref{fig:appendix_pf_early_inference_training_time} explores the balance between training time and inference time. Frameworks focusing on reducing computational costs during the training phase can result in larger or more complex models that require more resources during inference. Some frameworks that implement early stopping manage to reduce training time without substantially increasing inference time, ensuring an overall efficient performance during both phases. 

\clearpage
\subsection{Regret and time saved per framework}

\begin{figure}[h]
\begin{center}
\centering
\includegraphics[width=0.60\linewidth]{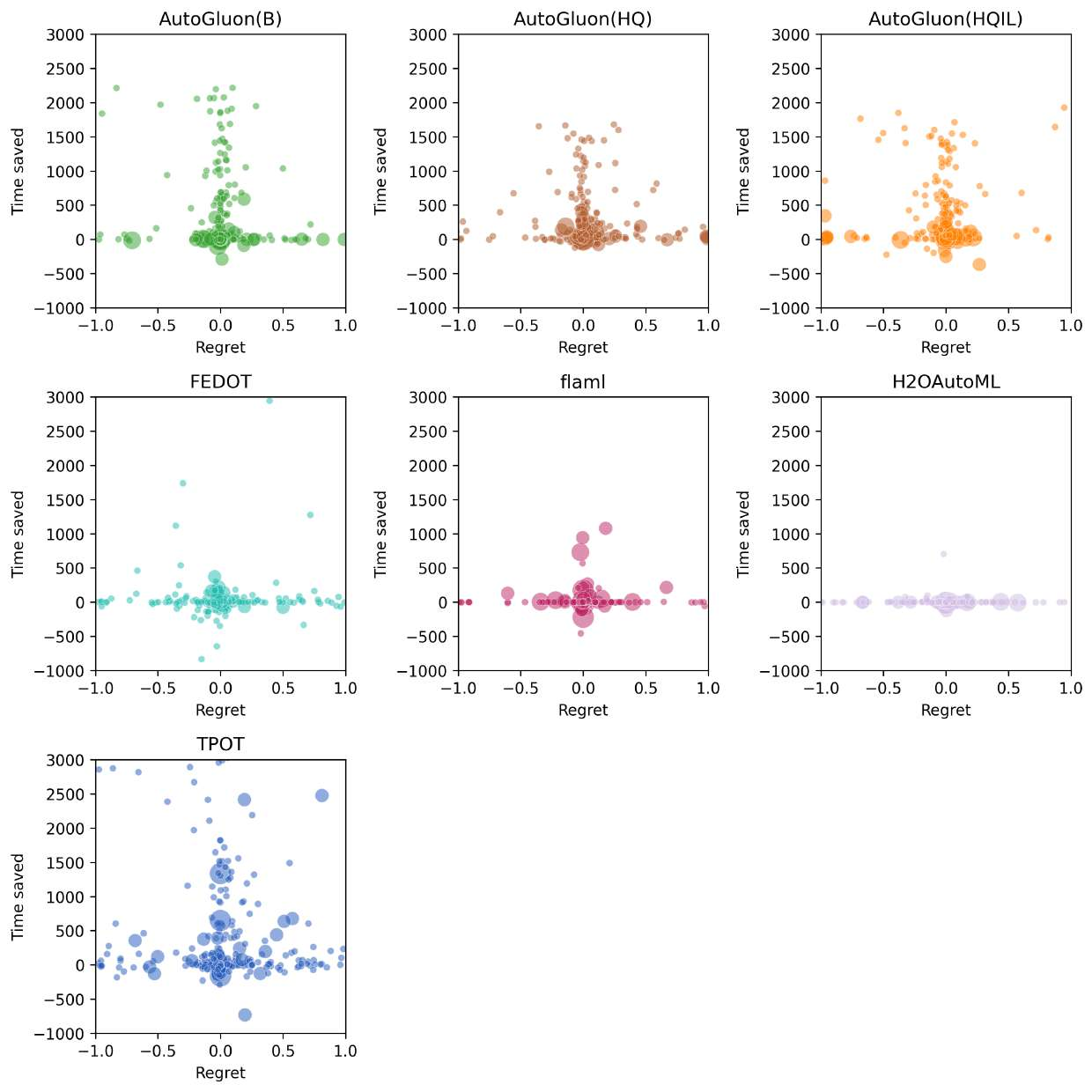}
\end{center}
\caption{\textit{Regret} vs. \textit{time saved} trade-off for different AutoML frameworks with early stopping mechanisms across various dataset sizes \label{fig:regret_delta_time_per_framework}}
\end{figure}

Lastly, in Figure \ref{fig:regret_delta_time_per_framework} demonstrates the \textit{regret} and \textit{time saved} trade-off per framework, where:

$\mathtt{AutoGluon}$ Variants: All three variants demonstrate a relatively concentrated range of \textit{regret} around 0, with some variance extending towards positive \textit{regret} (worse performance) but rarely surpassing a \textit{regret} of 0.5. $\mathtt{AutoGluon(B)}$ shows the highest variation in delta time, with a few instances reaching time savings exceeding 2000 seconds, but this often coincides with small negative or positive \textit{regret}. $\mathtt{AutoGluon(HQ)}$ has a cluster around $regret = 0$ with moderate time savings, but the spread is narrower compared to $\mathtt{AutoGluon(B)}$. $\mathtt{AutoGluon(HQIL)}$ has a similar pattern to $\mathtt{HQ}$, with more extreme \textit{time saved} values but less variation in \textit{regret}, indicating it tends to offer time savings without significantly degrading performance.

$\mathtt{FEDOT}$ maintains a very tight clustering around $regret = 0$, with most values lying close to zero. This indicates that early stopping in FEDOT rarely harms performance, and time varies minimally. 

$\mathtt{flaml}$ is similar to $\mathtt{FEDOT}$, $\mathtt{flaml}$ shows time-saving up for 1000 seconds for some tasks while showing minimal time differences. The effectiveness of $\mathtt{flaml}$ with early stopping is highly dataset-dependent.

$\mathtt{H2OAutoML}$ displays very little variation in both \textit{regret} and \textit{time saved}, with \textit{regret} values concentrated around 0. This suggests that early stopping in $\mathtt{H2OAutoML}$ has a very minimal impact on both time savings and performance degradation. Essentially, $\mathtt{H2OAutoML}$ remains consistent but shows little benefit from early stopping in terms of time savings.

$\mathtt{TPOT}$ presents the widest range of delta time savings and \textit{regret}, with \textit{regret} spanning from -1 to 1. This indicates that TPOT is highly sensitive to early stopping, leading to substantial performance changes (both positive and negative). The time saving often comes at the cost of increased \textit{regret}, particularly with larger datasets (larger bubbles). $\mathtt{TPOT}$’s trade-off between time savings and performance degradation is the most pronounced of all the frameworks.

\end{document}